\PassOptionsToPackage{dvipsnames}{xcolor}
\PassOptionsToPackage{export}{adjustbox}
\documentclass{article}

\usepackage{microtype}
\usepackage{graphicx}
\usepackage{booktabs} %
\usepackage{textcomp} %

\usepackage{tikz}
\usetikzlibrary{arrows, positioning, calc, decorations.pathreplacing}
\usepackage[mode=buildnew]{standalone} %
\usepackage{placeins} %
\usepackage{subcaption}

\usepackage{xurl}
\usepackage[breaklinks=true]{hyperref}

\usepackage[arxiv]{icml2024}

\usepackage{amsmath}
\usepackage{amssymb}
\usepackage{mathtools}
\usepackage{amsthm}
\usepackage{multirow}

\usepackage[capitalize,noabbrev]{cleveref}

\crefname{section}{Sec.}{Secs.}
\Crefname{section}{Section}{Sections}
\Crefname{table}{Table}{Tables}
\crefname{table}{Tab.}{Tabs.}

\newcommand{\figref}[1]{Fig.~\ref{#1}}

\theoremstyle{plain}

\theoremstyle{definition}

\theoremstyle{remark}

\usepackage[textsize=tiny]{todonotes}

\usepackage{etoolbox}
\newcommand{\settablefontsize}{\footnotesize}

\newcommand{\s}{\hphantom{0}}

\AtBeginEnvironment{tabular}{\settablefontsize}
\AtBeginEnvironment{table}{\settablefontsize}

\usepackage{caption}
\usepackage{enumitem}

\icmltitlerunning{Scaling Rectified Flow Transformers for High-Resolution Image Synthesis}

\usepackage{xspace}
\newcommand{\modelname}{\emph{MM-DiT}\xspace}
\usepackage{xcolor}
\newcommand{\TODO}[1]{\textcolor{red}{#1}}

\usepackage{bbm}

\makeatletter
\DeclareRobustCommand\onedot{\futurelet\@let@token\@onedot}
\def\@onedot{\ifx\@let@token.\else.\null\fi\xspace}

\def\eg{\emph{e.g}\onedot} 
\def\ie{\emph{i.e}\onedot} 
\def\cf{\emph{c.f}\onedot} 
 \def\vs{\emph{vs}\onedot}

\makeatother

\begin{document}

\twocolumn[
\icmltitle{Scaling Rectified Flow Transformers for High-Resolution Image Synthesis} %

\icmlsetsymbol{equal}{*}

\begin{icmlauthorlist}
\vspace{-1em}
\icmlauthor{Patrick Esser}{equal}\;
\icmlauthor{Sumith Kulal}{}\;
\icmlauthor{Andreas Blattmann}{}\;
\icmlauthor{Rahim Entezari}{}\;
\icmlauthor{Jonas Müller}{}\;
\icmlauthor{Harry Saini}{}\;
\icmlauthor{Yam Levi}{}\;
\icmlauthor{Dominik Lorenz}{}\,
\icmlauthor{Axel Sauer}{}\,
\icmlauthor{Frederic Boesel}{}\,
\icmlauthor{Dustin Podell}{}\,
\icmlauthor{Tim Dockhorn}{}\,
\icmlauthor{Zion English}{}\,

\icmlauthor{Kyle Lacey}{}\,
\icmlauthor{Alex Goodwin}{}\,
\icmlauthor{Yannik Marek}{}\,
\icmlauthor{Robin Rombach}{equal}\\
Stability AI
\end{icmlauthorlist}

\icmlcorrespondingauthor{}{\textless{}first.last\textgreater{}@stability.ai}
\icmlkeywords{Machine Learning, ICML}

{\vspace{0.75em}%
\includegraphics[width=0.99\textwidth]{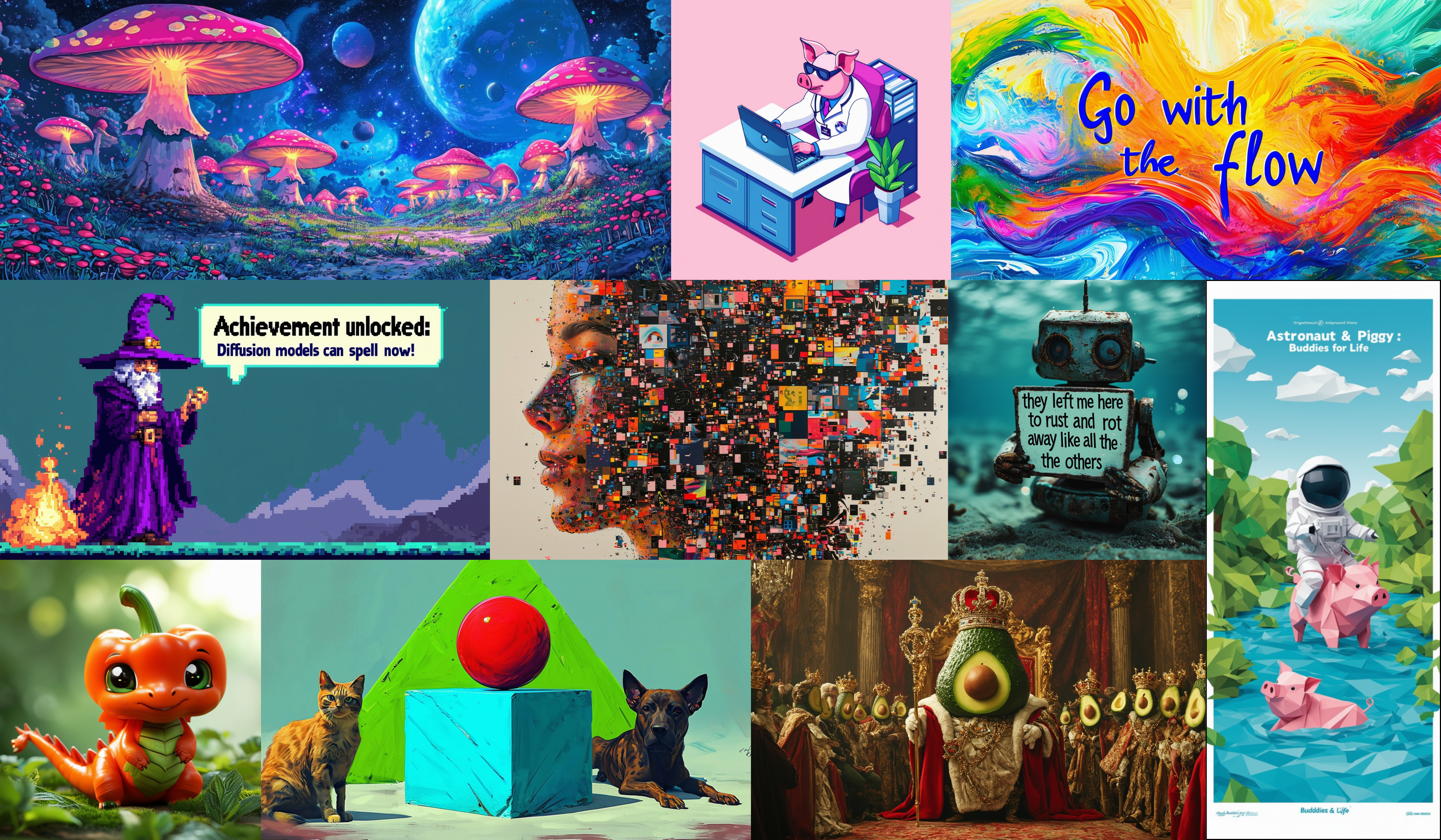}%
\vspace{-1em}
\captionof{figure}{
High-resolution samples from our 8B rectified flow model, showcasing its capabilities in typography, precise prompt following and spatial reasoning, attention to fine details, and high image quality across a wide variety of styles.
\vspace{-0.75em}}%
\label{fig:teaser}%
}

\vskip 0.3in
]

\printAffiliationsAndNotice{\icmlEqualContribution} %

\newcommand{\hscomment}[1]{\noindent{\textcolor{blue}{\textbf{\#\#\# HS:} \textsf{#1} \#\#\#}}}
\newcommand{\recomment}[1]{\noindent{\textcolor{blue}{\textbf{\#\#\# RE:} \textsf{#1} \#\#\#}}}
\newcommand{\oscomment}[1]{\noindent{\textcolor{blue}{\textbf{\#\#\# OS:} \textsf{#1} \#\#\#}}}
\newcommand{\nccomment}[1]{\noindent{\textcolor{blue}{\textbf{\#\#\# NC:} \textsf{#1} \#\#\#}}}
\newcommand{\bncomment}[1]{\noindent{\textcolor{blue}{\textbf{\#\#\# BN:} \textsf{#1} \#\#\#}}}

\newcommand{\F}{\text{Fr}}
\newcommand{\KL}{\text{KL}}

\newcommand{\figleft}{{m (Left)}}
\newcommand{\figcenter}{{\em (Center)}}
\newcommand{\figright}{{\em (Right)}}
\newcommand{\figtop}{{\em (Top)}}
\newcommand{\figbottom}{{\em (Bottom)}}
\newcommand{\captiona}{{\em (a)}}
\newcommand{\captionb}{{\em (b)}}
\newcommand{\captionc}{{\em (c)}}
\newcommand{\captiond}{{\em (d)}}

\newcommand{\newterm}[1]{{\bf #1}}

\def\figref#1{figure~\ref{#1}}
\def\Figref#1{Figure~\ref{#1}}
\def\twofigref#1#2{figures \ref{#1} and \ref{#2}}
\def\quadfigref#1#2#3#4{figures \ref{#1}, \ref{#2}, \ref{#3} and \ref{#4}}
\def\secref#1{section~\ref{#1}}
\def\Secref#1{Section~\ref{#1}}
\def\twosecrefs#1#2{sections \ref{#1} and \ref{#2}}
\def\secrefs#1#2#3{sections \ref{#1}, \ref{#2} and \ref{#3}}
\def\eqref#1{Equation~\ref{#1}}
\def\Eqref#1{Equation~\ref{#1}}
\def\plaineqref#1{\ref{#1}}
\def\chapref#1{chapter~\ref{#1}}
\def\Chapref#1{Chapter~\ref{#1}}
\def\rangechapref#1#2{chapters\ref{#1}--\ref{#2}}
\def\algref#1{algorithm~\ref{#1}}
\def\Algref#1{Algorithm~\ref{#1}}
\def\twoalgref#1#2{algorithms \ref{#1} and \ref{#2}}
\def\Twoalgref#1#2{Algorithms \ref{#1} and \ref{#2}}
\def\partref#1{part~\ref{#1}}
\def\Partref#1{Part~\ref{#1}}
\def\twopartref#1#2{parts \ref{#1} and \ref{#2}}

\def\ceil#1{\lceil #1 \rceil}
\def\floor#1{\lfloor #1 \rfloor}
\def\1{\bm{1}}
\newcommand{\train}{\mathcal{D}}
\newcommand{\valid}{\mathcal{D_{\mathrm{valid}}}}
\newcommand{\test}{\mathcal{D_{\mathrm{test}}}}

\def\eps{{\epsilon}}

\def\reta{{\textnormal{$\eta$}}}
\def\ra{{\textnormal{a}}}
\def\rb{{\textnormal{b}}}
\def\rc{{\textnormal{c}}}
\def\rd{{\textnormal{d}}}
\def\re{{\textnormal{e}}}
\def\rf{{\textnormal{f}}}
\def\rg{{\textnormal{g}}}
\def\rh{{\textnormal{h}}}
\def\ri{{\textnormal{i}}}
\def\rj{{\textnormal{j}}}
\def\rk{{\textnormal{k}}}
\def\rl{{\textnormal{l}}}
\def\rn{{\textnormal{n}}}
\def\ro{{\textnormal{o}}}
\def\rp{{\textnormal{p}}}
\def\rq{{\textnormal{q}}}
\def\rr{{\textnormal{r}}}
\def\rs{{\textnormal{s}}}
\def\rt{{\textnormal{t}}}
\def\ru{{\textnormal{u}}}
\def\rv{{\textnormal{v}}}
\def\rw{{\textnormal{w}}}
\def\rx{{\textnormal{x}}}
\def\ry{{\textnormal{y}}}
\def\rz{{\textnormal{z}}}

\def\rvepsilon{{\mathbf{\epsilon}}}
\def\rvtheta{{\mathbf{\theta}}}
\def\rva{{\mathbf{a}}}
\def\rvb{{\mathbf{b}}}
\def\rvc{{\mathbf{c}}}
\def\rvd{{\mathbf{d}}}
\def\rve{{\mathbf{e}}}
\def\rvf{{\mathbf{f}}}
\def\rvg{{\mathbf{g}}}
\def\rvh{{\mathbf{h}}}
\def\rvu{{\mathbf{i}}}
\def\rvj{{\mathbf{j}}}
\def\rvk{{\mathbf{k}}}
\def\rvl{{\mathbf{l}}}
\def\rvm{{\mathbf{m}}}
\def\rvn{{\mathbf{n}}}
\def\rvo{{\mathbf{o}}}
\def\rvp{{\mathbf{p}}}
\def\rvq{{\mathbf{q}}}
\def\rvr{{\mathbf{r}}}
\def\rvs{{\mathbf{s}}}
\def\rvt{{\mathbf{t}}}
\def\rvu{{\mathbf{u}}}
\def\rvv{{\mathbf{v}}}
\def\rvw{{\mathbf{w}}}
\def\rvx{{\mathbf{x}}}
\def\rvy{{\mathbf{y}}}
\def\rvz{{\mathbf{z}}}

\def\erva{{\textnormal{a}}}
\def\ervb{{\textnormal{b}}}
\def\ervc{{\textnormal{c}}}
\def\ervd{{\textnormal{d}}}
\def\erve{{\textnormal{e}}}
\def\ervf{{\textnormal{f}}}
\def\ervg{{\textnormal{g}}}
\def\ervh{{\textnormal{h}}}
\def\ervi{{\textnormal{i}}}
\def\ervj{{\textnormal{j}}}
\def\ervk{{\textnormal{k}}}
\def\ervl{{\textnormal{l}}}
\def\ervm{{\textnormal{m}}}
\def\ervn{{\textnormal{n}}}
\def\ervo{{\textnormal{o}}}
\def\ervp{{\textnormal{p}}}
\def\ervq{{\textnormal{q}}}
\def\ervr{{\textnormal{r}}}
\def\ervs{{\textnormal{s}}}
\def\ervt{{\textnormal{t}}}
\def\ervu{{\textnormal{u}}}
\def\ervv{{\textnormal{v}}}
\def\ervw{{\textnormal{w}}}
\def\ervx{{\textnormal{x}}}
\def\ervy{{\textnormal{y}}}
\def\ervz{{\textnormal{z}}}

\def\rmA{{\mathbf{A}}}
\def\rmB{{\mathbf{B}}}
\def\rmC{{\mathbf{C}}}
\def\rmD{{\mathbf{D}}}
\def\rmE{{\mathbf{E}}}
\def\rmF{{\mathbf{F}}}
\def\rmG{{\mathbf{G}}}
\def\rmH{{\mathbf{H}}}
\def\rmI{{\mathbf{I}}}
\def\rmJ{{\mathbf{J}}}
\def\rmK{{\mathbf{K}}}
\def\rmL{{\mathbf{L}}}
\def\rmM{{\mathbf{M}}}
\def\rmN{{\mathbf{N}}}
\def\rmO{{\mathbf{O}}}
\def\rmP{{\mathbf{P}}}
\def\rmQ{{\mathbf{Q}}}
\def\rmR{{\mathbf{R}}}
\def\rmS{{\mathbf{S}}}
\def\rmT{{\mathbf{T}}}
\def\rmU{{\mathbf{U}}}
\def\rmV{{\mathbf{V}}}
\def\rmW{{\mathbf{W}}}
\def\rmX{{\mathbf{X}}}
\def\rmY{{\mathbf{Y}}}
\def\rmZ{{\mathbf{Z}}}

\def\ermA{{\textnormal{A}}}
\def\ermB{{\textnormal{B}}}
\def\ermC{{\textnormal{C}}}
\def\ermD{{\textnormal{D}}}
\def\ermE{{\textnormal{E}}}
\def\ermF{{\textnormal{F}}}
\def\ermG{{\textnormal{G}}}
\def\ermH{{\textnormal{H}}}
\def\ermI{{\textnormal{I}}}
\def\ermJ{{\textnormal{J}}}
\def\ermK{{\textnormal{K}}}
\def\ermL{{\textnormal{L}}}
\def\ermM{{\textnormal{M}}}
\def\ermN{{\textnormal{N}}}
\def\ermO{{\textnormal{O}}}
\def\ermP{{\textnormal{P}}}
\def\ermQ{{\textnormal{Q}}}
\def\ermR{{\textnormal{R}}}
\def\ermS{{\textnormal{S}}}
\def\ermT{{\textnormal{T}}}
\def\ermU{{\textnormal{U}}}
\def\ermV{{\textnormal{V}}}
\def\ermW{{\textnormal{W}}}
\def\ermX{{\textnormal{X}}}
\def\ermY{{\textnormal{Y}}}
\def\ermZ{{\textnormal{Z}}}

\def\vzero{{\bm{0}}}
\def\vone{{\bm{1}}}
\def\vmu{{\bm{\mu}}}
\def\vtheta{{\bm{\theta}}}
\def\va{{\bm{a}}}
\def\vb{{\bm{b}}}
\def\vc{{\bm{c}}}
\def\vd{{\bm{d}}}
\def\ve{{\bm{e}}}
\def\vf{{\bm{f}}}
\def\vg{{\bm{g}}}
\def\vh{{\bm{h}}}
\def\vi{{\bm{i}}}
\def\vj{{\bm{j}}}
\def\vk{{\bm{k}}}
\def\vl{{\bm{l}}}
\def\vm{{\bm{m}}}
\def\vn{{\bm{n}}}
\def\vo{{\bm{o}}}
\def\vp{{\bm{p}}}
\def\vq{{\bm{q}}}
\def\vr{{\bm{r}}}
\def\vs{{\bm{s}}}
\def\vt{{\bm{t}}}
\def\vu{{\bm{u}}}
\def\vv{{\bm{v}}}
\def\vw{{\bm{w}}}
\def\vx{{\bm{x}}}
\def\vy{{\bm{y}}}
\def\vz{{\bm{z}}}

\def\evalpha{{\alpha}}
\def\evbeta{{\beta}}
\def\evepsilon{{\epsilon}}
\def\evlambda{{\lambda}}
\def\evomega{{\omega}}
\def\evmu{{\mu}}
\def\evpsi{{\psi}}
\def\evsigma{{\sigma}}
\def\evtheta{{\theta}}
\def\eva{{a}}
\def\evb{{b}}
\def\evc{{c}}
\def\evd{{d}}
\def\eve{{e}}
\def\evf{{f}}
\def\evg{{g}}
\def\evh{{h}}
\def\evi{{i}}
\def\evj{{j}}
\def\evk{{k}}
\def\evl{{l}}
\def\evm{{m}}
\def\evn{{n}}
\def\evo{{o}}
\def\evp{{p}}
\def\evq{{q}}
\def\evr{{r}}
\def\evs{{s}}
\def\evt{{t}}
\def\evu{{u}}
\def\evv{{v}}
\def\evw{{w}}
\def\evx{{x}}
\def\evy{{y}}
\def\evz{{z}}

\def\mA{{\bm{A}}}
\def\mB{{\bm{B}}}
\def\mC{{\bm{C}}}
\def\mD{{\bm{D}}}
\def\mE{{\bm{E}}}
\def\mF{{\bm{F}}}
\def\mG{{\bm{G}}}
\def\mH{{\bm{H}}}
\def\mI{{\bm{I}}}
\def\mJ{{\bm{J}}}
\def\mK{{\bm{K}}}
\def\mL{{\bm{L}}}
\def\mM{{\bm{M}}}
\def\mN{{\bm{N}}}
\def\mO{{\bm{O}}}
\def\mP{{\bm{P}}}
\def\mQ{{\bm{Q}}}
\def\mR{{\bm{R}}}
\def\mS{{\bm{S}}}
\def\mT{{\bm{T}}}
\def\mU{{\bm{U}}}
\def\mV{{\bm{V}}}
\def\mW{{\bm{W}}}
\def\mX{{\bm{X}}}
\def\mY{{\bm{Y}}}
\def\mZ{{\bm{Z}}}
\def\mBeta{{\bm{\beta}}}
\def\mPhi{{\bm{\Phi}}}
\def\mLambda{{\bm{\Lambda}}}
\def\mSigma{{\bm{\Sigma}}}

\newcommand{\tens}[1]{\bm{\mathsfit{#1}}}
\def\tA{{\tens{A}}}
\def\tB{{\tens{B}}}
\def\tC{{\tens{C}}}
\def\tD{{\tens{D}}}
\def\tE{{\tens{E}}}
\def\tF{{\tens{F}}}
\def\tG{{\tens{G}}}
\def\tH{{\tens{H}}}
\def\tI{{\tens{I}}}
\def\tJ{{\tens{J}}}
\def\tK{{\tens{K}}}
\def\tL{{\tens{L}}}
\def\tM{{\tens{M}}}
\def\tN{{\tens{N}}}
\def\tO{{\tens{O}}}
\def\tP{{\tens{P}}}
\def\tQ{{\tens{Q}}}
\def\tR{{\tens{R}}}
\def\tS{{\tens{S}}}
\def\tT{{\tens{T}}}
\def\tU{{\tens{U}}}
\def\tV{{\tens{V}}}
\def\tW{{\tens{W}}}
\def\tX{{\tens{X}}}
\def\tY{{\tens{Y}}}
\def\tZ{{\tens{Z}}}

\def\gA{{\mathcal{A}}}
\def\gB{{\mathcal{B}}}
\def\gC{{\mathcal{C}}}
\def\gD{{\mathcal{D}}}
\def\gE{{\mathcal{E}}}
\def\gF{{\mathcal{F}}}
\def\gG{{\mathcal{G}}}
\def\gH{{\mathcal{H}}}
\def\gI{{\mathcal{I}}}
\def\gJ{{\mathcal{J}}}
\def\gK{{\mathcal{K}}}
\def\gL{{\mathcal{L}}}
\def\gM{{\mathcal{M}}}
\def\gN{{\mathcal{N}}}
\def\gO{{\mathcal{O}}}
\def\gP{{\mathcal{P}}}
\def\gQ{{\mathcal{Q}}}
\def\gR{{\mathcal{R}}}
\def\gS{{\mathcal{S}}}
\def\gT{{\mathcal{T}}}
\def\gU{{\mathcal{U}}}
\def\gV{{\mathcal{V}}}
\def\gW{{\mathcal{W}}}
\def\gX{{\mathcal{X}}}
\def\gY{{\mathcal{Y}}}
\def\gZ{{\mathcal{Z}}}

\def\sA{{\mathbb{A}}}
\def\sB{{\mathbb{B}}}
\def\sC{{\mathbb{C}}}
\def\sD{{\mathbb{D}}}
\def\sF{{\mathbb{F}}}
\def\sG{{\mathbb{G}}}
\def\sH{{\mathbb{H}}}
\def\sI{{\mathbb{I}}}
\def\sJ{{\mathbb{J}}}
\def\sK{{\mathbb{K}}}
\def\sL{{\mathbb{L}}}
\def\sM{{\mathbb{M}}}
\def\sN{{\mathbb{N}}}
\def\sO{{\mathbb{O}}}
\def\sP{{\mathbb{P}}}
\def\sQ{{\mathbb{Q}}}
\def\sR{{\mathbb{R}}}
\def\sS{{\mathbb{S}}}
\def\sT{{\mathbb{T}}}
\def\sU{{\mathbb{U}}}
\def\sV{{\mathbb{V}}}
\def\sW{{\mathbb{W}}}
\def\sX{{\mathbb{X}}}
\def\sY{{\mathbb{Y}}}
\def\sZ{{\mathbb{Z}}}

\def\emLambda{{\Lambda}}
\def\emA{{A}}
\def\emB{{B}}
\def\emC{{C}}
\def\emD{{D}}
\def\emE{{E}}
\def\emF{{F}}
\def\emG{{G}}
\def\emH{{H}}
\def\emI{{I}}
\def\emJ{{J}}
\def\emK{{K}}
\def\emL{{L}}
\def\emM{{M}}
\def\emN{{N}}
\def\emO{{O}}
\def\emP{{P}}
\def\emQ{{Q}}
\def\emR{{R}}
\def\emS{{S}}
\def\emT{{T}}
\def\emU{{U}}
\def\emV{{V}}
\def\emW{{W}}
\def\emX{{X}}
\def\emY{{Y}}
\def\emZ{{Z}}
\def\emSigma{{\Sigma}}

\newcommand{\etens}[1]{\mathsfit{#1}}
\def\etLambda{{\etens{\Lambda}}}
\def\etA{{\etens{A}}}
\def\etB{{\etens{B}}}
\def\etC{{\etens{C}}}
\def\etD{{\etens{D}}}
\def\etE{{\etens{E}}}
\def\etF{{\etens{F}}}
\def\etG{{\etens{G}}}
\def\etH{{\etens{H}}}
\def\etI{{\etens{I}}}
\def\etJ{{\etens{J}}}
\def\etK{{\etens{K}}}
\def\etL{{\etens{L}}}
\def\etM{{\etens{M}}}
\def\etN{{\etens{N}}}
\def\etO{{\etens{O}}}
\def\etP{{\etens{P}}}
\def\etQ{{\etens{Q}}}
\def\etR{{\etens{R}}}
\def\etS{{\etens{S}}}
\def\etT{{\etens{T}}}
\def\etU{{\etens{U}}}
\def\etV{{\etens{V}}}
\def\etW{{\etens{W}}}
\def\etX{{\etens{X}}}
\def\etY{{\etens{Y}}}
\def\etZ{{\etens{Z}}}

\newcommand{\pdata}{p_{\rm{data}}}
\newcommand{\ptrain}{\hat{p}_{\rm{data}}}
\newcommand{\Ptrain}{\hat{P}_{\rm{data}}}
\newcommand{\pmodel}{p_{\rm{model}}}
\newcommand{\Pmodel}{P_{\rm{model}}}
\newcommand{\ptildemodel}{\tilde{p}_{\rm{model}}}
\newcommand{\pencode}{p_{\rm{encoder}}}
\newcommand{\pdecode}{p_{\rm{decoder}}}
\newcommand{\precons}{p_{\rm{reconstruct}}}

\newcommand{\laplace}{\mathrm{Laplace}} %

\newcommand{\E}{\mathbb{E}}
\newcommand{\Ls}{\mathcal{L}}
\newcommand{\R}{\mathbb{R}}
\newcommand{\emp}{\tilde{p}}
\newcommand{\lr}{\alpha}
\newcommand{\reg}{\lambda}
\newcommand{\rect}{\mathrm{rectifier}}
\newcommand{\softmax}{\mathrm{softmax}}
\newcommand{\sigmoid}{\sigma}
\newcommand{\softplus}{\zeta}
\newcommand{\Var}{\mathrm{Var}}
\newcommand{\standarderror}{\mathrm{SE}}
\newcommand{\Cov}{\mathrm{Cov}}
\newcommand{\normlzero}{L^0}
\newcommand{\normlone}{L^1}
\newcommand{\normltwo}{L^2}
\newcommand{\normlp}{L^p}
\newcommand{\normmax}{L^\infty}
\def\abs#1{\left| #1 \right|}

\newcommand{\parents}{Pa} %

\newcommand{\inner}[2]{\langle #1, #2 \rangle}
\newcommand{\norm}[1]{\left\|#1\right\|}

\newcommand{\IN}{\text{in}}
\newcommand{\OUT}{\text{out}}
\newcommand{\conv}{\text{conv}}
\newcommand{\pad}{\text{pad}}
\newcommand{\pool}{\text{pool}}
\newcommand{\patch}{\text{patch}}
\newcommand{\nc}{\mathcal{c}}

\let\ab\allowbreak
\newtheorem{claim}[theorem]{Claim}
\newtheorem{condition}{Condition}

\def\lv{\lVert}
\def\rv{\rVert}

\newcommand{\calA}{\mathcal{A}}
\newcommand{\calB}{\mathcal{B}}
\newcommand{\calC}{\mathcal{C}}
\newcommand{\calD}{\mathcal{D}}
\newcommand{\calE}{\mathcal{E}}
\newcommand{\calF}{\mathcal{F}}
\newcommand{\calG}{\mathcal{G}}
\newcommand{\calH}{\mathcal{H}}
\newcommand{\calI}{\mathcal{I}}
\newcommand{\calJ}{\mathcal{J}}
\newcommand{\calK}{\mathcal{K}}
\newcommand{\calL}{\mathcal{L}}
\newcommand{\calM}{\mathcal{M}}
\newcommand{\calN}{\mathcal{N}}
\newcommand{\calO}{\mathcal{O}}
\newcommand{\calP}{\mathcal{P}}
\newcommand{\calQ}{\mathcal{Q}}
\newcommand{\calR}{\mathcal{R}}
\newcommand{\calS}{\mathcal{S}}
\newcommand{\calT}{\mathcal{T}}
\newcommand{\calU}{\mathcal{U}}
\newcommand{\calV}{\mathcal{V}}
\newcommand{\calW}{\mathcal{W}}
\newcommand{\calX}{\mathcal{X}}
\newcommand{\calY}{\mathcal{Y}}
\newcommand{\calZ}{\mathcal{Z}}

\newcommand{\mathA}{\mathbb{A}}
\newcommand{\mathB}{\mathbb{B}}
\newcommand{\mathC}{\mathbb{C}}
\newcommand{\mathD}{\mathbb{D}}
\newcommand{\mathE}{\mathbb{E}}
\newcommand{\mathF}{\mathbb{F}}
\newcommand{\mathG}{\mathbb{G}}
\newcommand{\mathH}{\mathbb{H}}
\newcommand{\mathI}{\mathbb{I}}
\newcommand{\mathJ}{\mathbb{J}}
\newcommand{\mathK}{\mathbb{K}}
\newcommand{\mathL}{\mathbb{L}}
\newcommand{\mathM}{\mathbb{M}}
\newcommand{\mathN}{\mathbb{N}}
\newcommand{\mathO}{\mathbb{O}}
\newcommand{\mathP}{\mathbb{P}}
\newcommand{\mathQ}{\mathbb{Q}}
\newcommand{\mathR}{\mathbb{R}}
\newcommand{\mathS}{\mathbb{S}}
\newcommand{\mathT}{\mathbb{T}}
\newcommand{\mathU}{\mathbb{U}}
\newcommand{\mathV}{\mathbb{V}}
\newcommand{\mathW}{\mathbb{W}}
\newcommand{\mathX}{\mathbb{X}}
\newcommand{\mathY}{\mathbb{Y}}
\newcommand{\mathZ}{\mathbb{Z}}

\newcommand{\vecA}{\mathbf{A}}
\newcommand{\vecB}{\mathbf{B}}
\newcommand{\vecC}{\mathbf{C}}
\newcommand{\vecD}{\mathbf{D}}
\newcommand{\vecE}{\mathbf{E}}
\newcommand{\vecF}{\mathbf{F}}
\newcommand{\vecG}{\mathbf{G}}
\newcommand{\vecH}{\mathbf{H}}
\newcommand{\vecI}{\mathbf{I}}
\newcommand{\vecJ}{\mathbf{J}}
\newcommand{\vecK}{\mathbf{K}}
\newcommand{\vecL}{\mathbf{L}}
\newcommand{\vecM}{\mathbf{M}}
\newcommand{\vecN}{\mathbf{N}}
\newcommand{\vecO}{\mathbf{O}}
\newcommand{\vecP}{\mathbf{P}}
\newcommand{\vecQ}{\mathbf{Q}}
\newcommand{\vecR}{\mathbf{R}}
\newcommand{\vecS}{\mathbf{S}}
\newcommand{\vecT}{\mathbf{T}}
\newcommand{\vecU}{\mathbf{U}}
\newcommand{\vecV}{\mathbf{V}}
\newcommand{\vecW}{\mathbf{W}}
\newcommand{\vecX}{\mathbf{X}}
\newcommand{\vecY}{\mathbf{Y}}
\newcommand{\vecZ}{\mathbf{Z}}
\newcommand{\vecAlpha}{\boldsymbol{\alpha}}
\newcommand{\vecBeta}{\boldsymbol{\beta}}
\newcommand{\vecGamma}{\boldsymbol{\gamma}}
\newcommand{\vecLambda}{\boldsymbol{\lambda}}
\newcommand{\vecDelta}{\boldsymbol{\Delta}}
\newcommand{\vecdelta}{\boldsymbol{\delta}}
\newcommand{\vecphi}{\boldsymbol{\phi}}
\newcommand{\vectheta}{\boldsymbol{\theta}}

\newcommand{\veca}{\mathbf{a}}
\newcommand{\vecb}{\mathbf{b}}
\newcommand{\vecc}{\mathbf{c}}
\newcommand{\vecd}{\mathbf{d}}
\newcommand{\vece}{\mathbf{e}}
\newcommand{\vecf}{\mathbf{f}}
\newcommand{\vecg}{\mathbf{g}}
\newcommand{\vech}{\mathbf{h}}
\newcommand{\veci}{\mathbf{i}}
\newcommand{\vecj}{\mathbf{j}}
\newcommand{\veck}{\mathbf{k}}
\newcommand{\vecl}{\mathbf{l}}
\newcommand{\vecm}{\mathbf{m}}
\newcommand{\vecn}{\mathbf{n}}
\newcommand{\veco}{\mathbf{o}}
\newcommand{\vecp}{\mathbf{p}}
\newcommand{\vecq}{\mathbf{q}}
\newcommand{\vecr}{\mathbf{r}}
\newcommand{\vecs}{\mathbf{s}}
\newcommand{\vect}{\mathbf{t}}
\newcommand{\vecu}{\mathbf{u}}
\newcommand{\vecv}{\mathbf{v}}
\newcommand{\vecw}{\mathbf{w}}
\newcommand{\vecx}{\mathbf{x}}
\newcommand{\vecy}{\mathbf{y}}
\newcommand{\vecz}{\mathbf{z}}

\newcommand{\RR}{\mathbb{R}}

\newcommand{\modelfig}{
\begin{figure}
\includegraphics[width=0.45\textwidth]{img/modelfig}
\caption{\label{fig:modelfig} Our models}
\end{figure}
}

\newcommand{\figloctoscore}{
\begin{figure}\centering
\includegraphics[width=0.45\textwidth]{img/loctoscore}
\caption{\label{fig:loctoscore} CLIP and FID for various location of the
  logit normal sampler and for various sampler settings.}
\end{figure}
}

\newcommand{\fignormalizedvalloc}{
\begin{figure}\centering
\includegraphics[width=0.45\textwidth]{img/normalizedvalloc}
\caption{\label{fig:loctoscore} Normalized validation loss over timestep $t$
  for logit normal sampler with different location parameters.}
\end{figure}
}

\newcommand{\fignormalizedvalscale}{
\begin{figure}\centering
\includegraphics[width=0.45\textwidth]{img/normalizedvalscale}
\caption{\label{fig:loctoscore} Normalized validation loss over timestep $t$
  for logit normal sampler with centered location and different scales.}
\end{figure}
}

\newcommand{\modesampler}{
\begin{figure}
\includegraphics[width=0.495\textwidth]{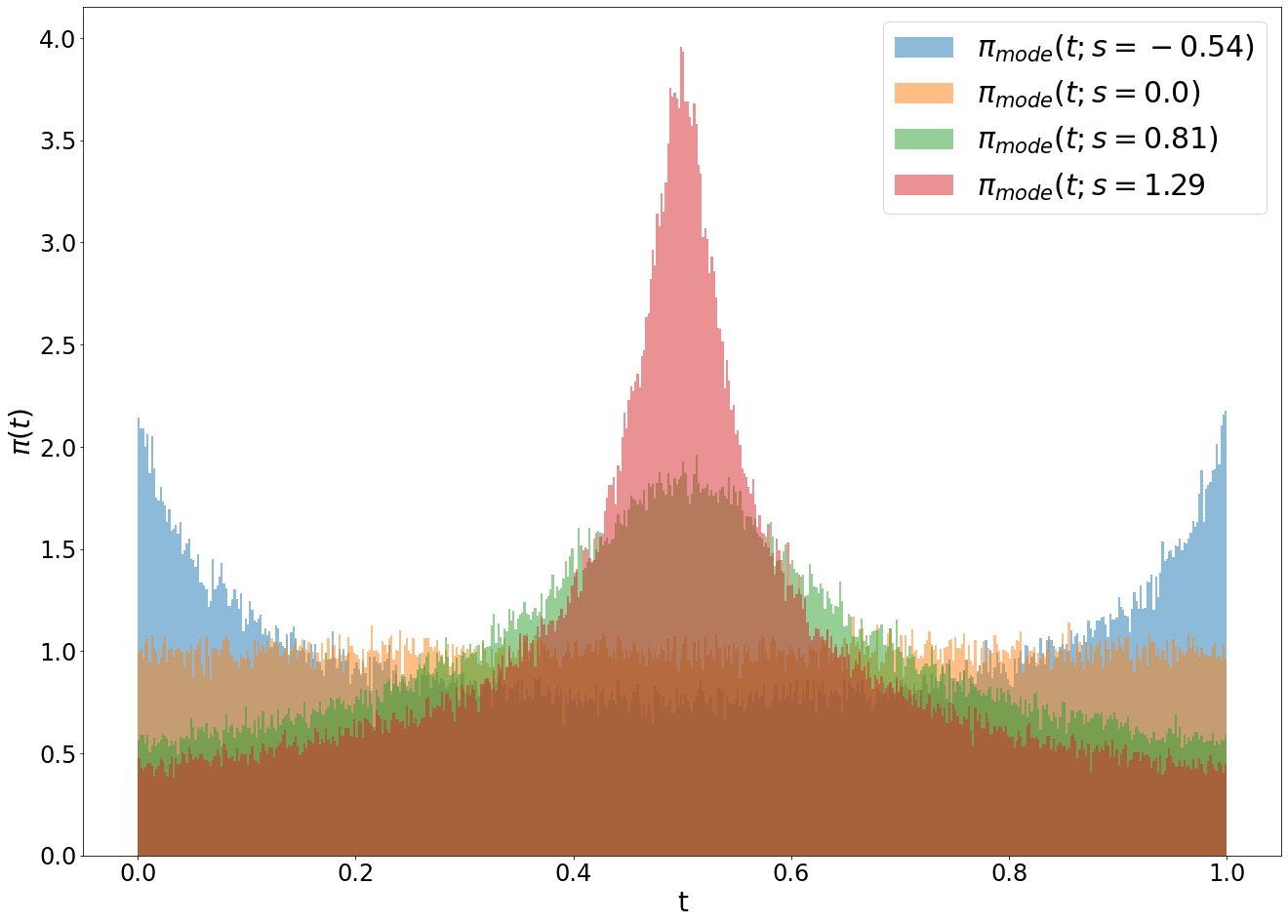}
\caption{\label{fig:modesampler} The mode sampler}
\end{figure}
}

\newcommand{\logitsampler}{
\begin{figure}
\includegraphics[width=0.495\textwidth]{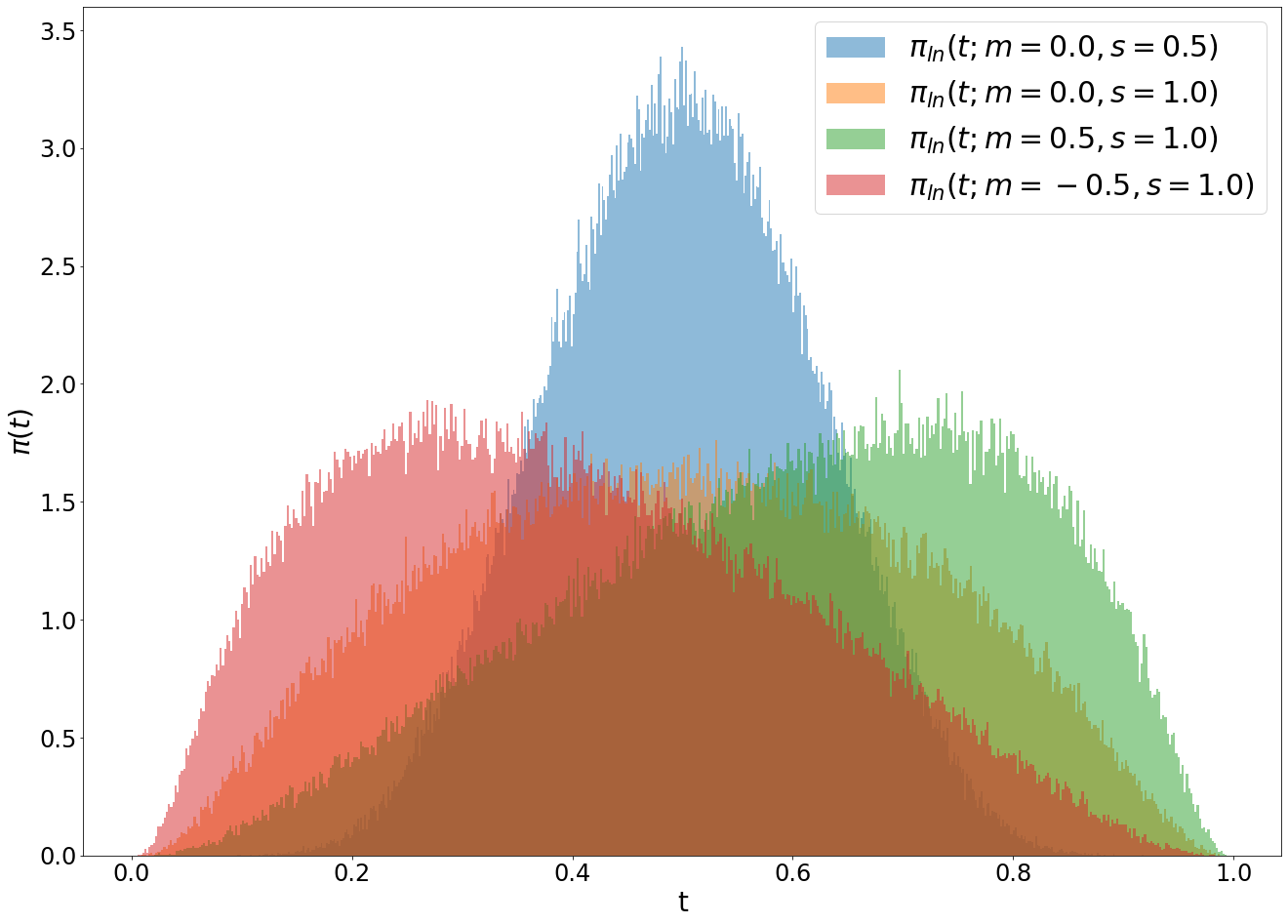}
\caption{\label{fig:logitsampler} The logit-normal sampler}
\end{figure}
}

\newcommand{\archstraining}{
\begin{figure*}[t!]
\includegraphics[width=0.32\textwidth]{img/archs/val_loss_level_6.png}
\includegraphics[width=0.32\textwidth]{img/archs/clip_scores_sampler_default_ema_True.png}
\includegraphics[width=0.32\textwidth]{img/archs/fid_scores_sampler_default_ema_True.png}
\caption{\label{fig:archstraining} 
\textbf{Training dynamics of model architectures.} 
Comparative analysis of  \emph{DiT},  \emph{CrossDiT},  \emph{UViT}, and \modelname on CC12M, focusing on validation loss, CLIP score, and FID. 
Our proposed \modelname performs favorably across all metrics. \vspace{-1em}
}
\end{figure*}
}

\newcommand{\archstrainingsqueezed}{
\setlength{\tabcolsep}{-2pt}
\begin{figure}
\begin{tabular}{cc}
\includegraphics[width=0.5\linewidth,valign=m]{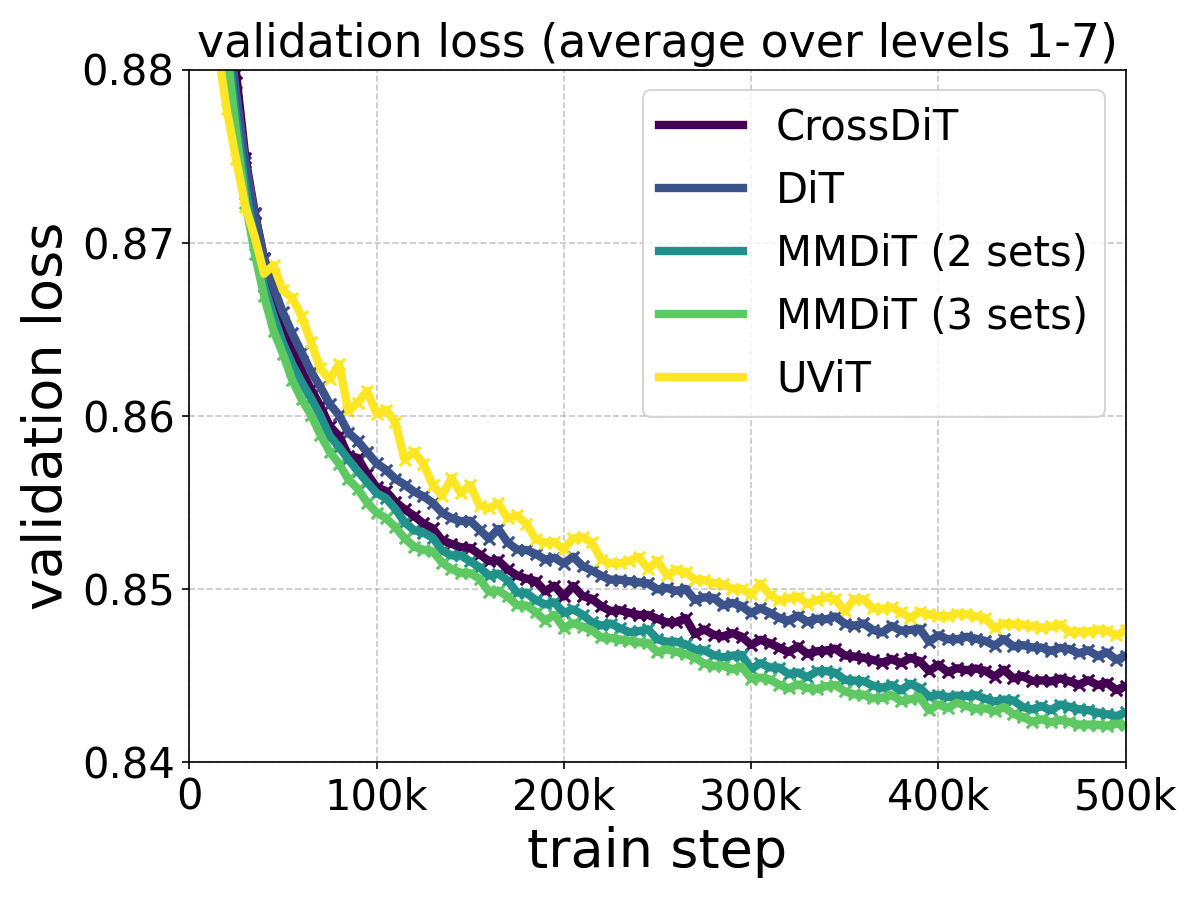} &
\includegraphics[width=0.5\linewidth,valign=m]{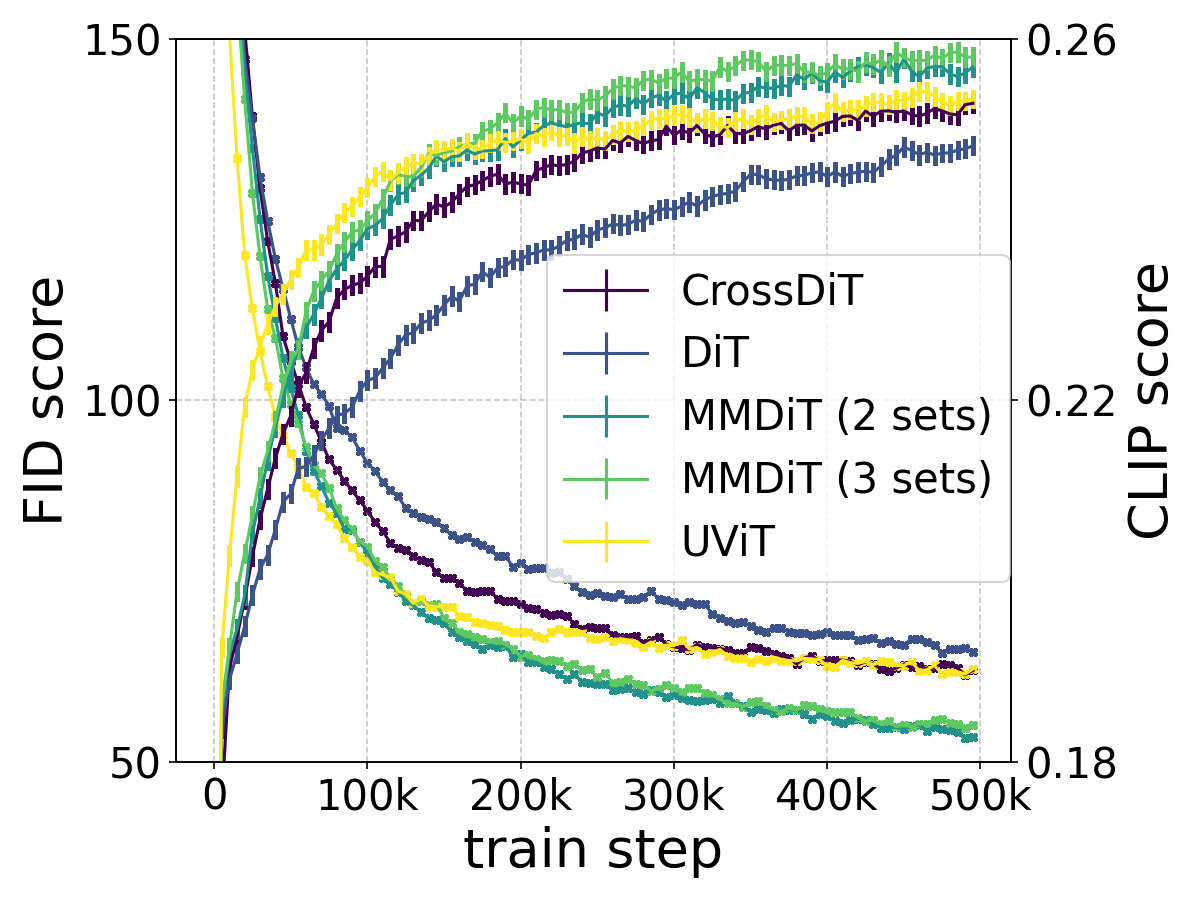}
\end{tabular}
\caption{\label{fig:archstraining}\textbf{Training dynamics of model architectures.} 
Comparative analysis of  \emph{DiT},  \emph{CrossDiT},  \emph{UViT}, and \modelname on CC12M, focusing on validation loss, CLIP score, and FID. 
Our proposed \modelname performs favorably across all metrics.}
\vspace{-1em}
\end{figure}

}

\newcommand{\aefidstudy}{
\begin{figure}\centering
\includegraphics[width=0.45\textwidth]{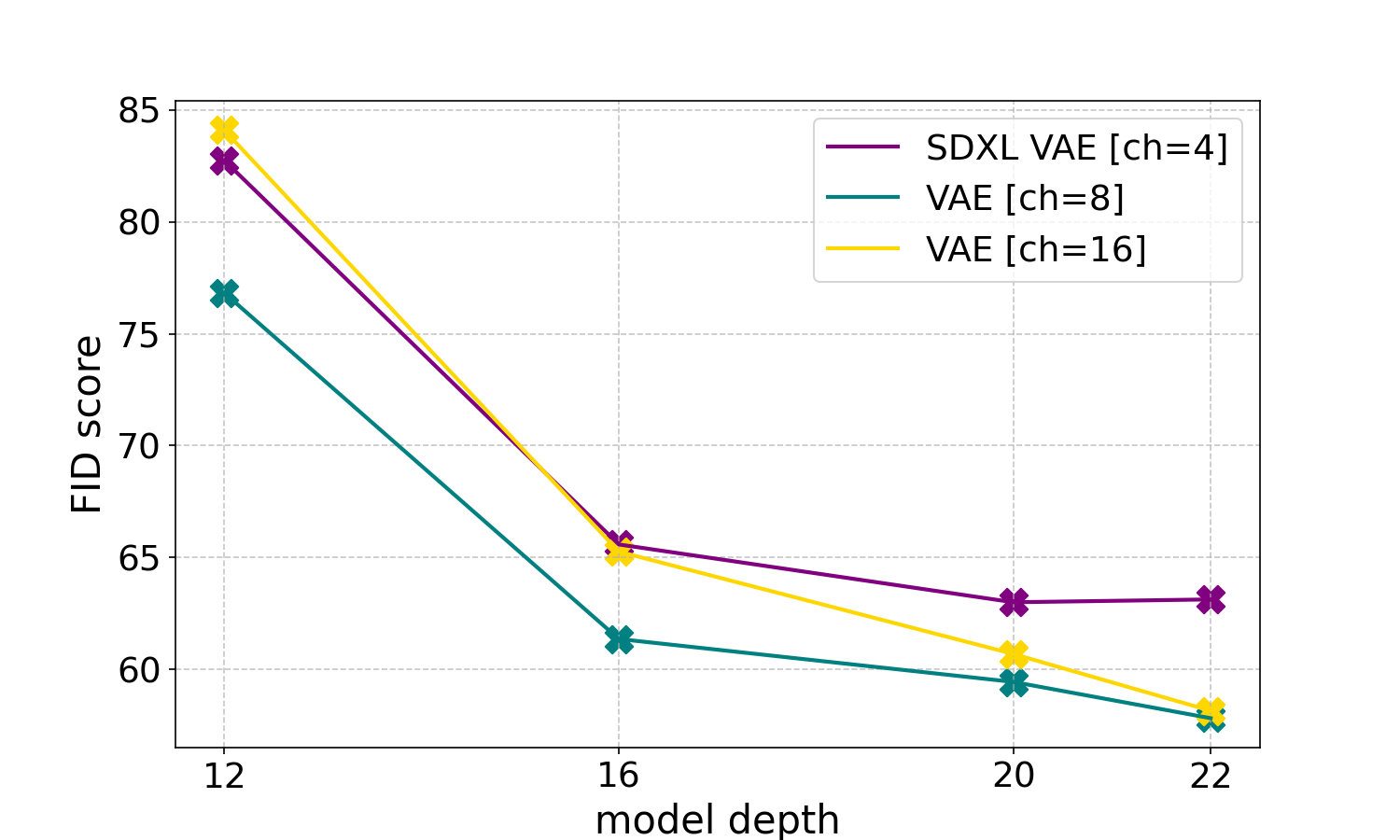}
\caption{\label{fig:aefidstudy} FID scores after training flow models with different sizes (parameterized via their depth) on the latent space of different autoencoders (4 latent channels, 8 channels and 16 channels) as discussed in \Cref{sec:improvedae}. As expected, the flow model trained on the 16-channel autoencoder space needs more model capacity to achieve similar performance. At depth $d=22$, the gap between 8-chn and 16-chn becomes negligible. We opt for the 16-chn model as we ultimately aim to scale to much larger model sizes.}
\end{figure}
}

\newcommand{\tsamplers}{
\begin{figure*}
  \begin{center}%
    \includegraphics[width=0.45\linewidth]{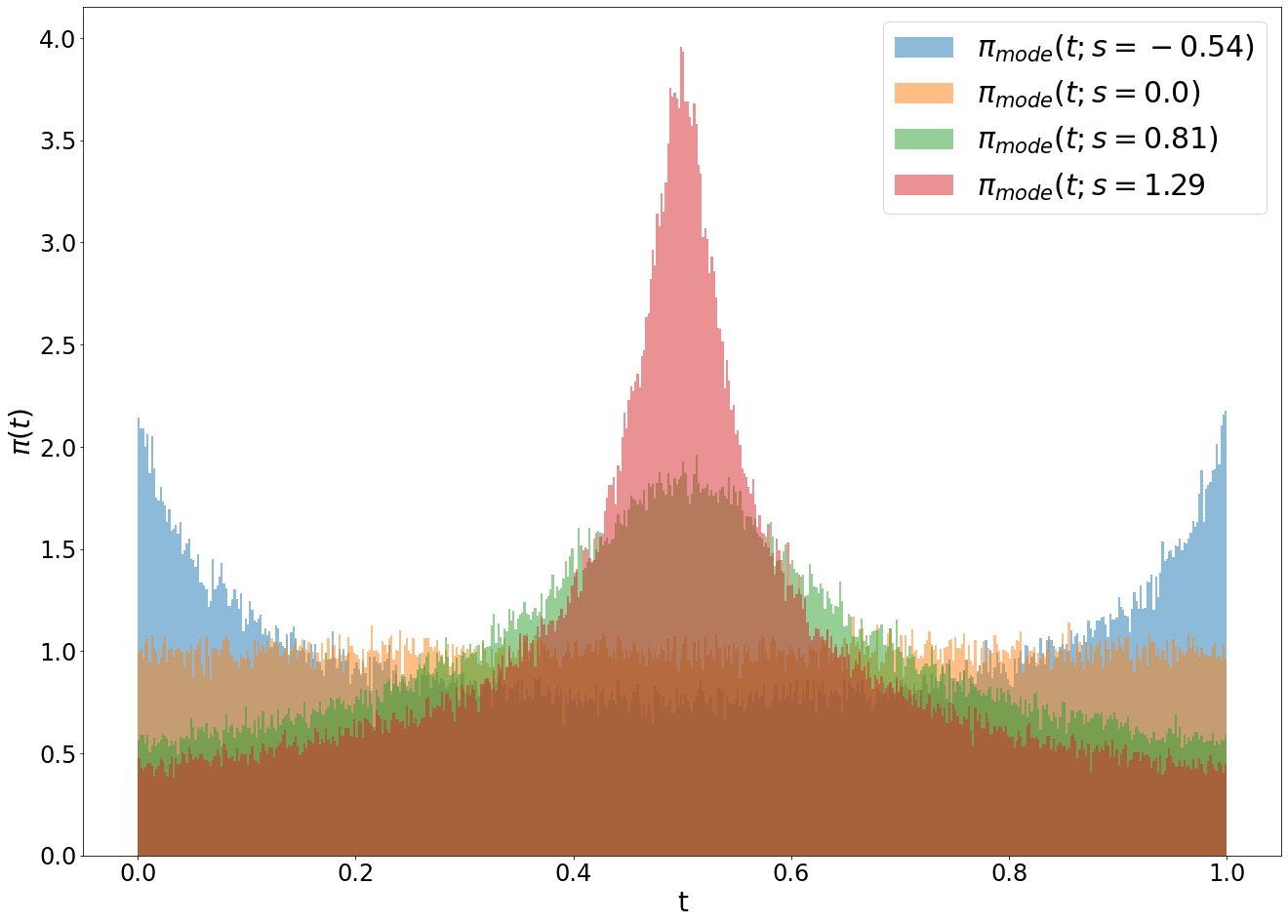}%
    \hfil%
    \includegraphics[width=0.45\linewidth]{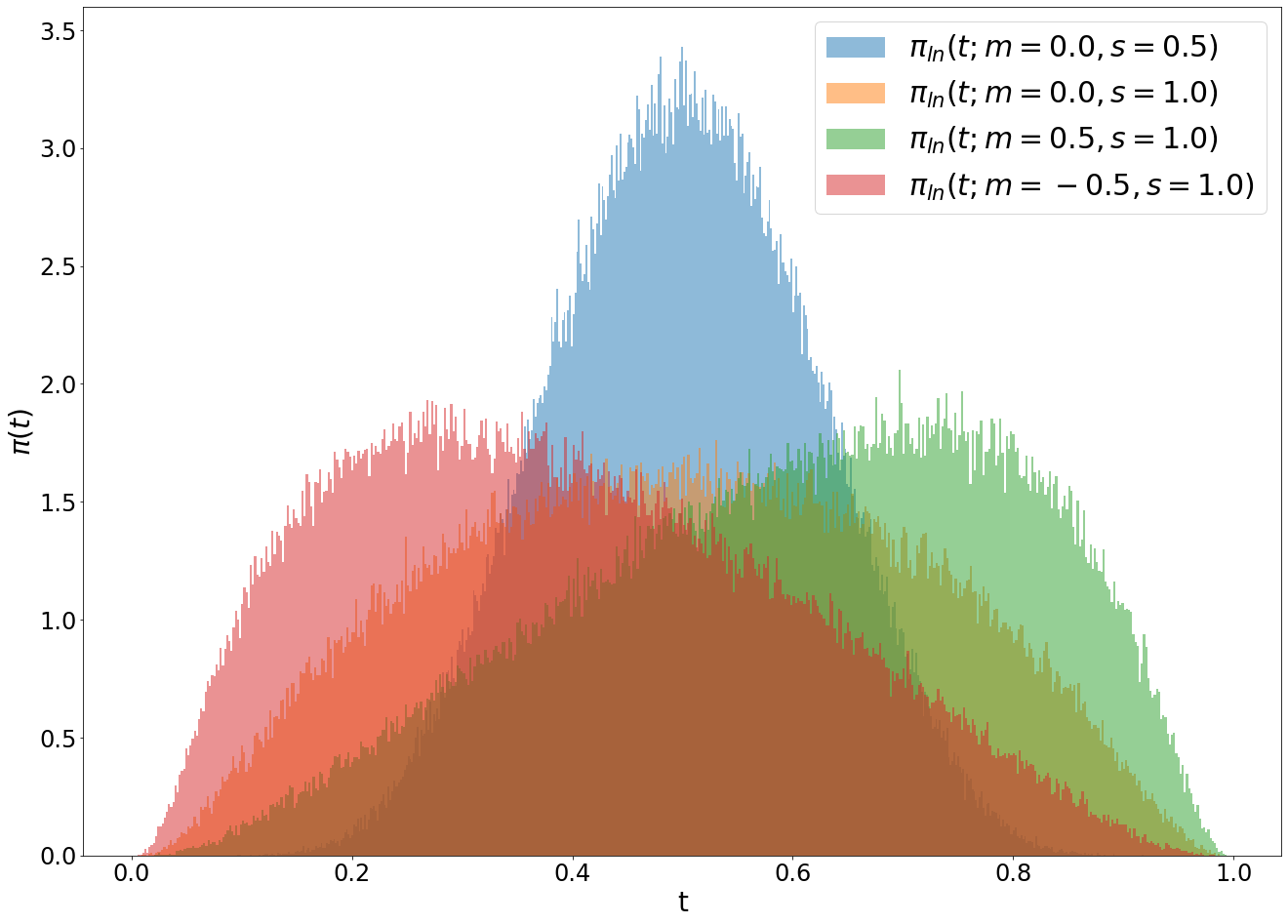}%
  \end{center}
  \caption{The mode (left) and logit-normal (right) distributions that we
  explore for biasing the sampling of training timesteps.}
  \label{fig:tsamplers}
\end{figure*}
}

\newcommand{\dropthememb}{
\begin{figure}[h]
\centering
\setlength{\tabcolsep}{1pt}
\begin{tabular}{cc}
\toprule
 \textit{All text-encoders} & \textit{w/o T5~\citep{raffel2019exploring}}\\
\midrule
 \includegraphics[width=.485\linewidth]{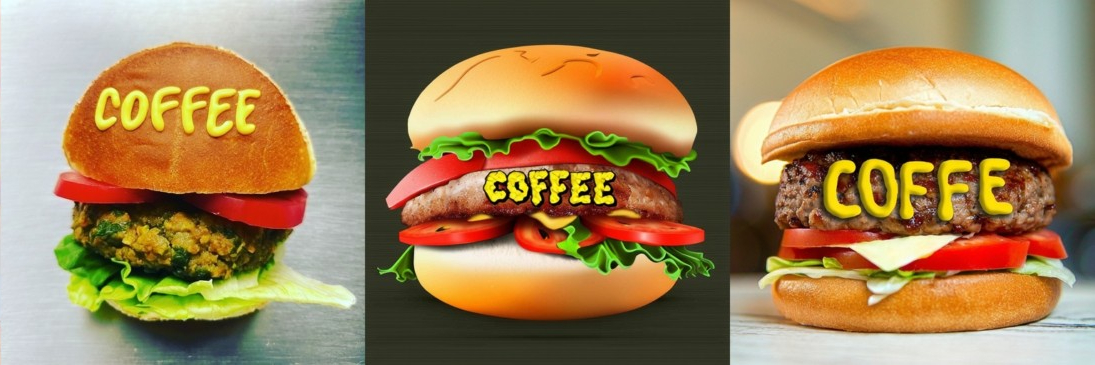}&
 \includegraphics[width=.485\linewidth]{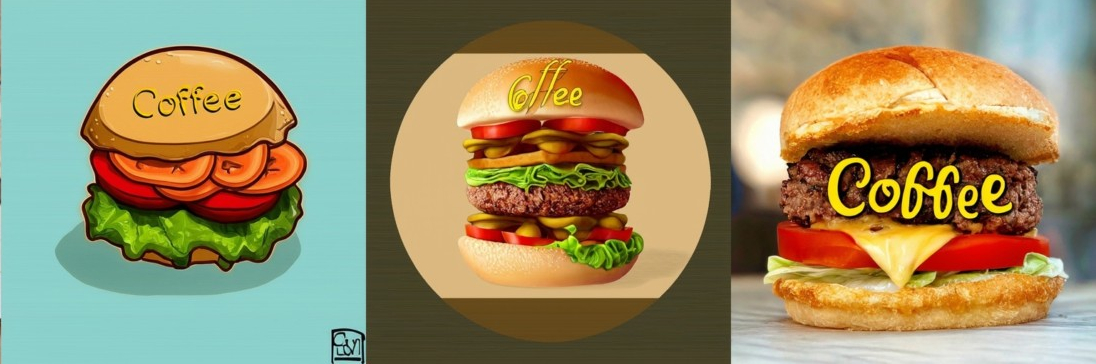} \\[-2pt]
\multicolumn{2}{c}{\tiny``A burger patty, with the bottom bun and lettuce and tomatoes. "COFFEE" written on it in mustard''} \\
 \midrule
 \includegraphics[width=.485\linewidth]{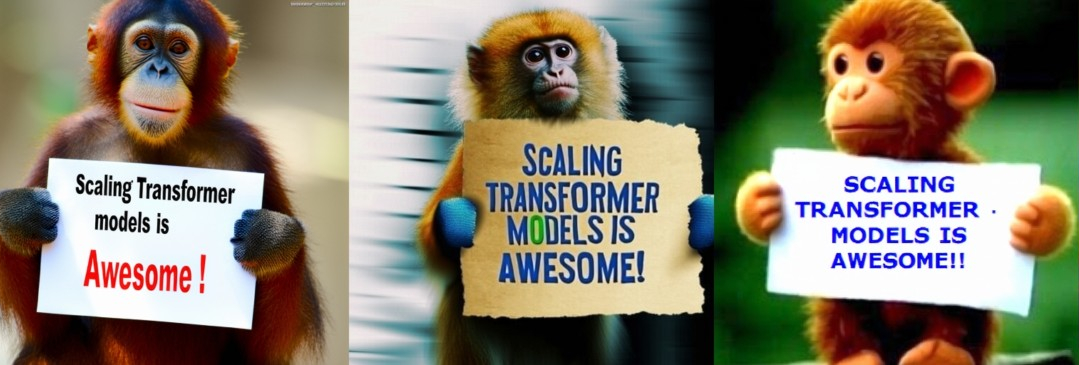}&
 \includegraphics[width=.485\linewidth]{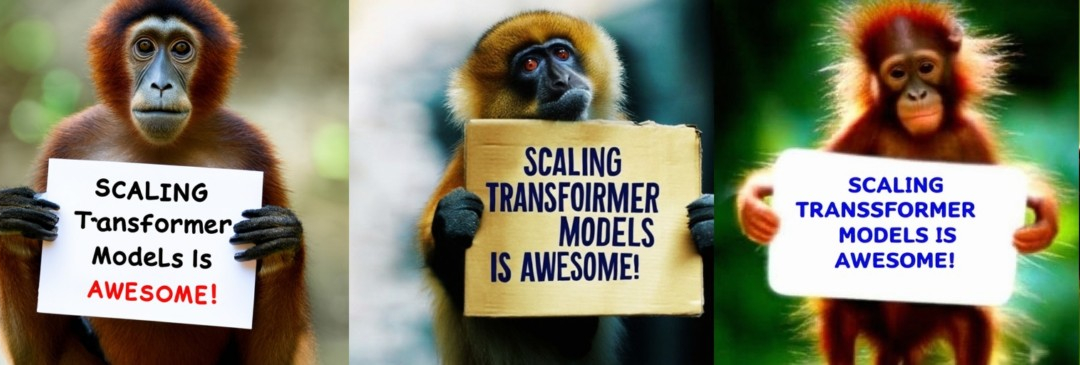} \\[-2pt]
\multicolumn{2}{c}{\tiny``A monkey holding a sign reading "Scaling transformer models is awesome!''} \\
\midrule
\includegraphics[width=.485\linewidth]{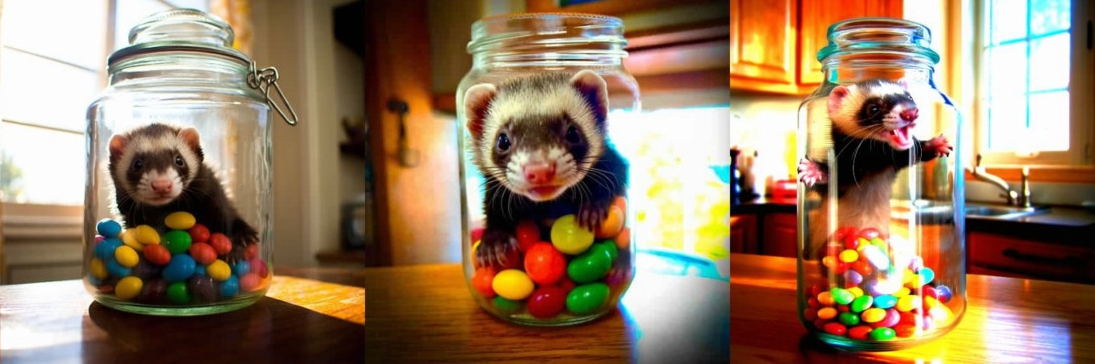}&
\includegraphics[width=.485\linewidth]{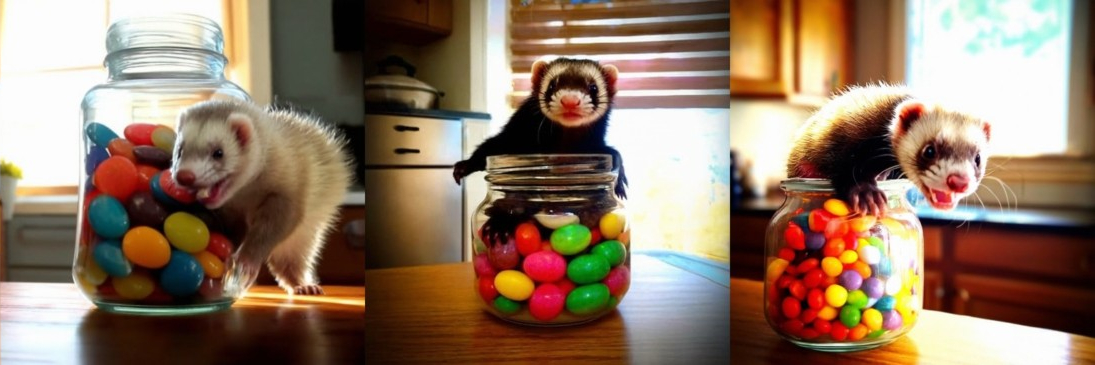} \\ 
\multicolumn{2}{c}{\parbox[b]{.97\linewidth}{\centering\tiny``A mischievous ferret with a playful grin squeezes itself into a
large glass jar, surrounded by colorful candy. The jar sits on
a wooden table in a cozy kitchen, and warm sunlight filters
through a nearby window''}} \\
\bottomrule
  \vspace{-1.75em}
\end{tabular}
\caption{\label{fig:embedder_dropping} \textbf{Impact of T5.} We observe T5 to
  be important for complex prompts e.g. such involving a high degree of detail
  or longer spelled text (rows 2 and 3). For most prompts, however, we find
  that removing T5 at inference time still achieves competitive
  performance.}
\vspace{-1.0em}
\end{figure}
}

\newcommand{\tabglobalrank}{
\begin{table}[h]
\centering
\small
\begin{tabular}{lccc}\toprule
& \multicolumn{3}{c}{rank averaged over} \\ \midrule
variant & all & 5 steps & 50 steps \\ \midrule
rf/lognorm(0.00, 1.00) & \s1.54 & \s1.25 & \s1.50 \\
rf/lognorm(1.00, 0.60) & \s2.08 & \s3.50 & \s2.00 \\
rf/lognorm(0.50, 0.60) & \s2.71 & \s8.50 & \s1.00 \\
rf/mode(1.29) & \s2.75 & \s3.25 & \s3.00 \\
rf/lognorm(0.50, 1.00) & \s2.83 & \s1.50 & \s2.50 \\
eps/linear & \s2.88 & \s4.25 & \s2.75 \\
rf/mode(1.75) & \s3.33 & \s2.75 & \s2.75 \\
rf/cosmap & \s4.13 & \s3.75 & \s4.00 \\
edm(0.00, 0.60) & \s5.63 & 13.25 & \s3.25 \\
rf & \s5.67 & \s6.50 & \s5.75 \\
v/linear & \s6.83 & \s5.75 & \s7.75 \\
edm(0.60, 1.20) & \s9.00 & 13.00 & \s9.00 \\
v/cos & \s9.17 & 12.25 & \s8.75 \\
edm/cos & 11.04 & 14.25 & 11.25 \\
edm/rf & 13.04 & 15.25 & 13.25 \\
edm(-1.20, 1.20) & 15.58 & 20.25 & 15.00 \\
\bottomrule
\end{tabular}
\vspace{-0.75em}
\caption{\label{tab:globalranking}
\textbf{Global ranking of variants.}
For this ranking, we apply non-dominated sorting averaged over EMA and non-EMA weights, two datasets and different sampling settings.}
\end{table}
}

\newcommand{\tablowsteprank}{
\begin{table}[!ht]
    \centering
    \caption{Non-dominated sorting of variants averaged over two sampler
    settings with 5 steps and two datasets.}
    \begin{tabular}{lr}
        variant & average rank \\ \midrule
        rf/lognorm\_0.00\_1.00 & 1.25 \\
        (rf/lognorm\_0.50\_1.00) & 1.50 \\
        rf/mode\_1.75 & 2.75 \\
        (rf/mode\_1.29) & 3.25 \\
        rf/cosmap & 3.75 \\
        eps/linear & 4.25 \\
        v/linear & 5.75 \\
        rf/mode\_0.00 & 6.50 \\
        v/cos & 12.25 \\
        edm/edm\_0.60\_1.20 & 13.00 \\
        edm/cos & 14.25 \\
        edm/edmrf & 15.25 \\
        edm/edm\_-1.20\_1.20 & 20.25 \\
    \end{tabular}
\end{table}
}
\newcommand{\tabhighsteprank}{
  \begin{table}[!ht]
    \centering
    \caption{Non-dominated sorting of variants averaged over two sampler
    settings with 50 steps and two datasets.}
    \begin{tabular}{lr}
        variant & average rank \\ \midrule
        rf/lognorm\_0.50\_0.60 & 1.0 \\
        (rf/lognorm\_0.00\_1.00) & 1.5 \\
        eps/linear & 2.75 \\
        rf/mode\_1.75 & 2.75 \\
        (rf/mode\_1.29) & 3.00 \\
        edm/edm\_0.00\_0.60 & 3.25 \\
        rf/cosmap & 4.00 \\
        rf/mode\_0.00 & 5.75 \\
        v/linear & 7.75 \\
        v/cos & 8.75 \\
        edm/cos & 11.25 \\
        edm/edmrf & 13.25 \\
        edm/edm\_-1.20\_1.20 & 15.00 \\
    \end{tabular}
\end{table}
}

\newcommand{\tabsota}{
\begin{table}[t]
\centering
\begin{tabular}{lcccc} \toprule
& \multicolumn{2}{c}{ImageNet} & \multicolumn{2}{c}{CC12M} \\ 
\cmidrule(lr){2-3} \cmidrule(lr){4-5}
variant & CLIP & FID & CLIP & FID \\ \midrule
rf                     & 0.247	           & \s49.70             & 0.217             & \s94.90 \\
edm(-1.20, 1.20)	   & 0.236	           & \s63.12             & 0.200             & 116.60 \\
eps/linear	           & 0.245	           & \s48.42             & 0.222             & \textit{\s90.34} \\
v/cos	               & 0.244	           & \s50.74             & 0.209             & \s97.87 \\
v/linear	           & 0.246	           & \s51.68             & 0.217             & 100.76 \\ \midrule
rf/lognorm(0.50, 0.60) & \textbf{0.256}	   & \s80.41             & \underline{0.233} & 120.84 \\
rf/mode(1.75)          & \textit{0.253}    & \textbf{\s44.39}    & 0.218             & \s94.06 \\
rf/lognorm(1.00, 0.60) & \underline{0.254} & 114.26            & \textbf{0.234}    & 147.69 \\
rf/lognorm(-0.50, 1.00)& 0.248             & \s\underline{45.64} & 0.219             & \s\textbf{89.70} \\ \midrule
rf/lognorm(0.00, 1.00) & 0.250             & \textit{\s45.78}    & \textit{0.224}    & \s\underline{89.91} \\ \bottomrule
\end{tabular}
\vspace{-0.75em}
\caption{\label{tab:tabsota} 
\textbf{Metrics for different variants.}
FID and CLIP scores of different variants with
25 sampling steps. We highlight the \textbf{best}, \underline{second best},
and \textit{third best} entries.}
\end{table}
}

\newcommand{\figqualitativescaleparti}{
  \begin{figure*}
    \centering
    \small
\begin{tabular}{c@{\hspace{3pt}}c@{\hspace{3pt}}c@{\hspace{3pt}}c}
  \includegraphics[width=.23\linewidth,valign=m]{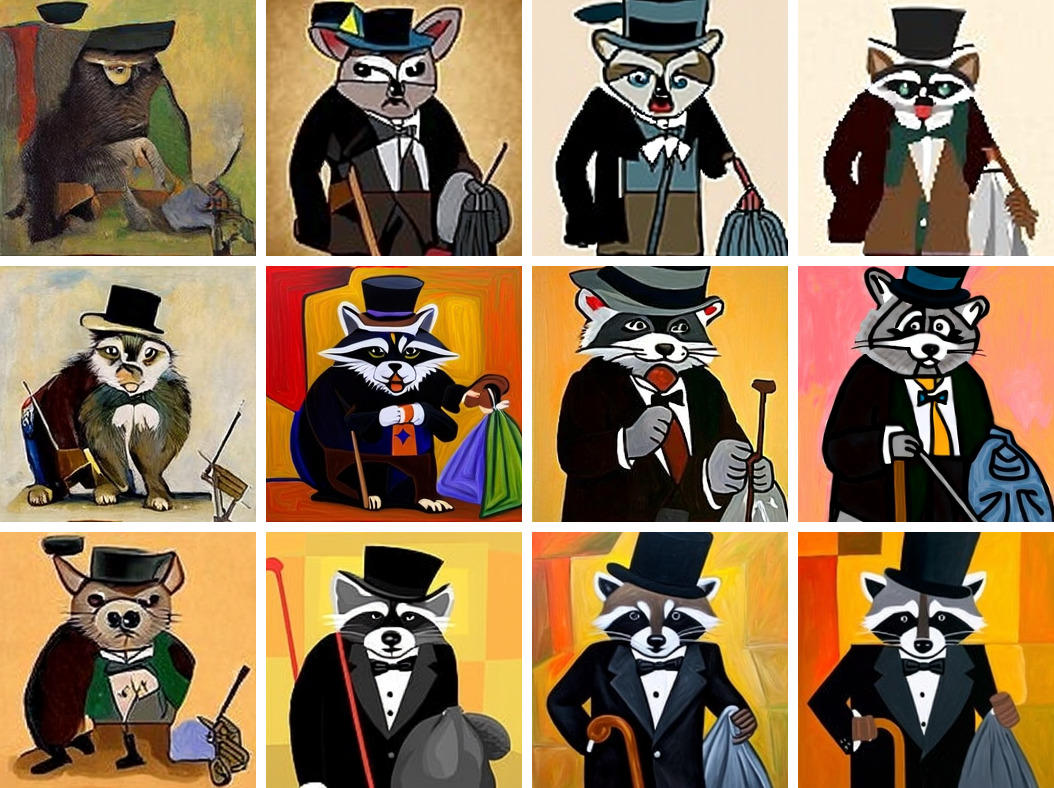}
  &
  \includegraphics[width=.23\linewidth,valign=m]{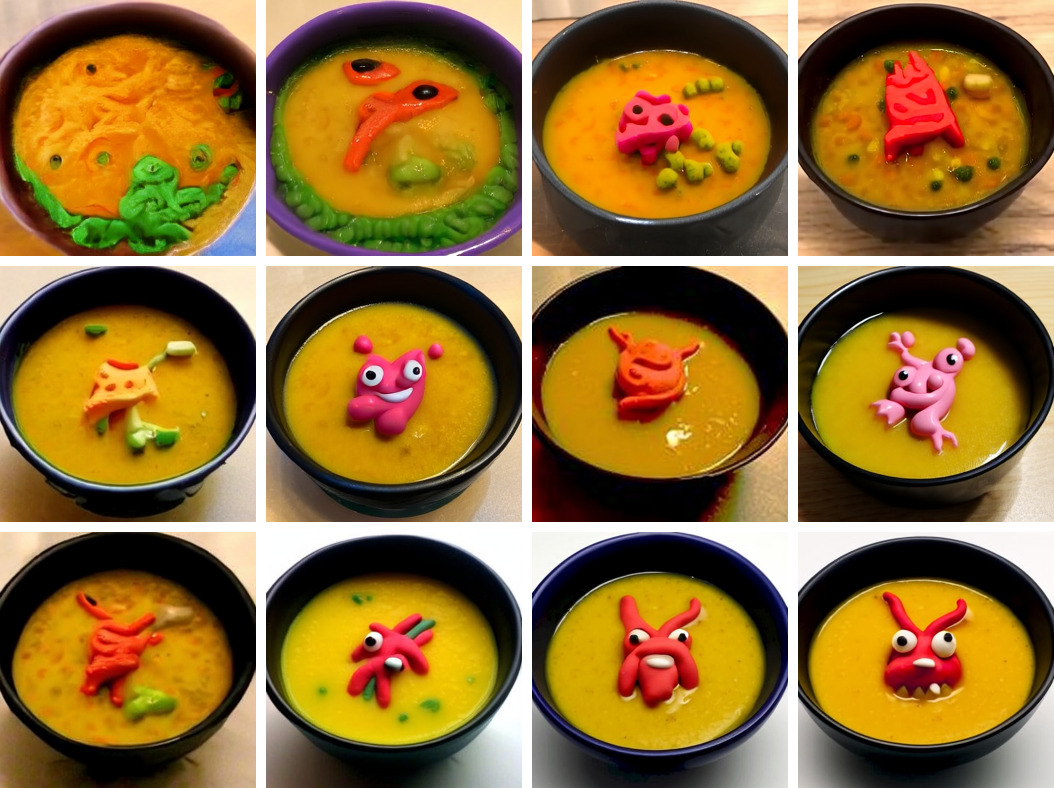}
  &
  \includegraphics[width=.23\linewidth,valign=m]{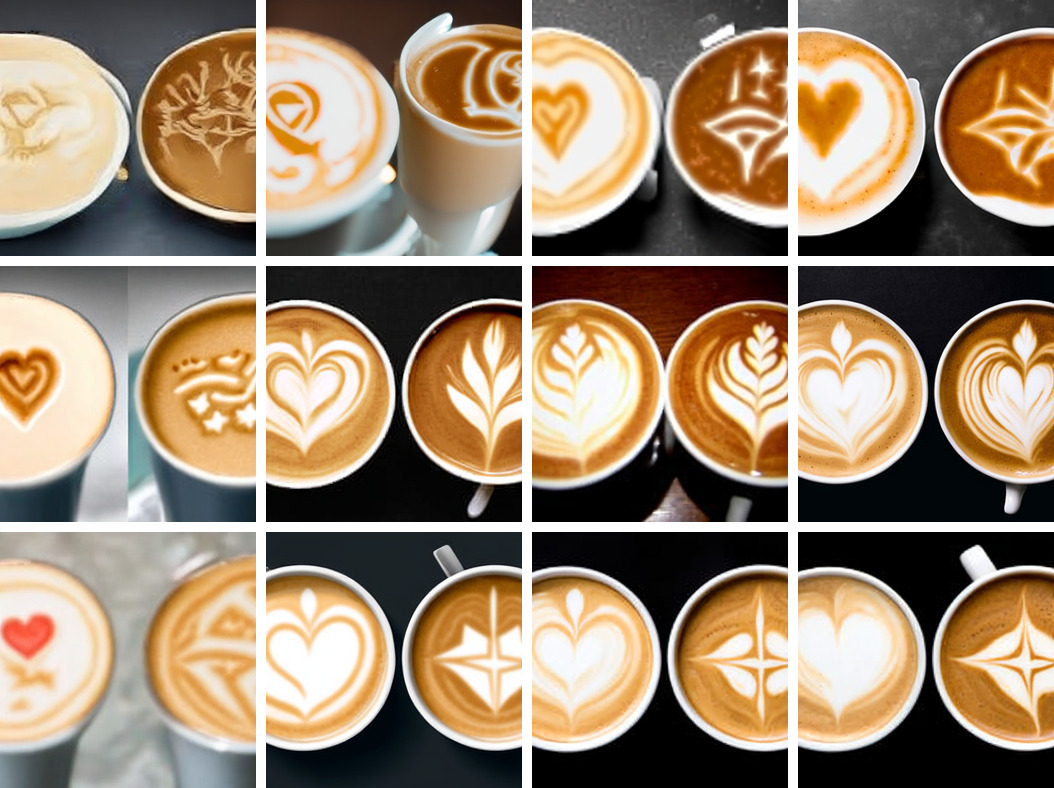}
  &
  \includegraphics[width=.23\linewidth,valign=m]{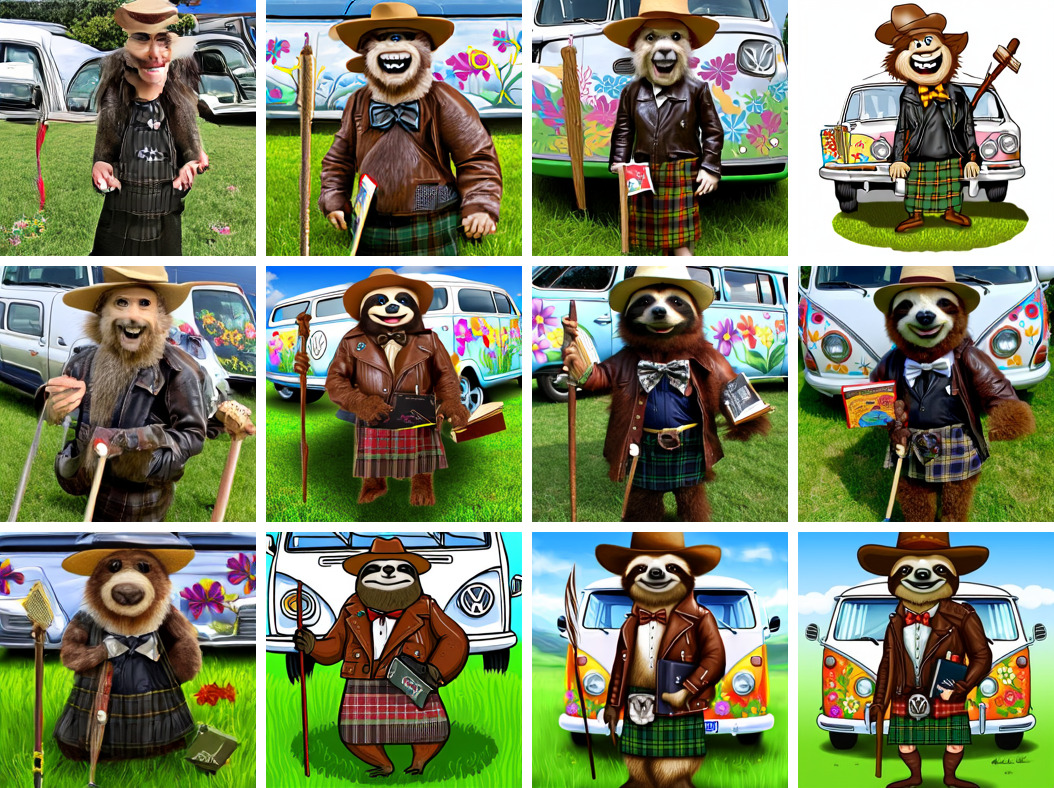}\\
  \multirow{3}{*}{
    \parbox[b]{.23\linewidth}{\centering \tiny ``A raccoon wearing formal clothes, wearing a tophat and holding a cane. The raccoon is holding a garbage bag. Oil painting in the style of abstract cubism.''}
    }
    &
    \multirow{3}{*}{
    \parbox[b]{.23\linewidth}{\centering \tiny ``A bowl of soup that looks like a monster made out of plasticine''}
    }
    &
    \multirow{3}{*}{
    \parbox[b]{.23\linewidth}{\centering \tiny ``Two cups of coffee, one with latte art of a heart. The other has latte art of stars.''}
    }
    &
    \multirow{3}{*}{
    \parbox[b]{.23\linewidth}{\centering \tiny ``A smiling sloth is wearing a leather jacket, a cowboy hat, a kilt and a bowtie. The sloth is holding a quarterstaff and a big book. The sloth is standing on grass a few feet in front of a shiny VW van with flowers painted on it. wide-angle lens from below.''}
    }\\
    &&& \\
    &&& \\
    
\end{tabular}
\caption{
\textbf{Qualitative effects of scaling.} Displayed are examples demonstrating the impact of scaling training steps (left to right: 50k, 200k, 350k, 500k) and model sizes (top to bottom: depth=15, 30, 38) on PartiPrompts, highlighting the influence of training duration and model complexity.
}
\label{fig:qualitativescaleparti}
\end{figure*}
}

\newcommand{\figvalscalingsqueeze}{
\setlength{\tabcolsep}{-1pt}
\begin{figure}
\begin{tabular}{cc}
\includegraphics[width=.5\linewidth,valign=m]{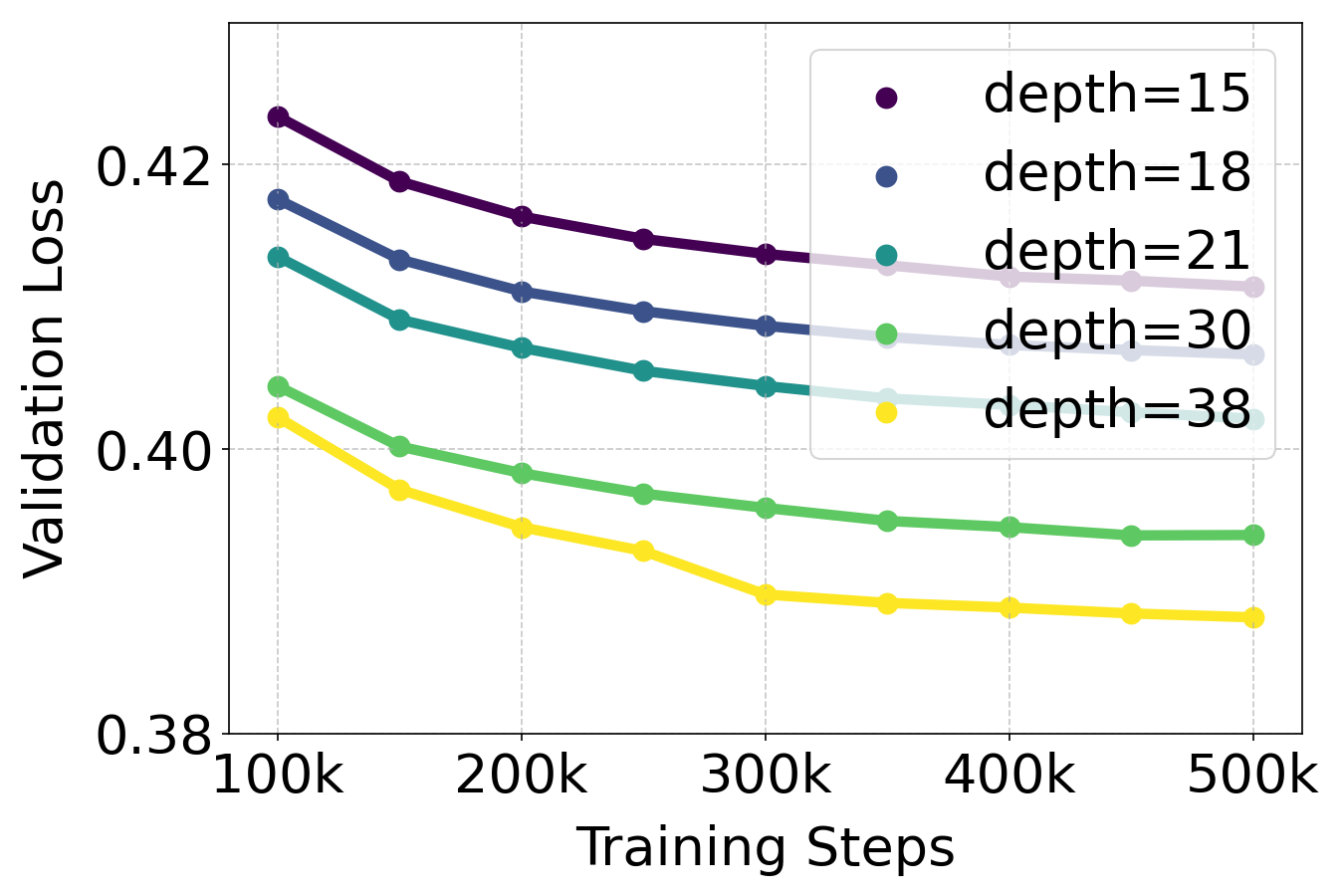}
&
\includegraphics[width=.5\linewidth,valign=m]{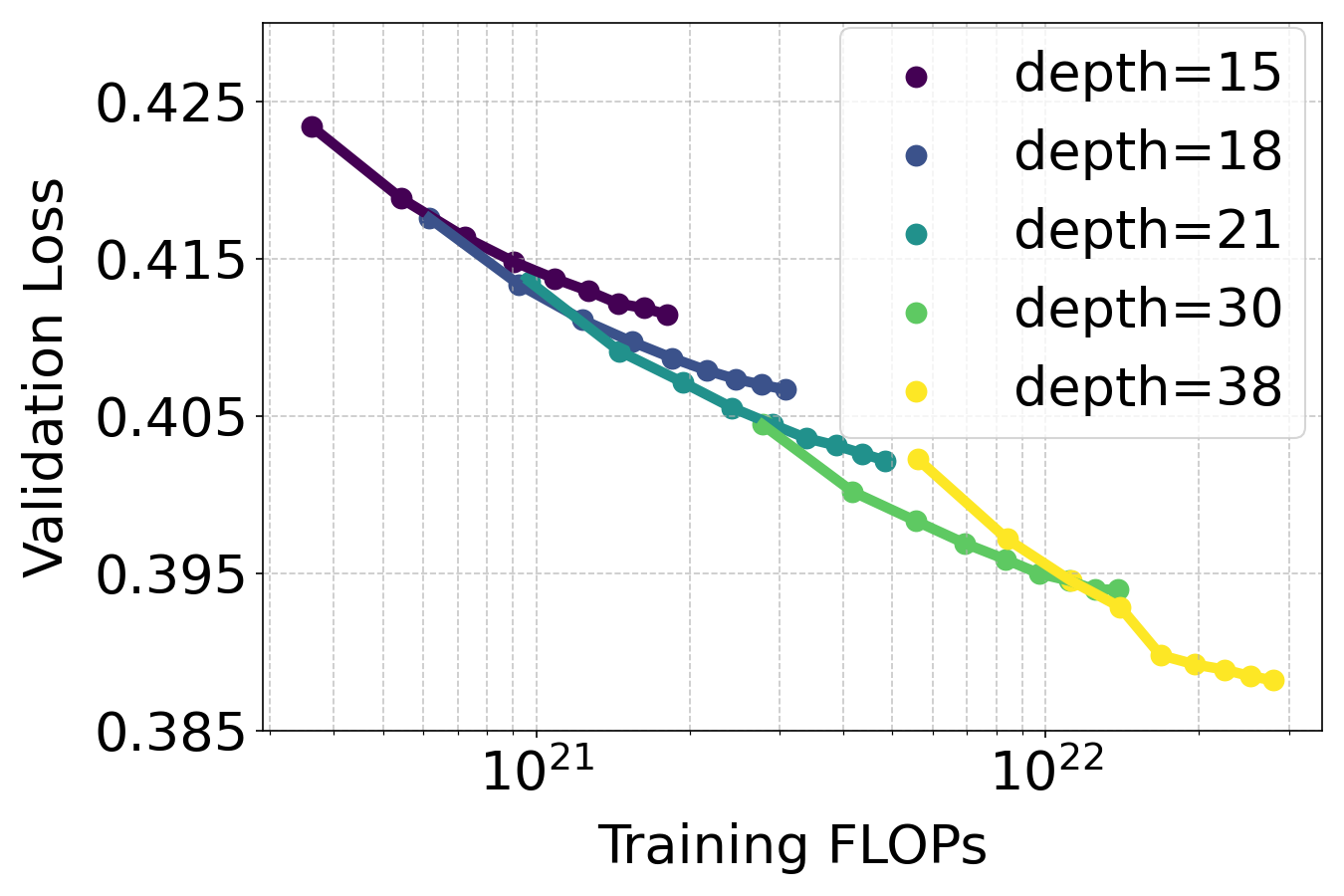} \\
\includegraphics[width=.5\linewidth,valign=m]{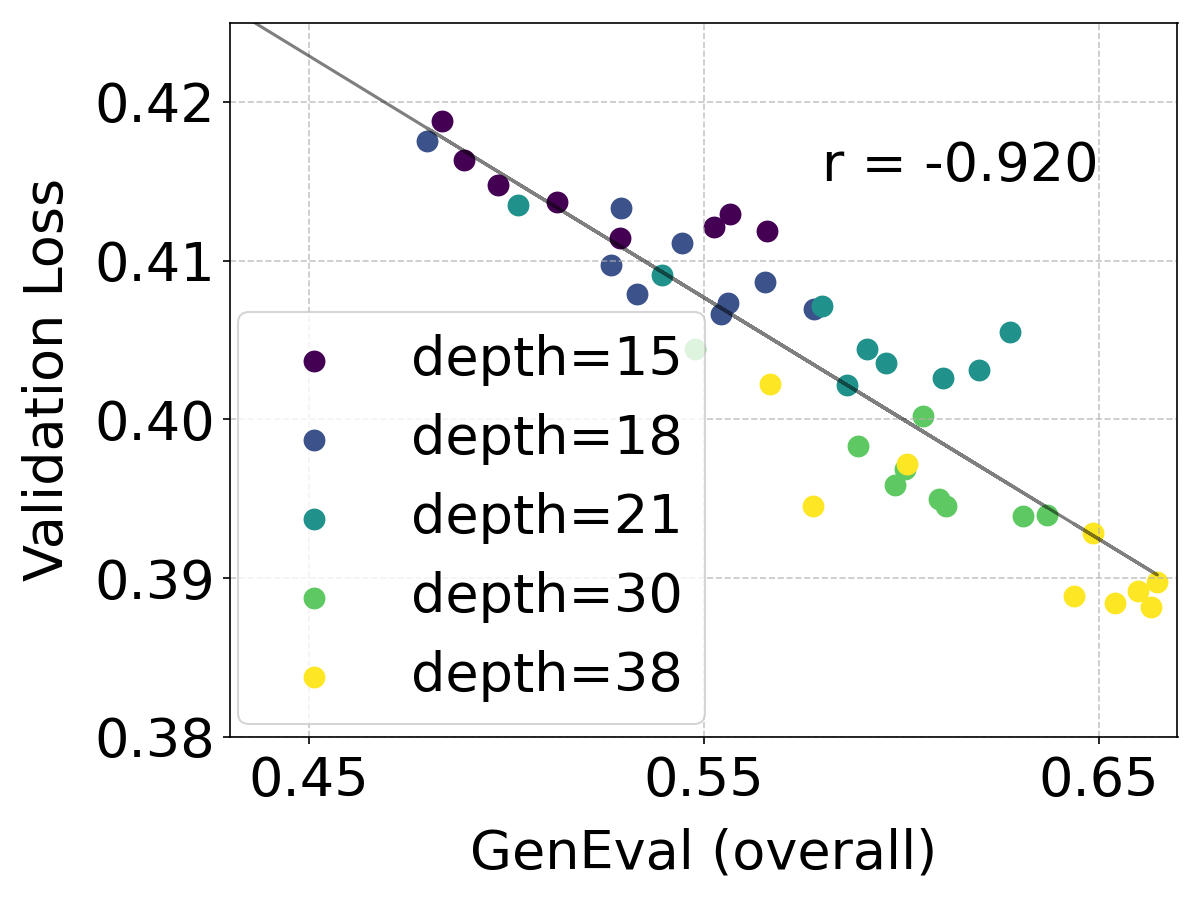}
&
\includegraphics[width=.5\linewidth,valign=m]{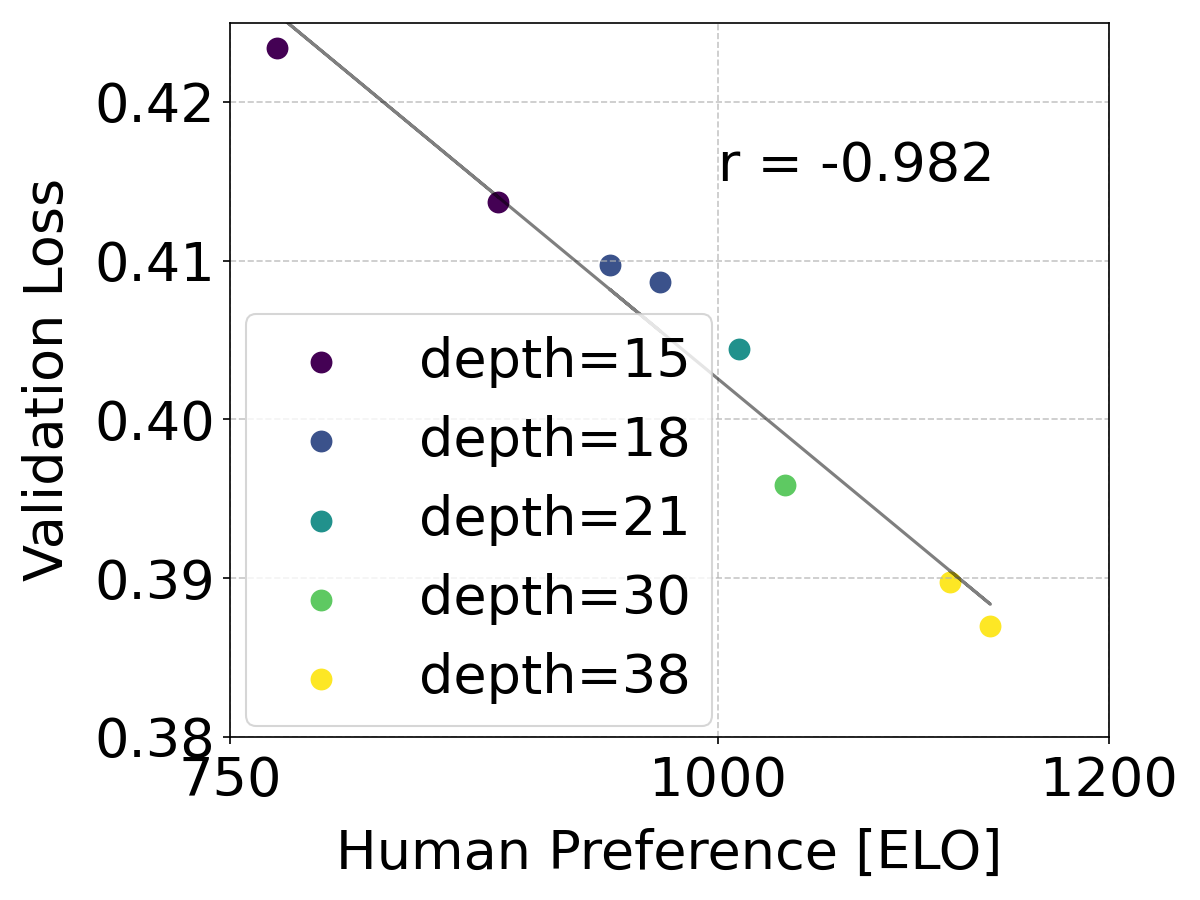} 
\end{tabular}
\caption{\label{fig:valscaling} \textbf{Quantitative effects of scaling.} \TODO{\textbf{larger plot. add third eval from appendix. add video exps, do video val loss vs human eval metrics}} We analyze the impact of model size on performance, maintaining consistent training hyperparameters throughout. 
An exception is depth=38, where learning rate adjustments at $3 \times 10^5$ steps were necessary to prevent divergence. 
(Top) Validation loss smoothly decreases as a function of both model size and training steps. 
(Bottom) Validation loss is a \textit{strong predictor of overall model performance}. There is a marked correlation between validation loss and holistic image evaluation metrics, including 
GenEval~\cite{ghosh2023geneval}, column 1, human preference, column 2, and T2I-CompBench~\cite{huang2023t2i}, column 3. For video models we observe a similar correlation between validation loss and human preference, column 4.
.}
\end{figure}
}

\newcommand{\figvalscaling}{
\setlength{\tabcolsep}{-1pt}
\begin{figure*}
\begin{tabular}{cccc}
\includegraphics[width=.25\linewidth,valign=m]{img/scale_val_squeeze/00_coco_val_loss_train-step.png}
&
\includegraphics[width=.25\linewidth,valign=m]{img/scale_val_squeeze/00_coco_val_loss_train-flops.png} &
\includegraphics[width=.25\linewidth,valign=m]{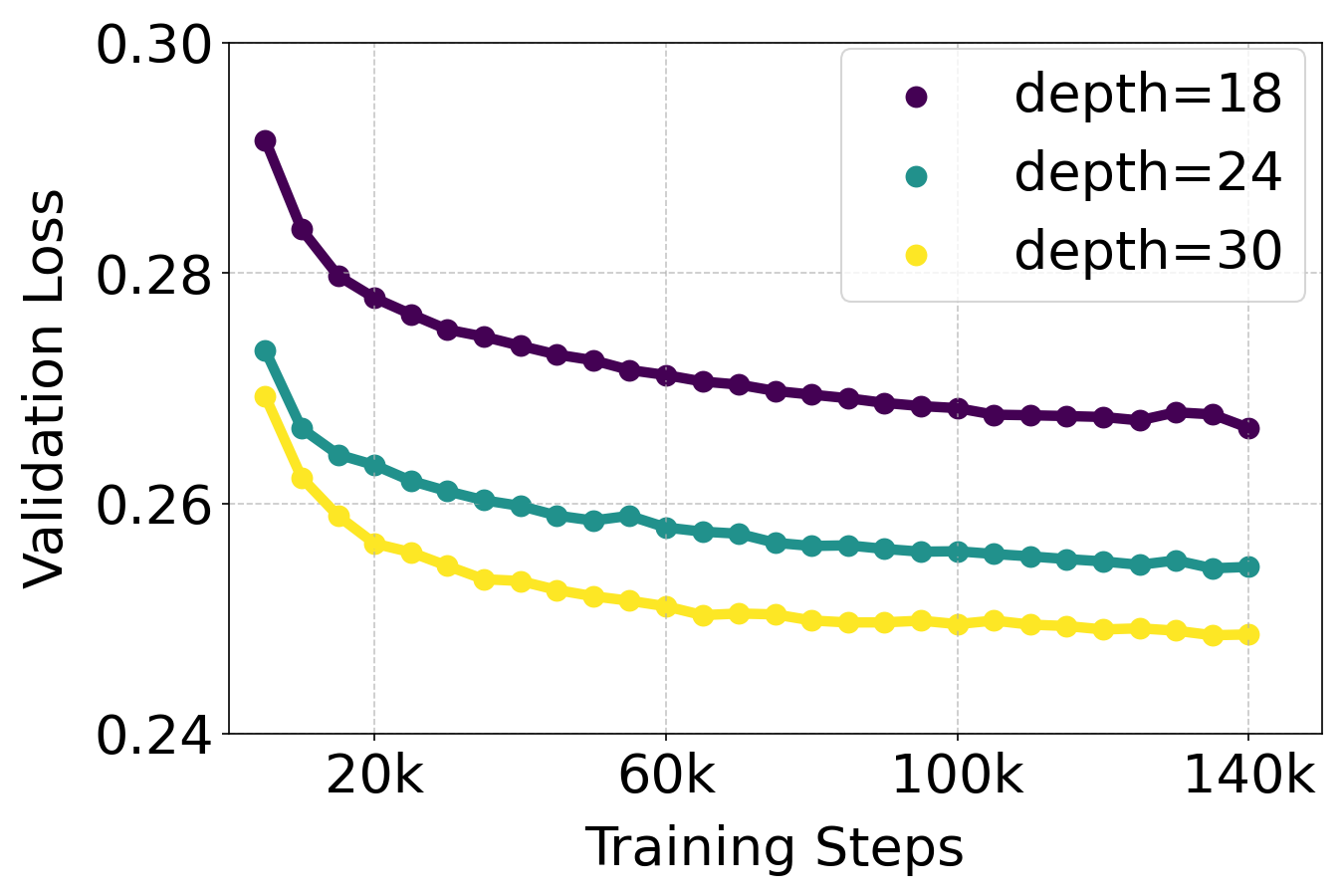} &
\includegraphics[width=.25\linewidth,valign=m]{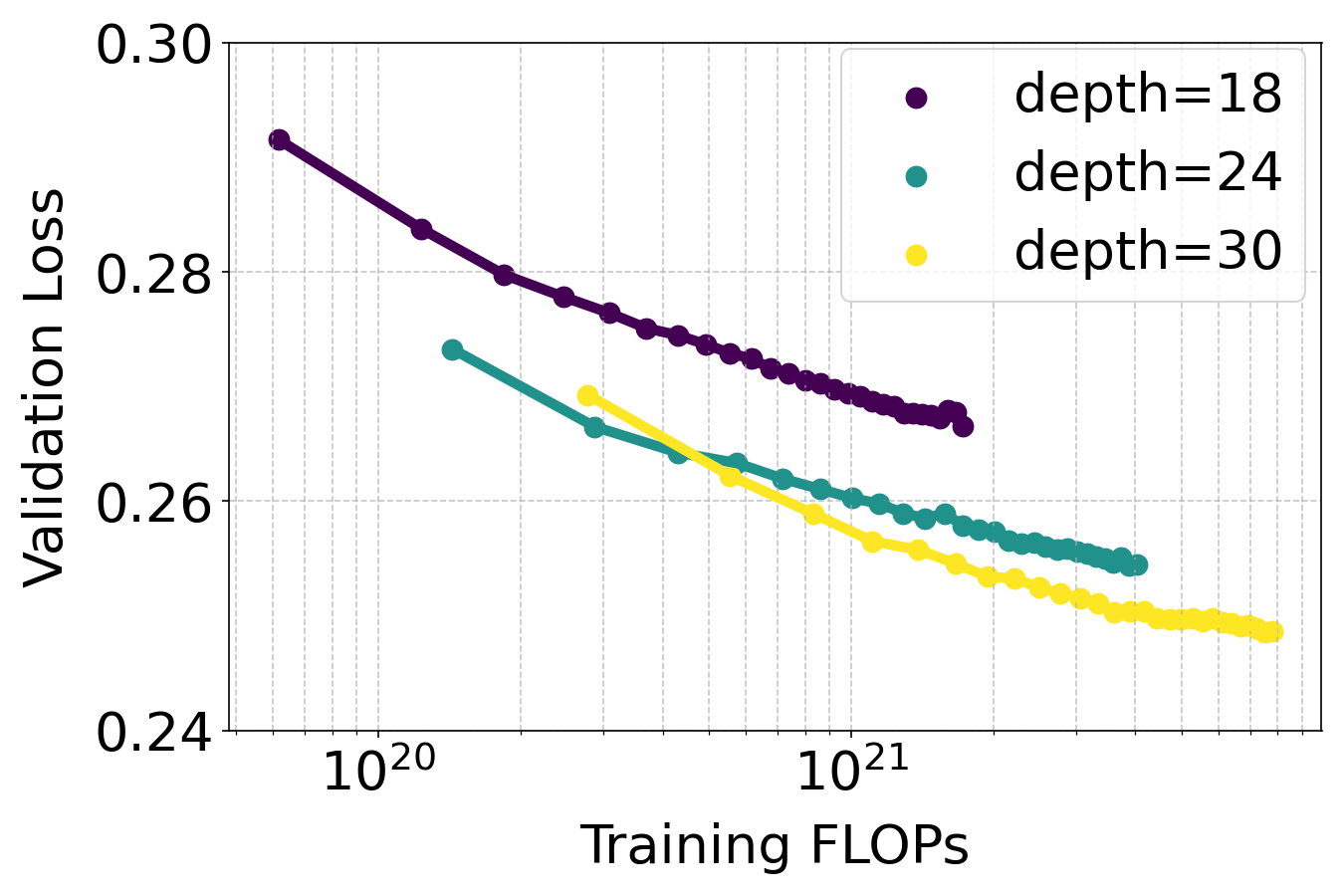}
 \\
\includegraphics[width=.25\linewidth,valign=m]{img/scale_val_squeeze/00_coco_val_loss_gen-eval.png}
&
\includegraphics[width=.25\linewidth,valign=m]{img/scale_val_squeeze/00_coco_val_loss_elo.png} 
&
\includegraphics[width=.25\linewidth,valign=m]{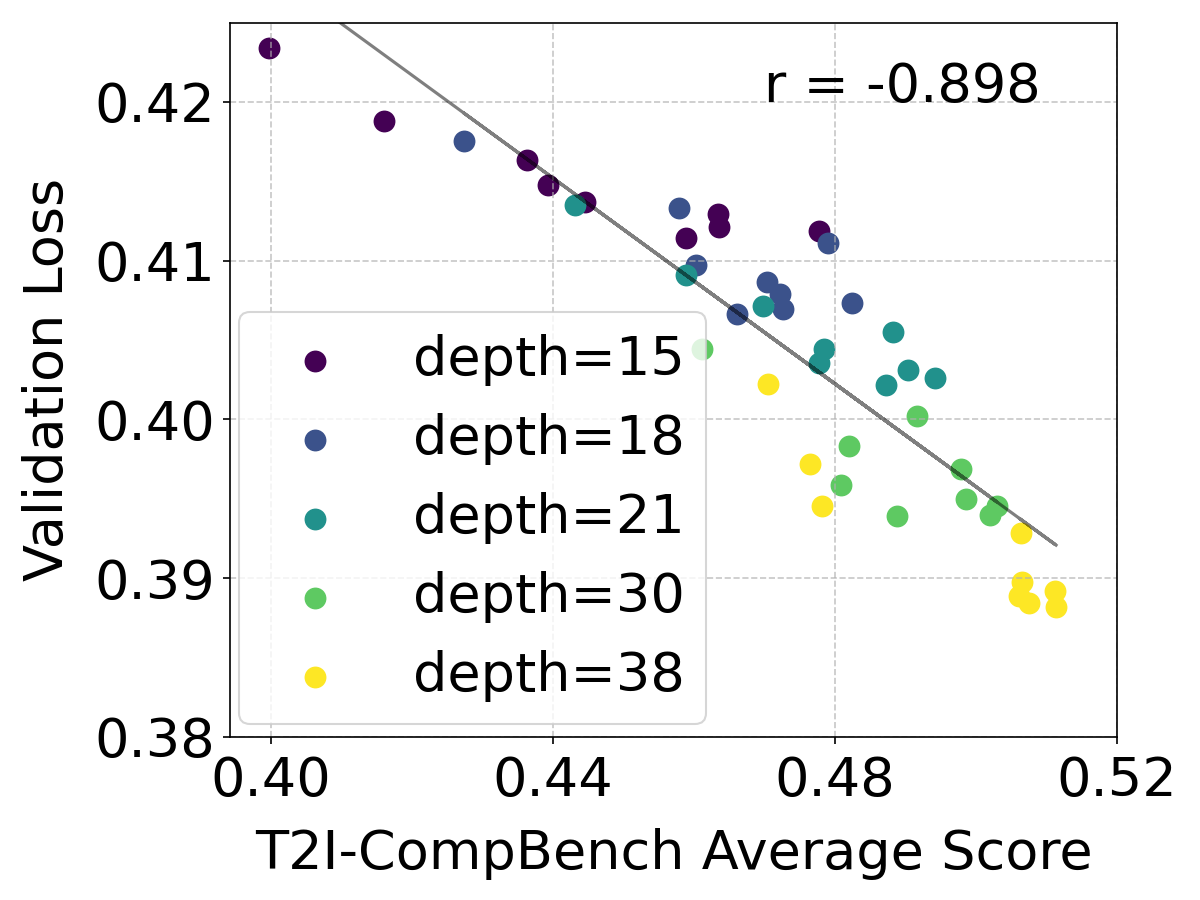}
&
\includegraphics[width=.25\linewidth,valign=m]{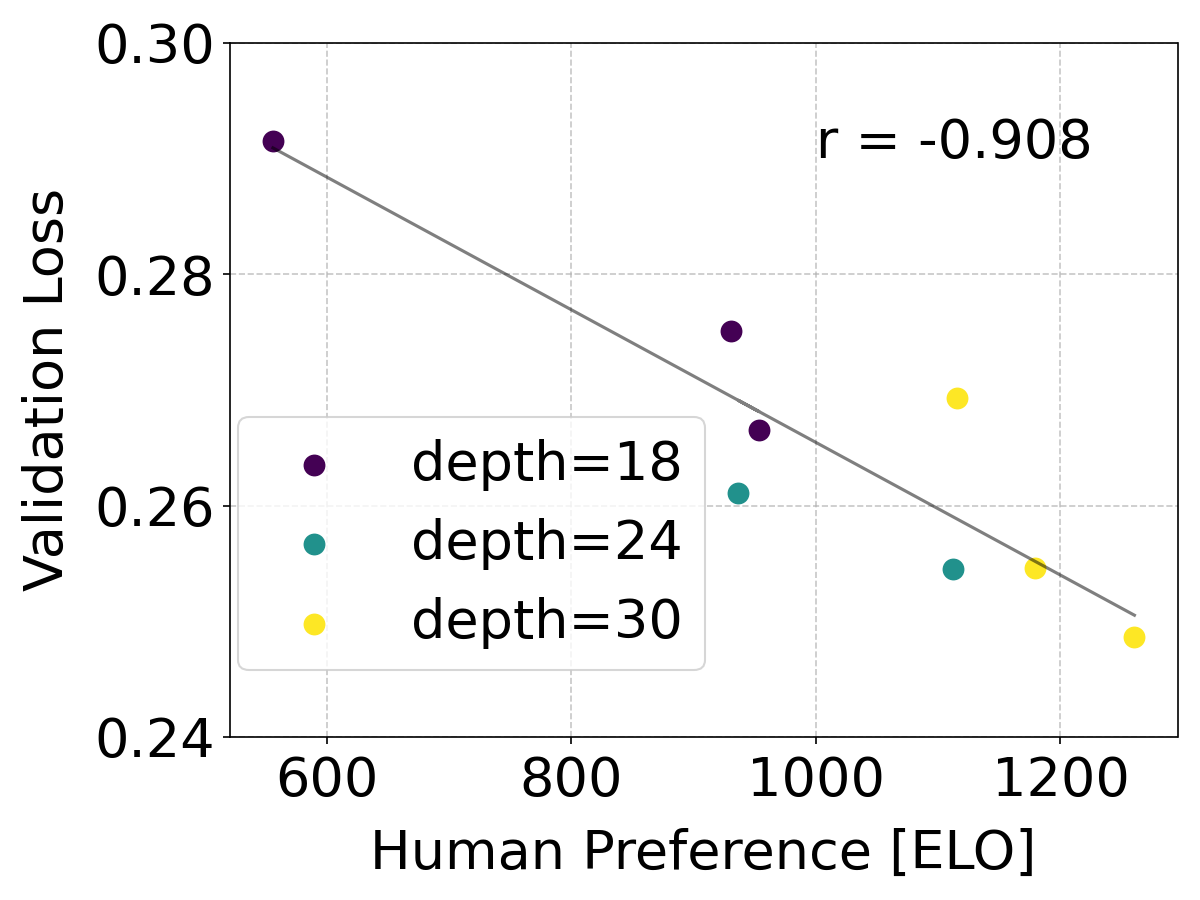} 
\end{tabular}
\caption{\label{fig:valscaling} \textbf{Quantitative effects of scaling.}  We analyze the impact of model size on performance, maintaining consistent training hyperparameters throughout. 
An exception is depth=38, where learning rate adjustments at $3 \times 10^5$ steps were necessary to prevent divergence. 
(Top) Validation loss smoothly decreases as a function of both model size and training steps for both image (columns 1 and 2) and video models (columns 3 and 4). 
(Bottom) Validation loss is a \textit{strong predictor of overall model performance}. There is a marked correlation between validation loss and holistic image evaluation metrics, including 
GenEval~\cite{ghosh2023geneval}, column 1, human preference, column 2, and T2I-CompBench~\cite{huang2023t2i}, column 3. For video models we observe a similar correlation between validation loss and human preference, column 4.
.}
\end{figure*}
}

\newcommand{\combenchcorr}{
\begin{figure}
\centering
\includegraphics[width=.5\linewidth,valign=m]{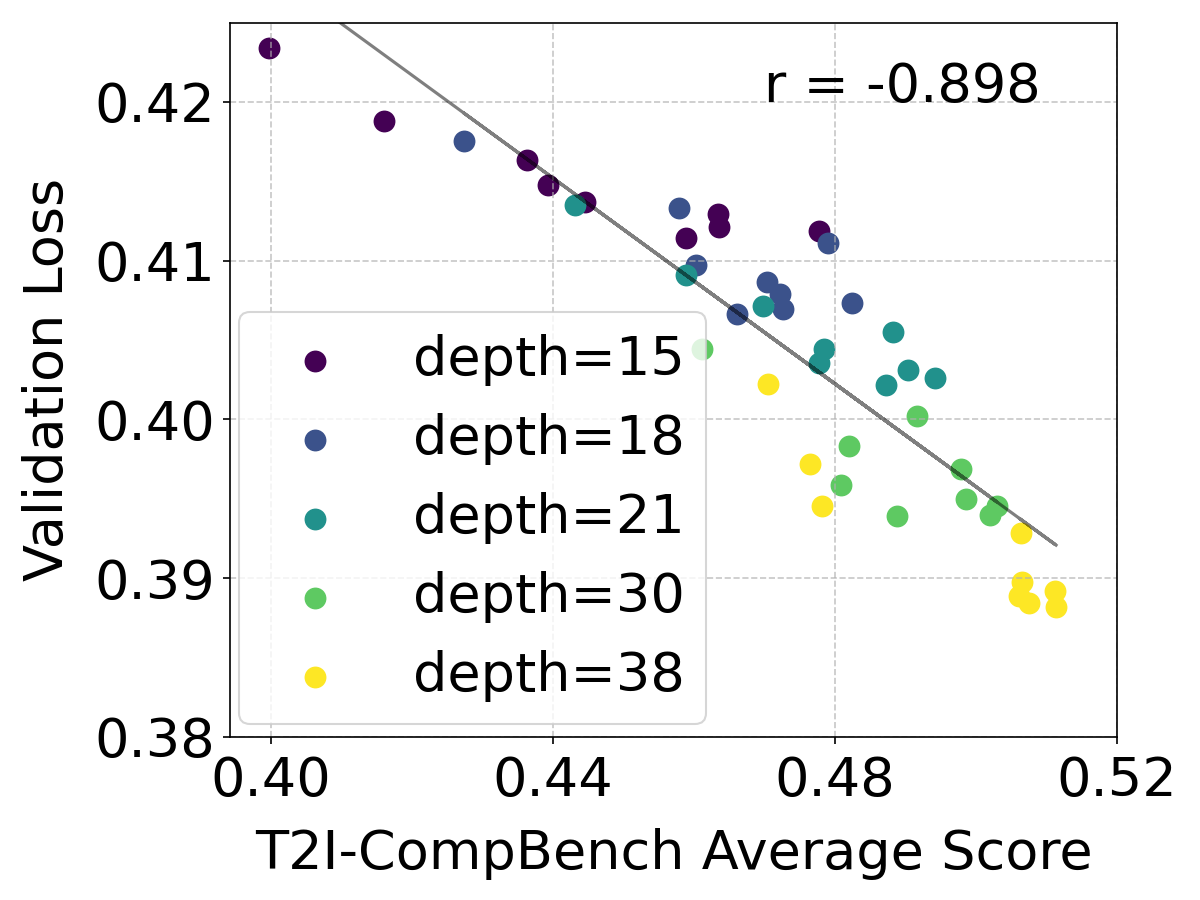}
\caption{\label{fig:combench_corr} \textbf{Correlation of validation loss and model performance as measured by CompBench~\cite{huang2023t2i}.} Similar to the findings for GenEval~\citep{ghosh2023geneval} and human preference presented in \Cref{fig:valscaling}, we find a strong correlation between validation loss and model performance on T2I-CompBench~\citep{huang2023t2i}, which provides further evidence that the validation loss is a strong predictor of model performance.}
\end{figure}
}

\newcommand{\figmembaselinessd}{
\begin{figure}\centering
\includegraphics[width=0.85\textwidth]{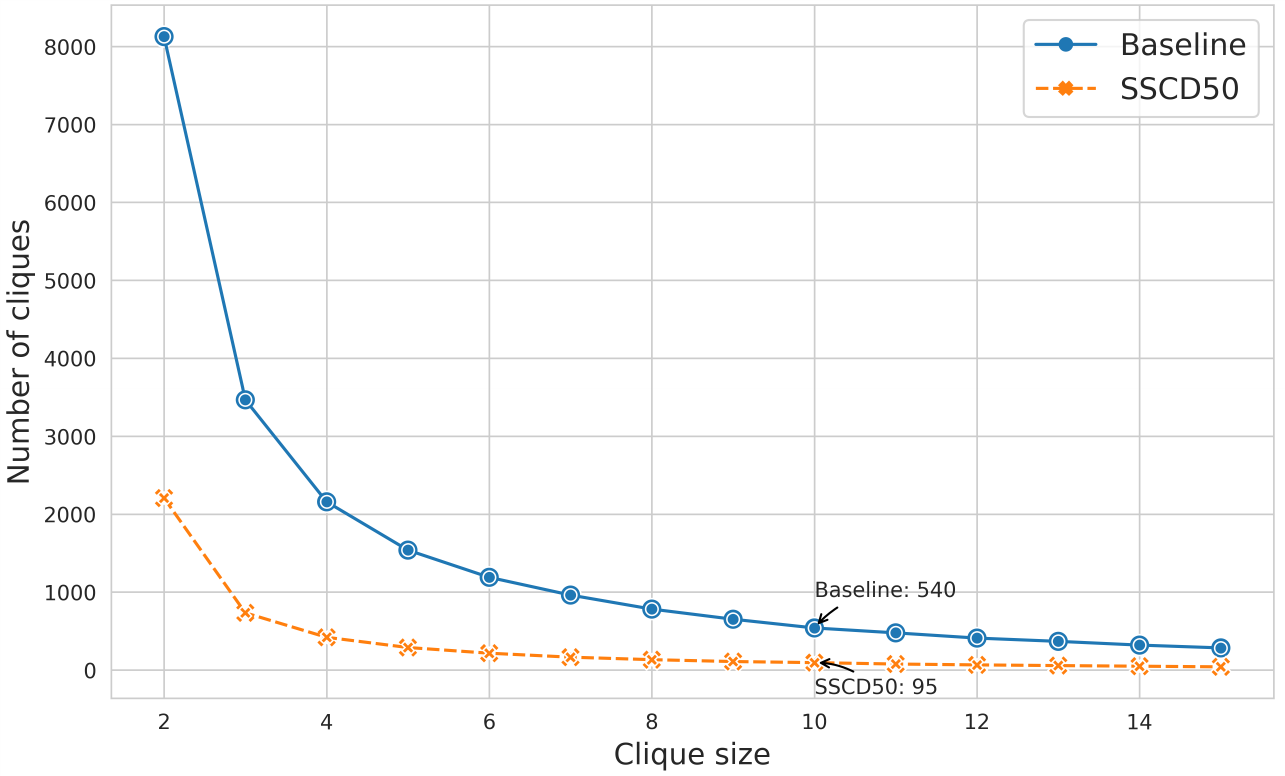}
\caption{\label{fig:memorization_sscd} \textbf{SSCD-based deduplication prevents memorization.} 
To assess how well our SSCD-based deduplication works, we extract memorized samples from
small, specifically for this purpose trained models and compare them before and after deduplication. 
We plot a comparison between number of memorized samples, before and after using SSCD with the threshold of 0.5 to remove near-duplicated samples. \citet{carlini2023extracting} mark images within clique size of 10 as memorized samples. Here we also explore different sizes for cliques. For all clique thresholds, SSCD is able to significantly reduce the number of memorized samples. Specifically, when the clique size is 10,  models on the deduplicated training samples cut off at SSCD$=0.5$ show a $5 \times $ reduction in potentially memorized examples.}
\end{figure}
}

\newcommand{\figmemsimilarity}{
\begin{figure}
\includegraphics[width=0.85\textwidth]{img/memorization/dup.pdf}
\caption{\label{fig:memorization_dup} Effect of model conditioning in memorization. Dup both means that both images and caotion are duplicated. We use BLIP when duplicating only images. Similar to ~\citet{somepalli2023understanding} when we train for 10K iterations, dulicating only images and using different captions for replicated images shows lower similarity score (memorization). This is a reminiscent of data augmentation effect, as a helping measure for generalization. However, our observation shows that this is very dependent on number of finetuning iterations and also guidance scale when sampling. }
\end{figure}
}

\newcommand{\figmemcaption}{
\begin{figure}
\includegraphics[width=0.85\textwidth]{img/memorization/dup_scale_7.5.pdf}
\caption{\label{fig:memorization_caption} Effect of different captioning in memorization. Using different captions when employing partial duplication (dup image) lead to different similarity scores.}
\end{figure}
}

\newcommand{\figdupcombined}{
\begin{figure}
\centering
\begin{subfigure}[t]{0.49\textwidth}\centering
        \includegraphics[width=\textwidth]{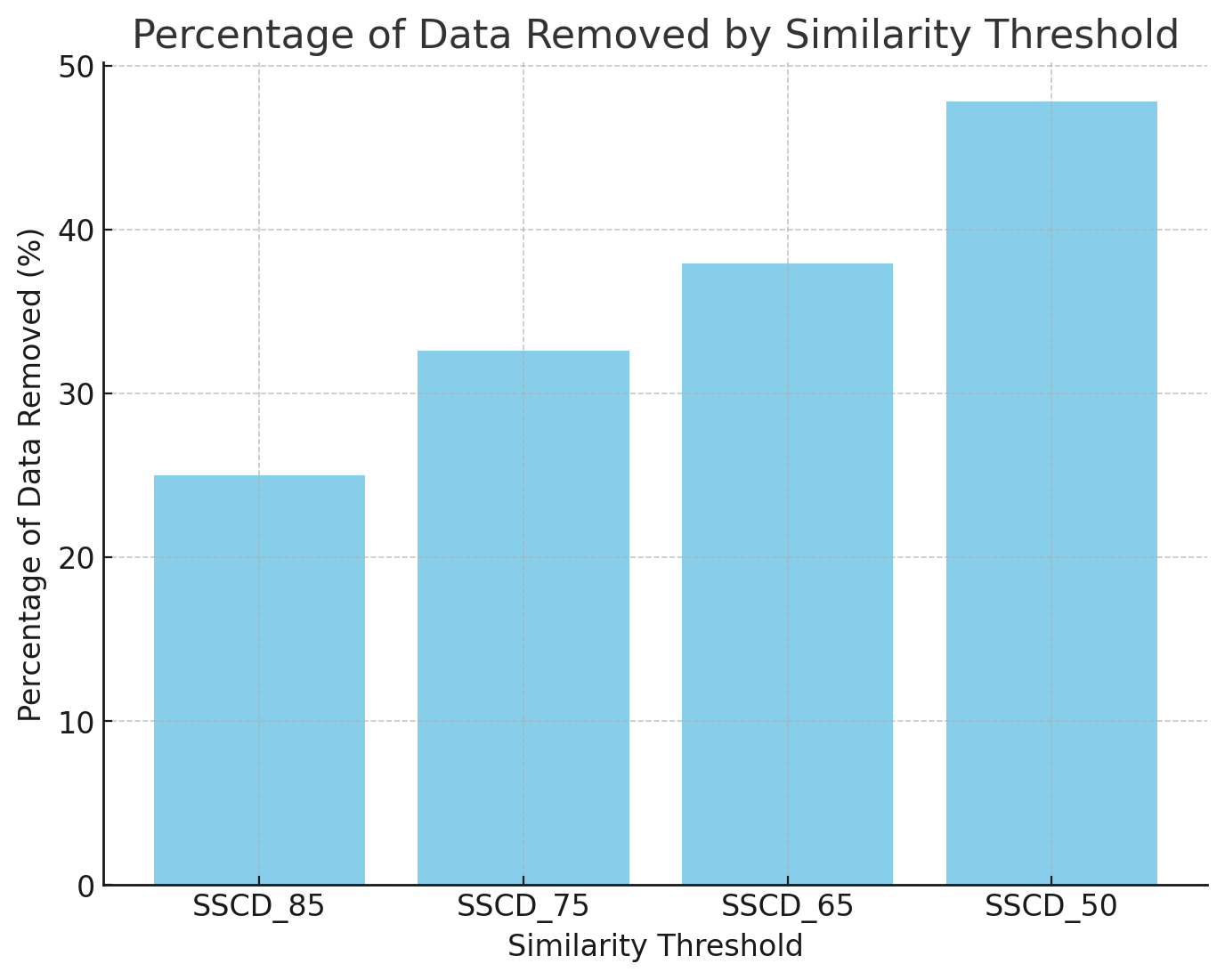}
        \caption{\label{fig:sscd_final_result}Final result of SSCD deduplication over the entire dataset}
    \end{subfigure}\hfill%
    \begin{subfigure}[t]{0.49\textwidth}\centering
        \includegraphics[width=\textwidth]{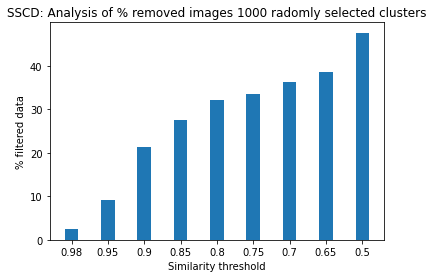}
        \caption{\label{fig:sscd_ablation}Result of SSCD deduplication with various thresholds over 1000 random clusters}
    \end{subfigure}
\caption{\label{fig:dedup} Results of deduplicating our training datasets for various filtering thresholds.}
\end{figure}
}

\newcommand{\figdupsscdrandom}{
\begin{figure}\centering
\includegraphics[width=0.45\textwidth]{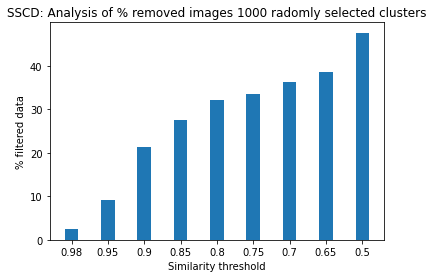}
\caption{\label{fig:sscd_ablation}Result of SSCD deduplication with various thresholds over 1000 random clusters}
\end{figure}
}

\newcommand{\figdupsscdfinal}{
\begin{figure}\centering
\includegraphics[width=0.45\textwidth]{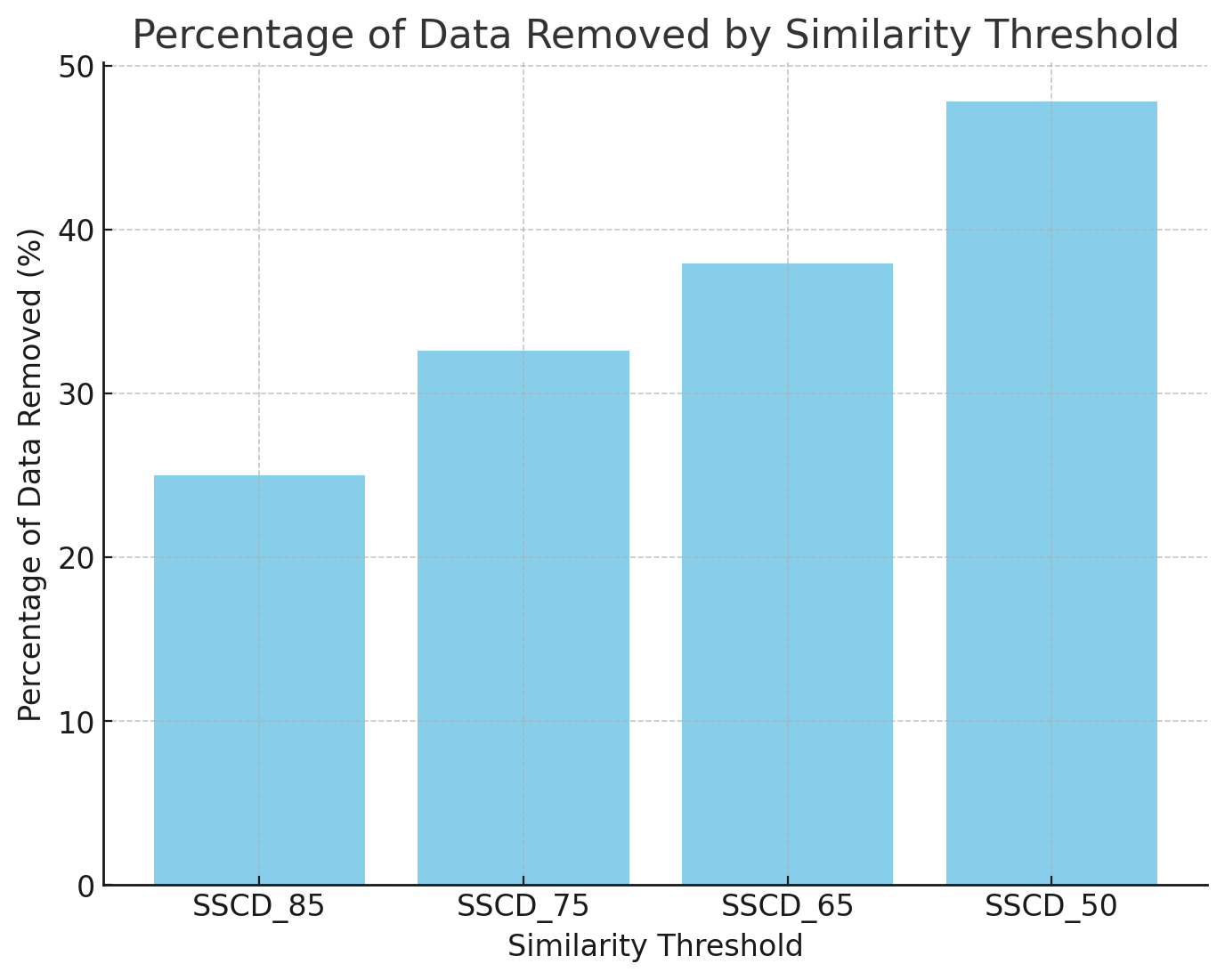}
\caption{\label{fig:sscd_final_result}Final result of SSCD deduplication over the entire dataset}
\end{figure}
}

\newcommand{\figfidvssteps}{
\begin{figure}
\includegraphics[width=0.45\textwidth]{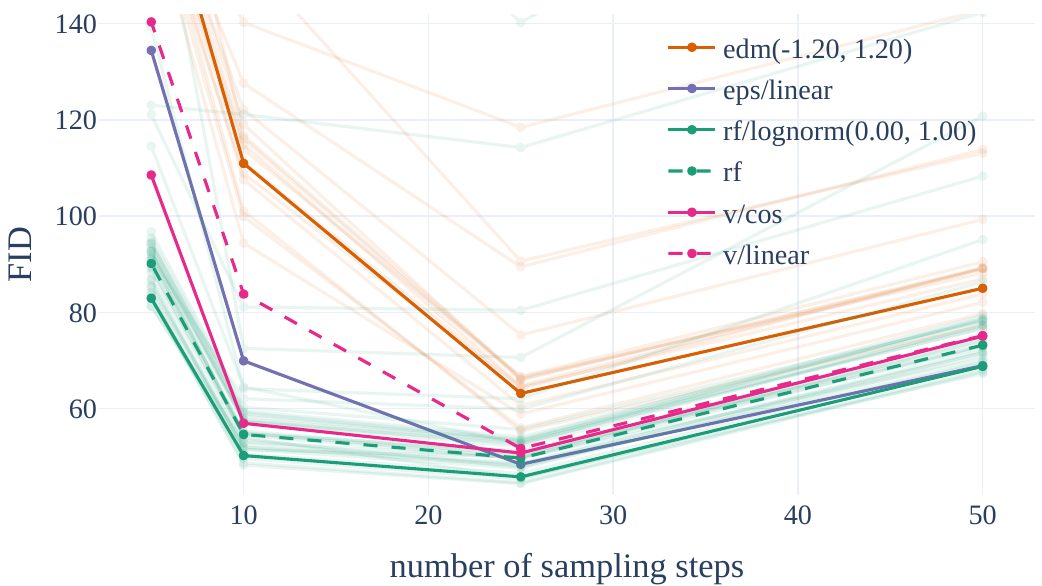}
\vspace{-0.5em}
\caption{\label{fig:fidvssteps}
\textbf{Rectified flows are sample efficient.}
Rectified Flows perform better then other formulations when sampling fewer steps. For 25 and more steps, only \texttt{rf/lognorm(0.00, 1.00)} remains competitive to \texttt{eps/linear}. \vspace{-0.75em}}
\end{figure}
}

\newcommand{\algclusterdeduplication}{%
\begin{algorithm}
\caption{Finding Duplicate Items in a Cluster}
\label{alg:cluster_deduplication}
\begin{algorithmic}[1]
     \REQUIRE $\mathtt{vecs}$ -- List of vectors in a single cluster, 
             $\mathtt{items}$ -- List of item IDs corresponding to vecs, 
             $\mathtt{index}$ -- FAISS index for similarity search within the cluster,  
             $\mathtt{thresh}$ -- Threshold for determining duplicates 

    \textbf{Output:} $\mathtt{dups}$ -- Set of duplicate item IDs 
    \STATE $\mathtt{dups} \gets \text{new set}()$
    \FOR{$i \gets 0$ \textbf{to} $\mathrm{length}(\mathtt{vecs}) - 1$}
        \STATE $\mathtt{qs} \gets \mathtt{vecs}[i]$ \COMMENT{Current vector}
        \STATE $\mathtt{qid} \gets \mathtt{items}[i]$ \COMMENT{Current item ID}
        \STATE $\mathtt{lims}, D, I \gets \mathtt{index}.\mathrm{range\_search}(\mathtt{qs}, \mathtt{thresh})$
        \IF{$\mathtt{qid} \in \mathtt{dups}$}
            \STATE \textbf{continue}
        \ENDIF
        \STATE $\mathtt{start} \gets \mathtt{lims}[0]$
        \STATE $\mathtt{end} \gets \mathtt{lims}[1]$
        \STATE $\mathtt{duplicate\_indices} \gets I[start:end]$ 
        \STATE $\mathtt{duplicate\_ids} \gets \mathrm{new\ list}()$
        \FOR{$j$ \textbf{in} $\mathtt{duplicate\_indices}$}
            \IF{$\mathtt{items}[j] \neq \mathtt{qid}$}
                \STATE $\mathtt{duplicate\_ids}.\mathrm{append}(\mathtt{items}[j])$
            \ENDIF
        \ENDFOR
        \STATE $\mathtt{dups}.\mathrm{update}(\mathtt{duplicate\_ids})$
    \ENDFOR
    \STATE \textbf{{Return} $\mathtt{dups}$} \COMMENT{Final set of duplicate IDs}
\end{algorithmic}
\end{algorithm}%
}

\newcommand{\algmemorizationdetection}{%
\begin{algorithm}
\caption{Detecting Memorization in Generated Images}
\label{alg:memorization_detection}
\begin{algorithmic}[1]
    \REQUIRE Set of prompts $P$, Number of generations per prompt $N$, Similarity threshold $\epsilon=0.15$, Memorization threshold $T$
    \ENSURE Detection of memorized images in generated samples
    \STATE Initialize $D$ to the set of most-duplicated examples
    \FOR{each prompt $p \in P$}
        \FOR{$i = 1$ to $N$}
            \STATE Generate image $\mathrm{Gen}(p; r_i)$ with random seed $r_i$
        \ENDFOR
    \ENDFOR
    \FOR{each pair of generated images $x_i, x_j$}
        \IF{distance $d(x_i, x_j) < \epsilon$}
            \STATE Connect $x_i$ and $x_j$ in graph $G$
        \ENDIF
    \ENDFOR
    \FOR{each node in $G$}
        \STATE Find largest clique containing the node
        \IF{size of clique $\geq T$}
            \STATE Mark images in the clique as memorized
        \ENDIF
    \ENDFOR
\end{algorithmic}
\end{algorithm}
}

\newcommand{\figdpo}{
\begin{figure}[htbp]
\vspace{-1em}
\centering
\begin{tabular}{c@{\hspace{5pt}}c@{\hspace{3pt}}c}
\toprule
& \footnotesize{\shortstack{\emph{``a peaceful lakeside landscape with} \\ \emph{migrating herd of sauropods''}}} & \footnotesize{\shortstack{\emph{``a book with the words `Don't Panic!`,} \\ \emph{written on it''}}} \\ 
\midrule
\rotatebox[origin=l,y=1.2em]{90}{{\textbf{2B base}}} &
\includegraphics[width=.47\textwidth]{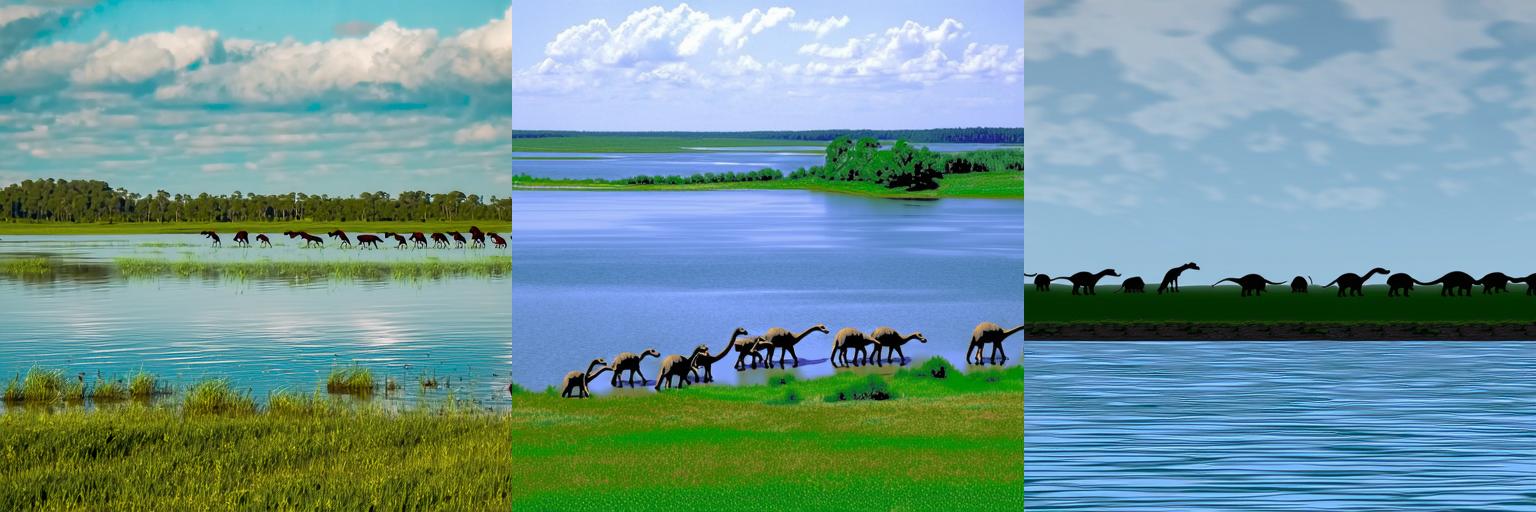} &
\includegraphics[width=.47\textwidth]{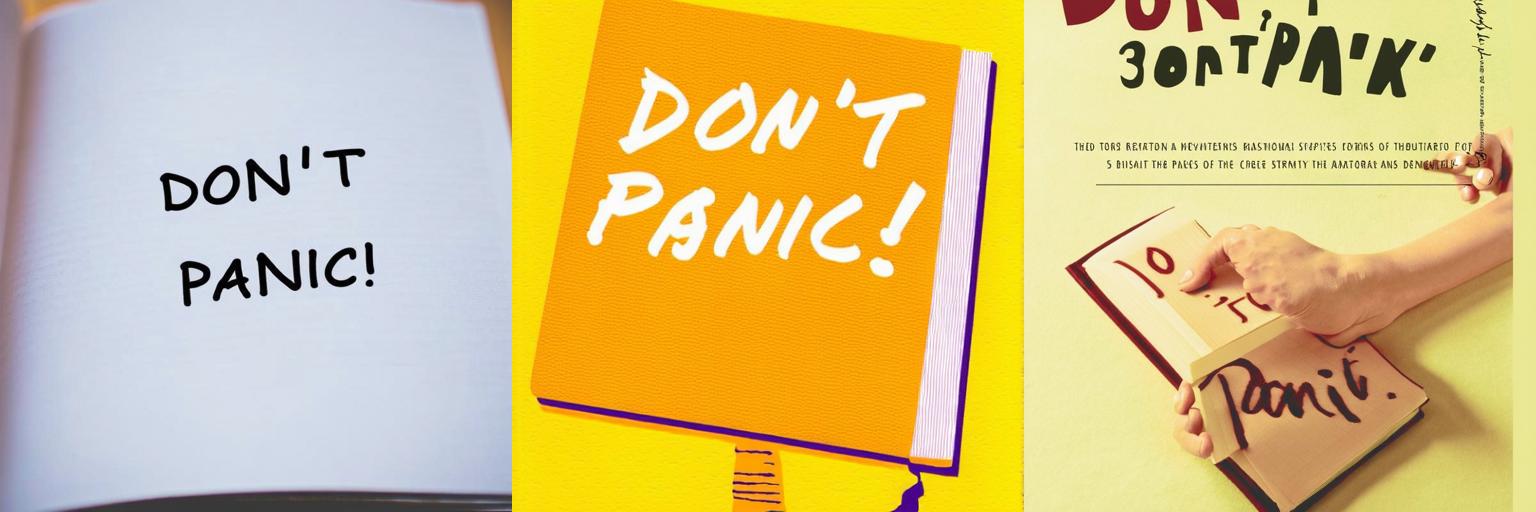} \\ %

\rotatebox[origin=l,y=1.2em]{90}{{\textbf{2B w/ DPO}}} &
\includegraphics[width=.47\textwidth]{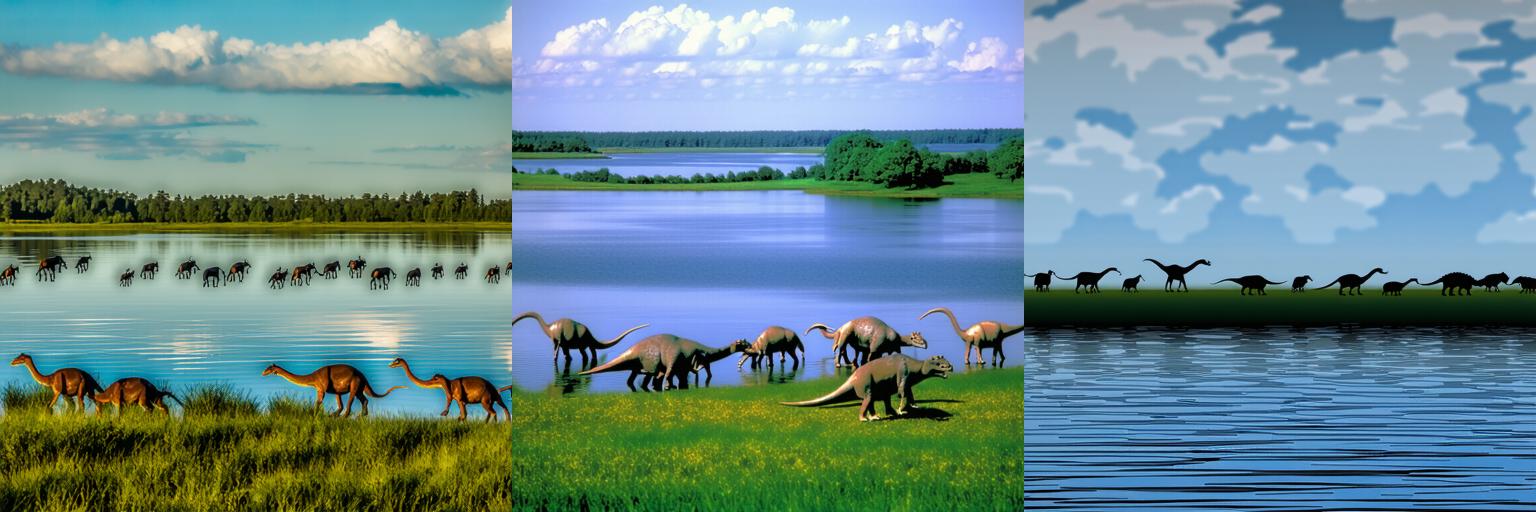} &
\includegraphics[width=.47\textwidth]{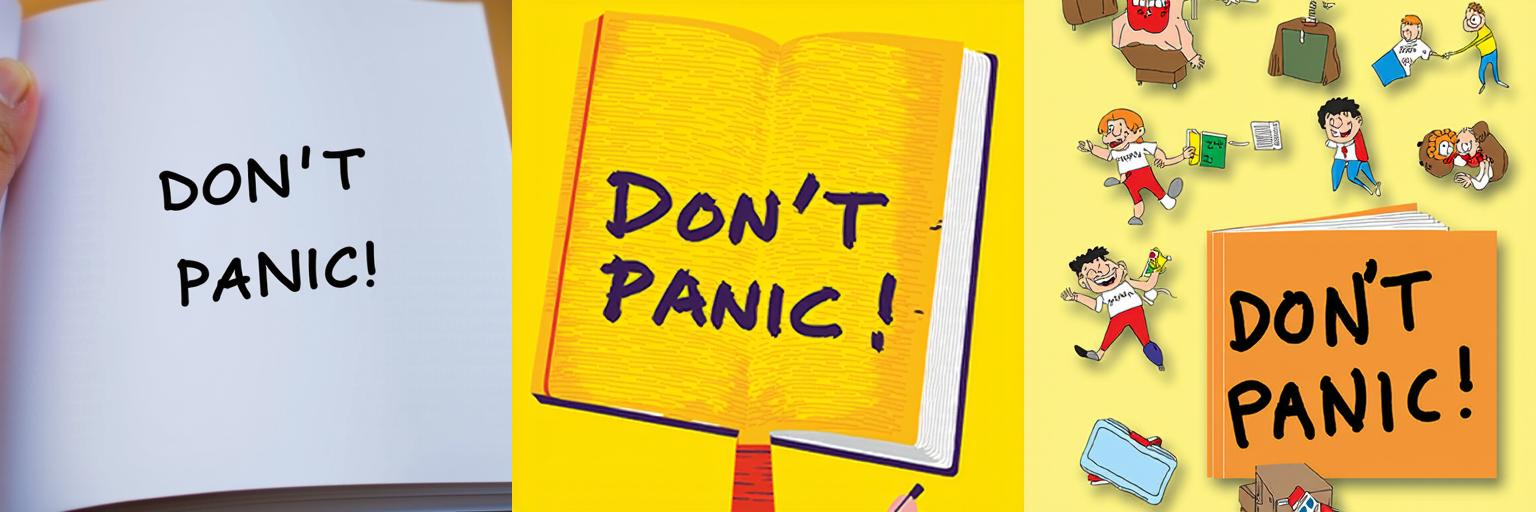} \\ %

\midrule

\rotatebox[origin=l,y=1.2em]{90}{{\textbf{8b base}}} &
\includegraphics[width=.47\textwidth]{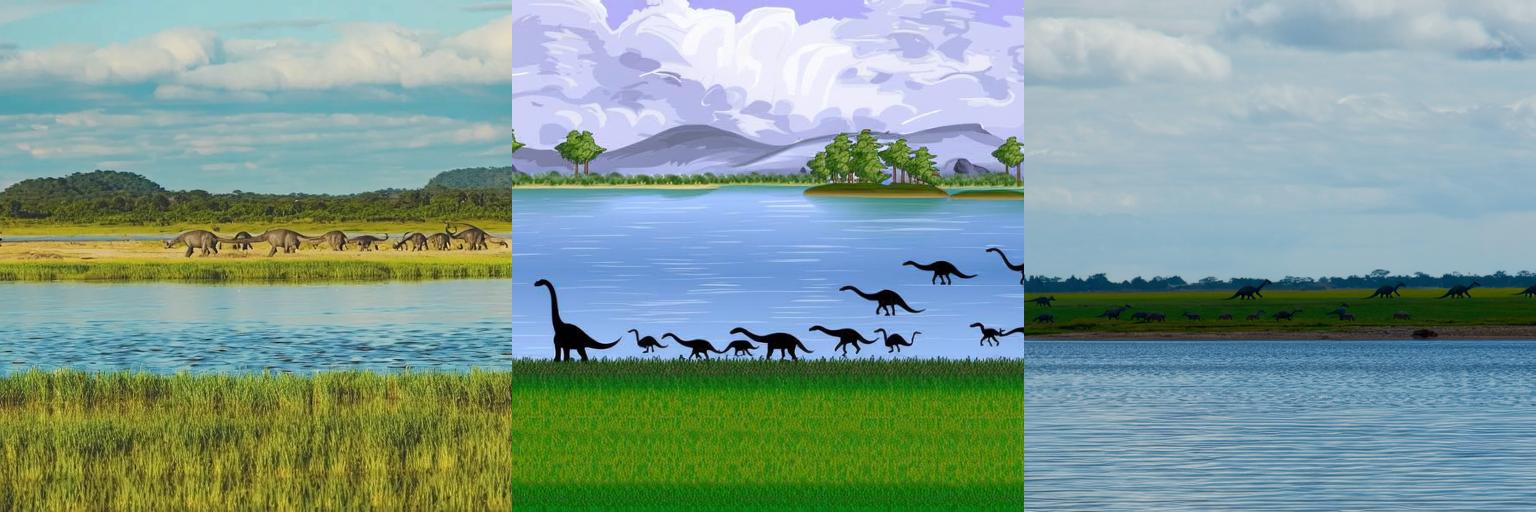} &
\includegraphics[width=.47\textwidth]{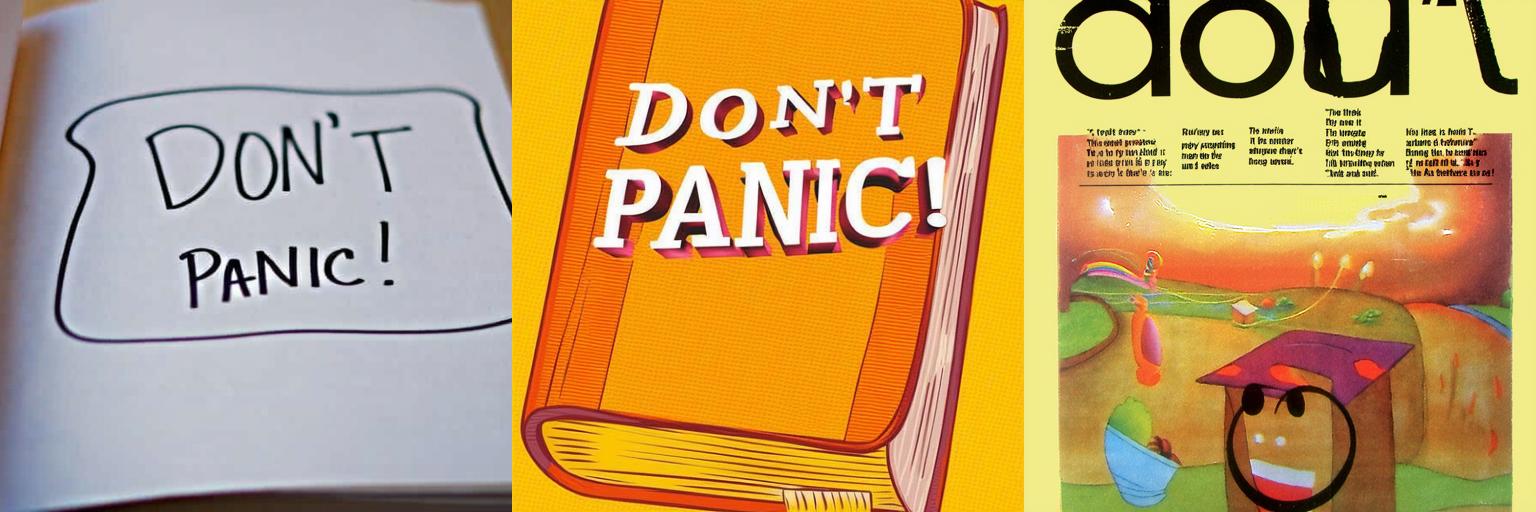} \\ %

\rotatebox[origin=l,y=1.2em]{90}{{\textbf{8b w/ DPO}}} &
\includegraphics[width=.47\textwidth]{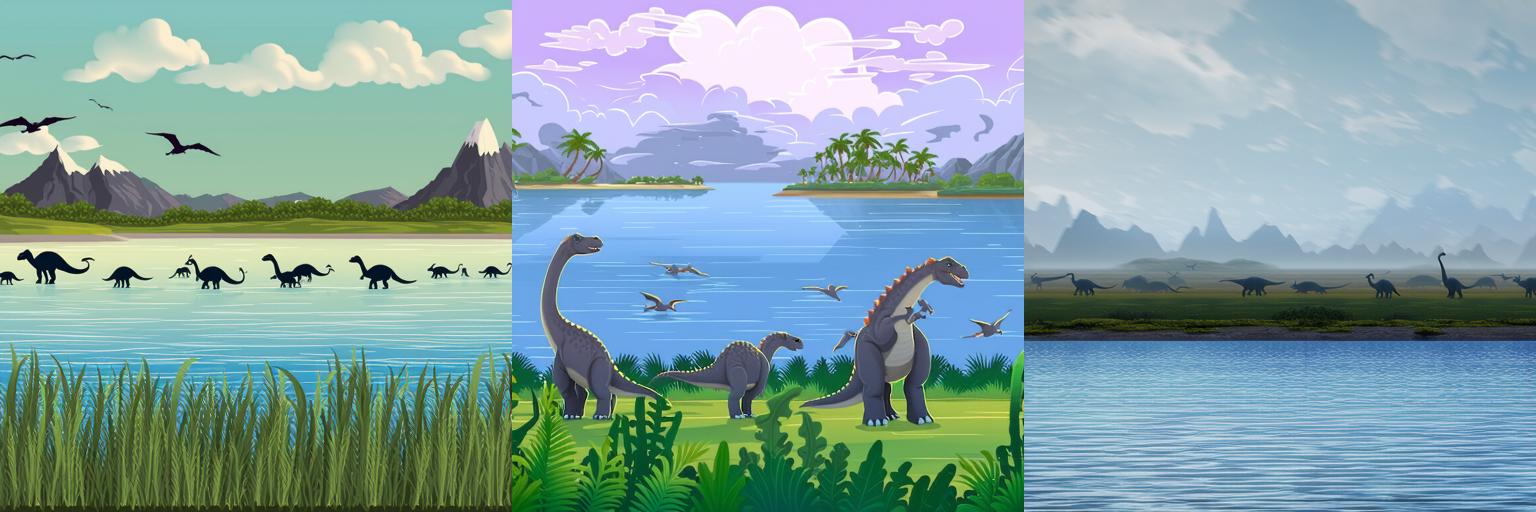} &
\includegraphics[width=.47\textwidth]{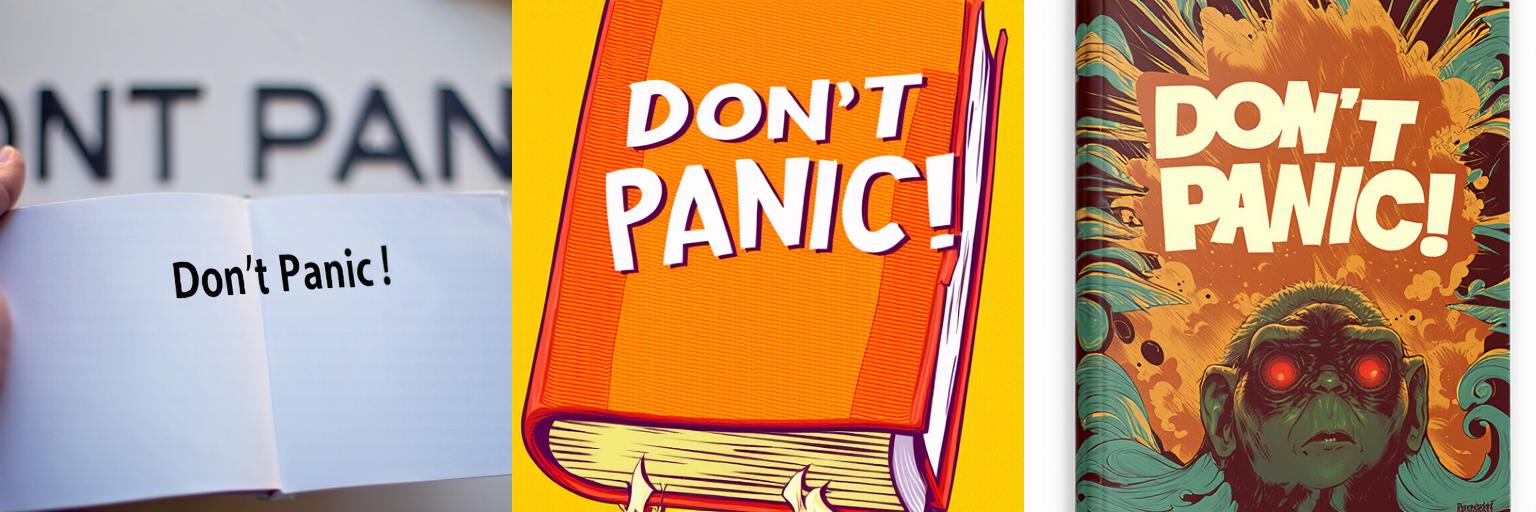} \\ %
\bottomrule
\end{tabular}
\caption{Comparison between base models and DPO-finetuned models. DPO-finetuning generally results in more aesthetically pleasing samples with better spelling.
} \label{fig:dpo}
\end{figure}
}

\newcommand{\sotaeval}{
\begin{figure}
\includegraphics[width=\linewidth,valign=m]{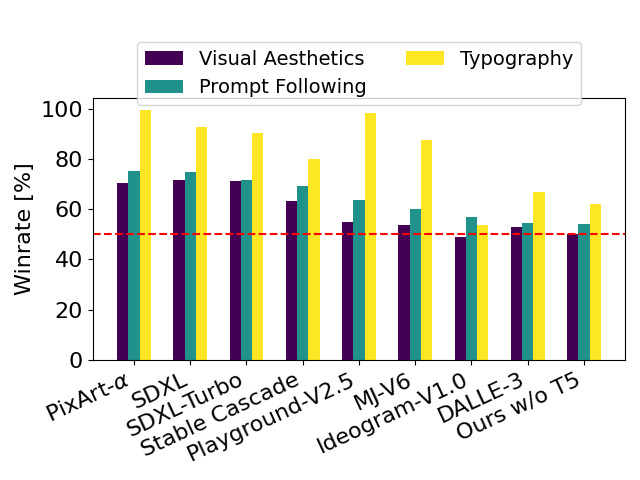}
\caption{\label{fig:sota_human_eval} \textbf{Human Preference Evaluation against currrent closed and open SOTA generative image models.} Our 8B model compares favorable against 
current state-of-the-art text-to-image models when evaluated on the parti-prompts~\cite{yu2022scaling} across the categories \emph{visual quality}, \emph{prompt following} and \emph{typography generation}.}
\end{figure}
}

\newcommand{\figdpohumaneval}{
\begin{figure}
\centering
\includegraphics[width=0.85\textwidth]{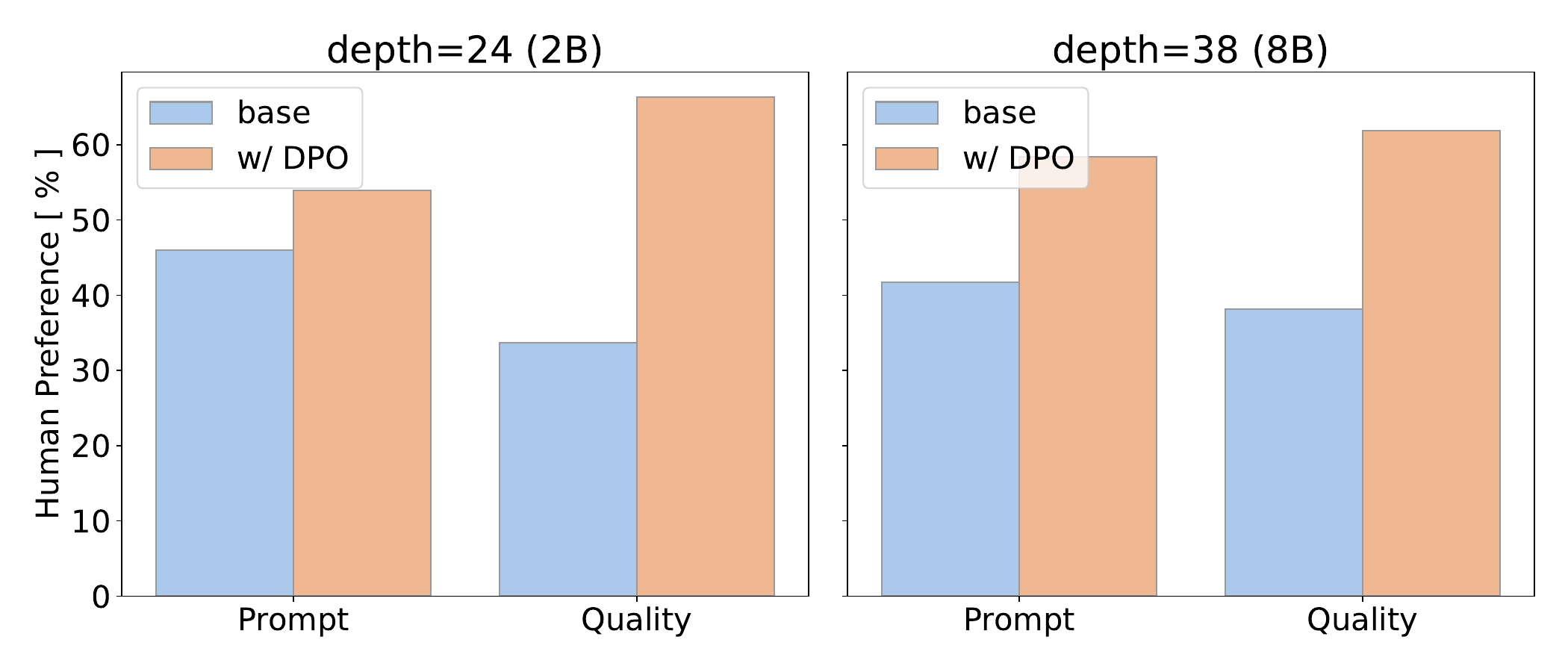}
\caption{Human preference evaluation between base models and DPO-finetuned models. Human evaluators prefer DPO-finetuned models for both prompt following and general quality.} \label{fig:dpo_human_eval}
\end{figure}
}

\newcommand{\modelarchitecture}{%
\begin{figure*}[t]
\centering
\begin{subfigure}[t]{0.49\textwidth}\centering
        \resizebox{!}{0.46\textheight}{\begin{tikzpicture}[
        block/.style={draw, fill=white, rectangle, minimum width=3cm,rounded corners=0.2cm},
        trainableblock/.style={block,fill=blue!10},
        frozenblock/.style={block,fill=blue!10},
        fnblock/.style={block,fill=green!10},
        operation/.style={draw, fill=red!10, circle, minimum width=2em},
        tensor/.style={draw, fill=VioletRed!50, circle},
        input/.style={block,fill=WildStrawberry!40},
    ]

    \pgfdeclarelayer{arrowlayer}
    \pgfdeclarelayer{residuallayer}
    \pgfdeclarelayer{decorations}
    \pgfdeclarelayer{bg}
    \pgfsetlayers{bg,arrowlayer,residuallayer,main,decorations}

    \newcommand{\clipGcolor}{LimeGreen!40}
    \newcommand{\clipLcolor}{ForestGreen!40}
    \newcommand{\tfiveColor}{YellowGreen!40}

    \node [input] (text) {Caption};

    \node [frozenblock, fill=\clipGcolor, below=1cm of text] (clipG) {CLIP-L/14};
    \node [frozenblock, fill=\clipLcolor, left=1.5cm of clipG.center] (clipL) {CLIP-G/14};
    \node [frozenblock, fill=\tfiveColor, right=1.5cm of clipG.center] (t5) {T5 XXL\phantom{/}};

    \coordinate [below=2.5cm of clipG] (conditioning);
    \coordinate [left=3.5cm of conditioning] (pool_conditioning);
    \draw [fill=\tfiveColor] (conditioning) |-++ (1.5, 1) coordinate (conditioning_br) --++ (0, -1.5) coordinate (conditioning_tr) -| cycle;
    \draw [fill=\clipLcolor] ($(conditioning)+(0, 0.25)$) -|++ (-1.5, 0.5) coordinate (conditioning_bl) -| cycle;
    \draw [fill=\clipGcolor] ($(conditioning)+(0, 0.75)$) -|++ (-1.5, 0.25) -| cycle;
    \draw [fill=black!10] ($(conditioning)+(0, 0.25)$) -|++ (-1.5, -0.75) coordinate (conditioning_tl) -| cycle;

    \draw [->] (t5) to [out=270, in=0] ($(conditioning_br)!0.5!(conditioning_tr)+(-0.5, 0)$);
    \draw [->] (clipG) to [out=270, in=90] ($(conditioning_bl)+(0.75, 0.125)$);
    \draw [->] (clipL) to [out=270, in=180] ($(conditioning_bl)+(0.2, -0.25)$);

    \draw [fill=\clipGcolor] ($(pool_conditioning)+(0, 0.5)$) -|++ (0.2, 0.5) -|++ (-0.4, -0.5) -- cycle;
    \draw [fill=\clipLcolor] ($(pool_conditioning)+(0, 0.5)$) -|++ (0.2, -1) -|++ (-0.4, 1) -- cycle;
    \path ($(pool_conditioning)+(-0.2, -0.5)$) --++ (0, 1.5) node [midway, rotate=90, yshift=0.2cm] {Pooled};

    \draw [->] (clipG) to [out=270, in=90] ($(pool_conditioning)+(0, 0.75)$);
    \draw [->] (clipL) to [out=270, in=0] ($(pool_conditioning)+(0, 0.)$);

    \begin{pgfonlayer}{decorations}
        \path [decoration={brace, amplitude=8pt, raise=4pt}, decorate, draw] ($(conditioning_bl)+(0, 0.25)$) -- (conditioning_br) node [midway, yshift=0.7cm, fill=white] {$77 + 77$ tokens};
        \path [decoration={brace, amplitude=8pt, raise=4pt}, decorate, draw] (conditioning_br) -- (conditioning_tr) node [midway, xshift=1.0cm, align=center] {$4096$\\ channel};
    \end{pgfonlayer}

    \node [trainableblock, below=1cm of conditioning] (c_lin) {Linear};
    \node [tensor, below=0.5cm of c_lin] (c_in) {$c$};

    \node [trainableblock, below=1cm of pool_conditioning] (y_mlp) {MLP};

    \begin{pgfonlayer}{arrowlayer}
        \draw [->] (text.south) to [out=270, in=90] (clipG);
        \draw [->] (text.south) to [out=270, in=90] (clipL);
        \draw [->] (text.south) to [out=270, in=90] (t5);

        \draw [->] (conditioning) -- (c_in);
    \end{pgfonlayer}


    \node [trainableblock, below=2cm of y_mlp] (t_emb) {MLP};
    \node [fnblock, below=0.2cm of t_emb] (t_enc) {Sinusoidal Encoding};
    \node [input, below=0.5cm of t_enc] (timestep) {Timestep};

    \node [operation, below=0.5cm of y_mlp] (y_plus) {$+$};
    \node [tensor, right=1cm of y_plus] (y_final) {y};

    \begin{pgfonlayer}{arrowlayer}
        \draw [->] (timestep) -- (t_emb) -- (y_plus);
        \draw [->] (pool_conditioning) -- (y_plus);
        \draw [->] (y_plus) -- (y_final);
    \end{pgfonlayer}

    \node [input, below right=2.15cm and 2.5cm of text] (image) {Noised Latent};

    \node [fnblock, below=0.5cm of image] (patching) {Patching};
    \node [trainableblock, below=0.2cm of patching] (patch_proj) {Linear};
    \node [operation, below=0.5cm of patch_proj] (image_plus_pos) {$+$};
    \node [trainableblock, left=0.2cm of image_plus_pos,align=center] (pos_emb) {Positional\\ Embedding};

    \node [tensor, below=0.5cm of image_plus_pos] (x_in) {$x$};
    \begin{pgfonlayer}{arrowlayer}
        \draw [->] (image) -- (image_plus_pos);
        \draw [->] (pos_emb) -- (image_plus_pos);
        \draw [->] (image_plus_pos) -- (x_in);
    \end{pgfonlayer}

    \node [trainableblock, below=1.5cm of $(c_in)!0.5!(x_in)$, minimum width=5cm] (attn1) {\emph{MM-DiT}-Block 1};
    \node [trainableblock, below=0.3cm of attn1, minimum width=5cm] (attn2) {\emph{MM-DiT}-Block 2};
    \node [below=0.4cm of attn2, minimum width=5cm] {$\ldots$};
    \node [trainableblock, below=1.3cm of attn2, minimum width=5cm] (attnN) {\emph{MM-DiT}-Block $d$};

    \coordinate (attn1_mid_left) at ($(attn1.north west)!0.5!(attn1.north)$);
    \coordinate (attn1_mid_right) at ($(attn1.north east)!0.5!(attn1.north)$);

    \coordinate (attnN_mid_left) at ($(attnN.north west)!0.5!(attnN.north)$);
    \coordinate (attnN_mid_right) at ($(attnN.north east)!0.5!(attnN.north)$);

    \node [trainableblock, below=1.5cm of attnN_mid_right] (out_mod) {Modulation};
    \node [trainableblock, below=0.2cm of out_mod] (out_mlp) {Linear};
    \node [fnblock, below=0.2cm of out_mlp] (depatch) {Unpatching};
    \node [input, below=0.5cm of depatch] (output) {Output};

    \coordinate (attn1_left) at ($(attn1.west)+(-2, 0)$);

    \begin{pgfonlayer}{arrowlayer}
        \draw [->] (x_in) -| (attn1_mid_right) -- (output);
        \draw [->] (c_in) -| (attn1_mid_left) -- (attnN_mid_left);

        \draw [->] (y_final) -| (attn1_left) -- (attn1);
        \draw [->] (attn1_left) |- (attn2);
        \draw [->] (attn1_left) |- (attnN);
        \draw [->] (attn1_left) |- (out_mod);
    \end{pgfonlayer}

    \begin{pgfonlayer}{bg}
        \draw [fill=blue!5] ($(attn1.north east)+(1, 0.5)$) -| ($(attnN.south west)+(-1, -0.5)$) -| cycle;
    \end{pgfonlayer}

\end{tikzpicture}}
        \caption{Overview of all components.}
        \label{fig:modelfig:overview_mmdit_only}
    \end{subfigure}\hfill%
    \begin{subfigure}[t]{0.49\textwidth}\centering
        \resizebox{!}{0.46\textheight}{\input{tikz/mm_block2}}
        \caption{One \modelname block}
        \label{fig:modelfig:mm}
    \end{subfigure}
    \caption{\textbf{Our model architecture.} Concatenation is indicated by $\odot$ and element-wise multiplication by $*$. The RMS-Norm for $Q$ and $K$ can be added to stabilize training runs. Best viewed zoomed in. \vspace{-1em}}
    \label{fig:modelfig:total}
\end{figure*}
}

\newcommand{\textmanipulation}{%
\begin{figure*}[h]
\centering
\begin{tabular}{@{\hspace{0\tabcolsep}}c@{\hspace{0.2\tabcolsep}}c@{\hspace{0.2\tabcolsep}}c@{\hspace{0.2\tabcolsep}}c}
& Input & Output 1 & Output 2\\
{
\begin{tabular}[x]{@{}c@{}} \small Write "go small \\ \small go home"  \\ \small instead \end{tabular}}
 & 
\raisebox{-.5\height}{
\includegraphics[width=0.25\linewidth]{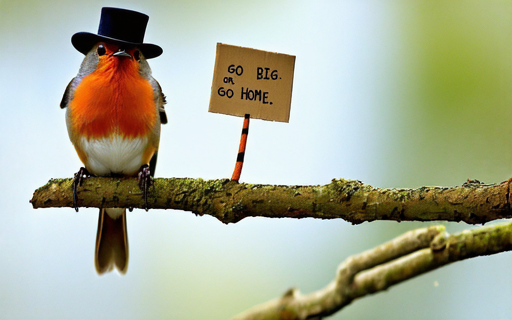}}&
\raisebox{-.5\height}{
\includegraphics[width=0.25\linewidth]{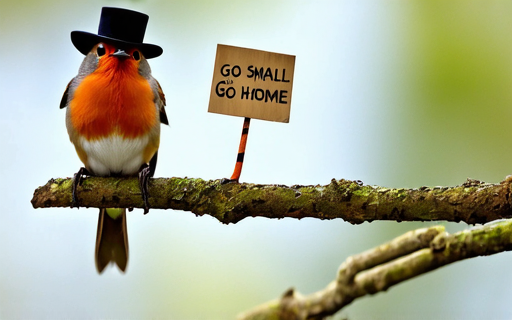}}&
\raisebox{-.5\height}{
\includegraphics[width=0.25\linewidth]{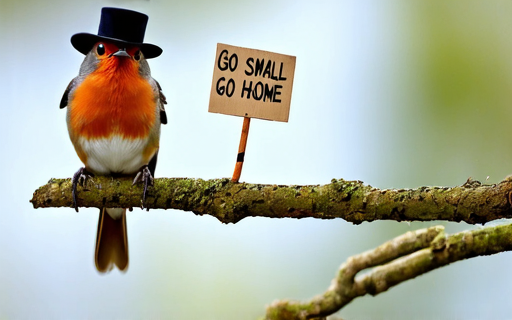}}\\
{
\begin{tabular}[x]{@{}c@{}} \small GO BIG OR GO UNET \\ \small is written on \\ \small the blackboard \end{tabular}}&
\raisebox{-.5\height}{
\includegraphics[width=0.25\linewidth]{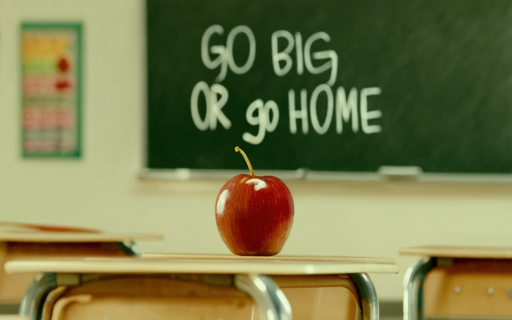}}&
\raisebox{-.5\height}{
\includegraphics[width=0.25\linewidth]{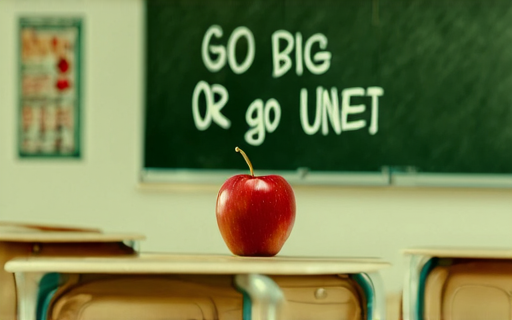}}
&
\raisebox{-.5\height}{
\includegraphics[width=0.25\linewidth]{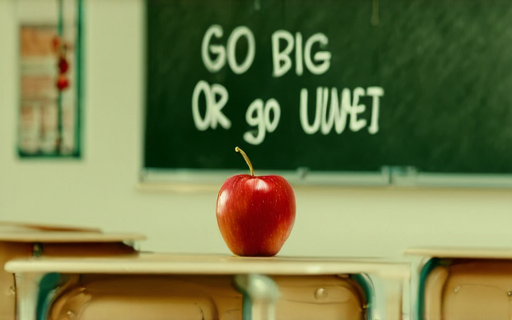}}\\
{
\begin{tabular}[x]{@{}c@{}} \small change the \\ \small word to \\ \small UNOT \end{tabular}}&
\raisebox{-.5\height}{
\includegraphics[width=0.25\linewidth]{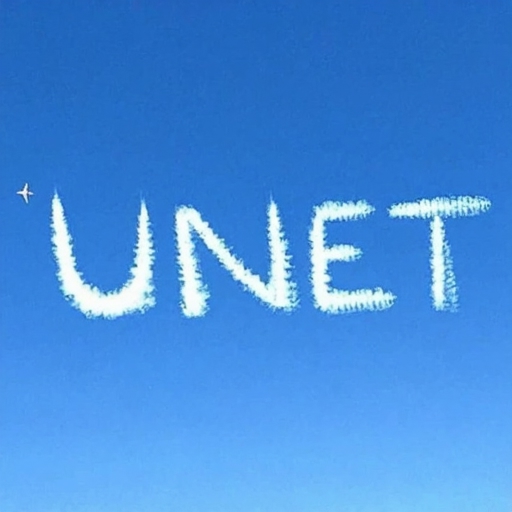}}&
\raisebox{-.5\height}{
\includegraphics[width=0.25\linewidth]{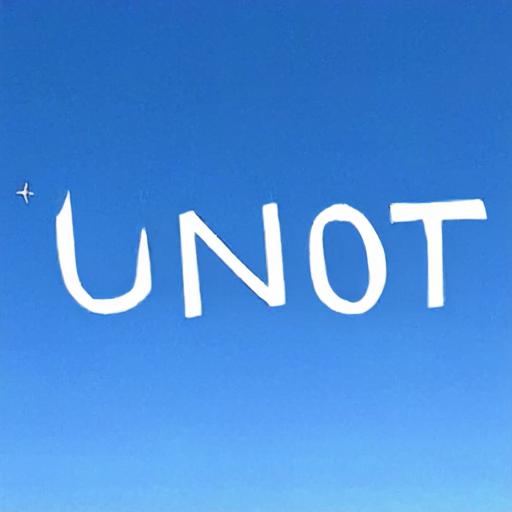}}&
\raisebox{-.5\height}{
\includegraphics[width=0.25\linewidth]{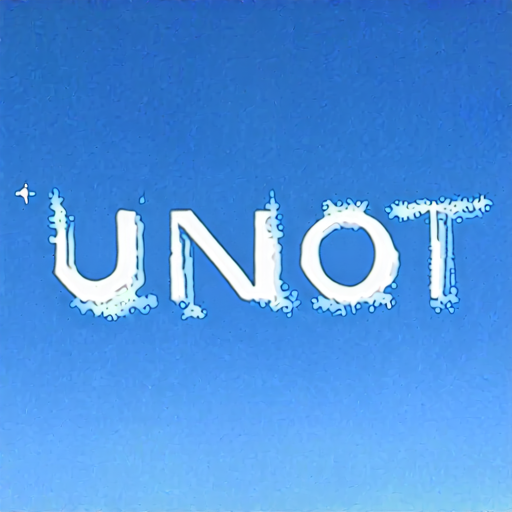}} \\
{
\begin{tabular}[x]{@{}c@{}} \small make the  \\ \small sign say \\ \small MMDIT rules \end{tabular}}&
\raisebox{-.5\height}{
\includegraphics[width=0.25\linewidth]{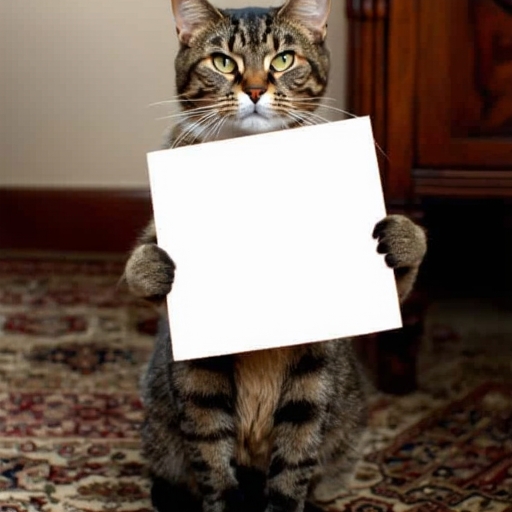}}&
\raisebox{-.5\height}{
\includegraphics[width=0.25\linewidth]{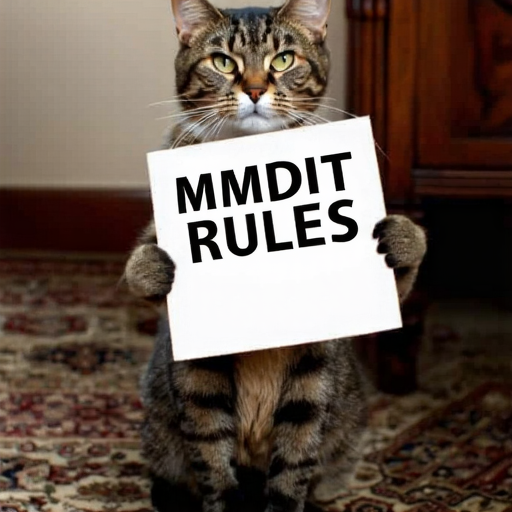}}&
\raisebox{-.5\height}{
\includegraphics[width=0.25\linewidth]{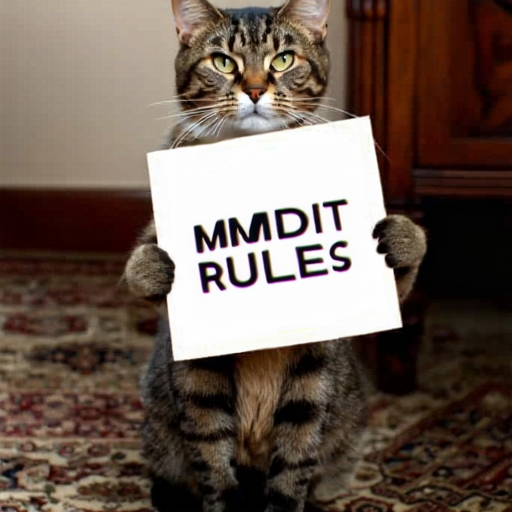}}
\end{tabular}
\caption{Zero Shot Text manipulation and insertion with the 2B Edit model}
\label{fig:textmanipulation} %
\end{figure*}
}

\newcommand{\gobig}{
\begin{figure*}
\includegraphics[width=0.99\textwidth]{img/samples/grid01.jpg}
\caption{\label{fig:gobig} More 768 $\times$ 1344 samples from our 8B rectified flow model (w/o DPO training).}
\end{figure*}
}

\newcommand{\maxattnlogitqk}{
\begin{figure}
\includegraphics[width=0.49\linewidth,valign=m]{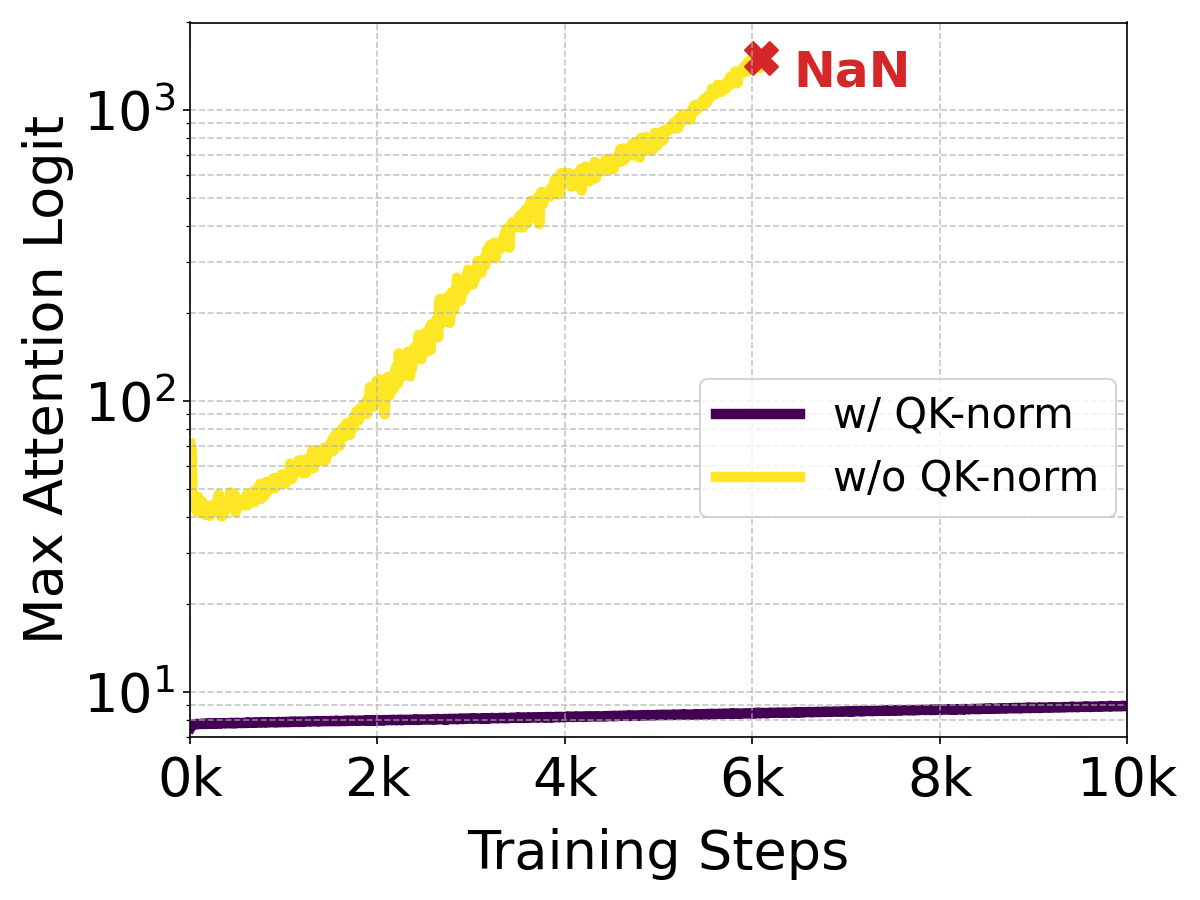}
\includegraphics[width=0.49\linewidth,valign=m]{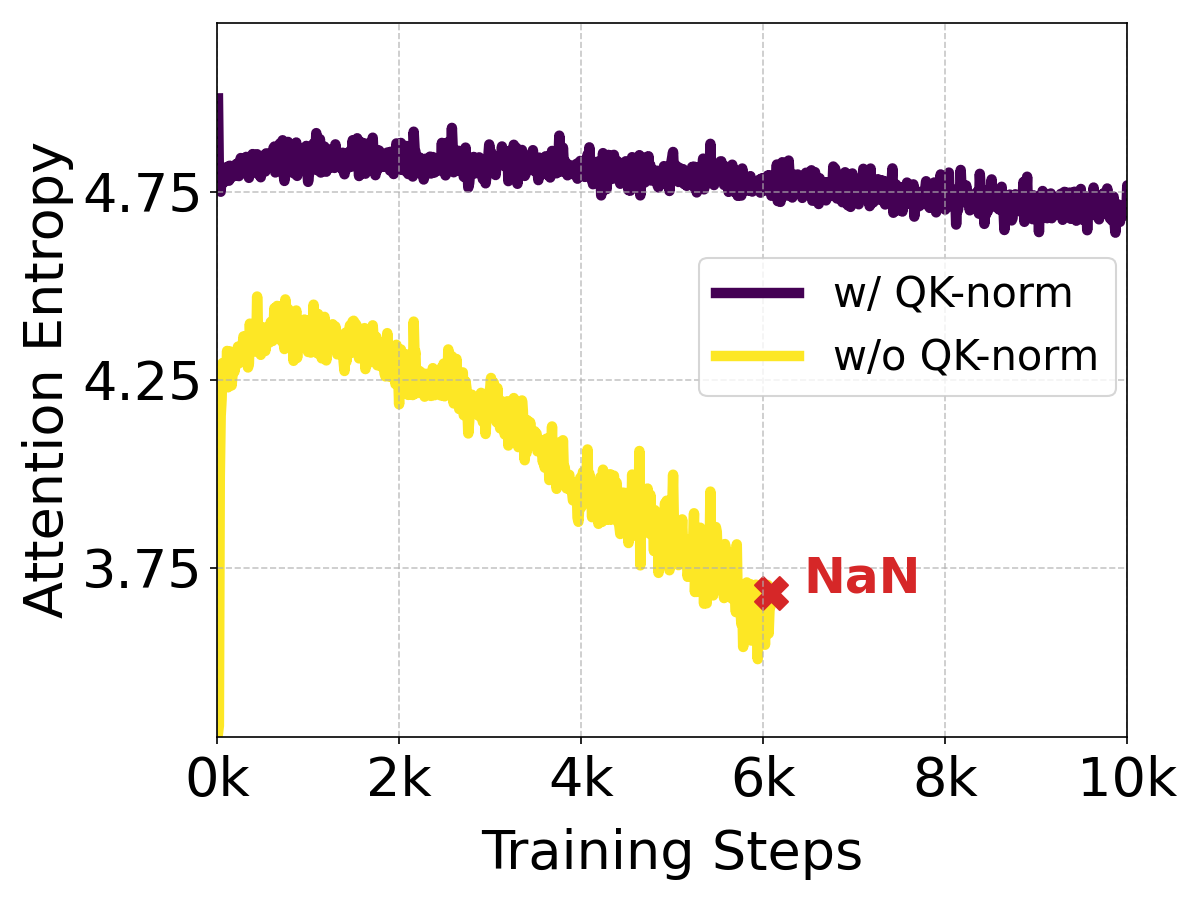}

\caption{\label{fig:qk_vis} \textbf{Effects of QK-normalization.} Normalizing the Q- and K-embeddings before calculating the attention matrix prevents the attention-logit growth instability (\emph{left}), which causes the attention entropy to collapse (\emph{right}) and has been previously reported in the discriminative ViT literature~\cite{dehghani2023scaling,wortsman2023smallscale}. In contrast with these previous works, we observe this instability in the last transformer blocks of our networks. Maximum attention logits and attention entropies are shown averaged over the last 5 blocks of a 2B (d=24) model.}
\end{figure}
}

\newcommand{\figtimeshift}{
\begin{figure}
\includegraphics[width=1.0\linewidth,valign=m]{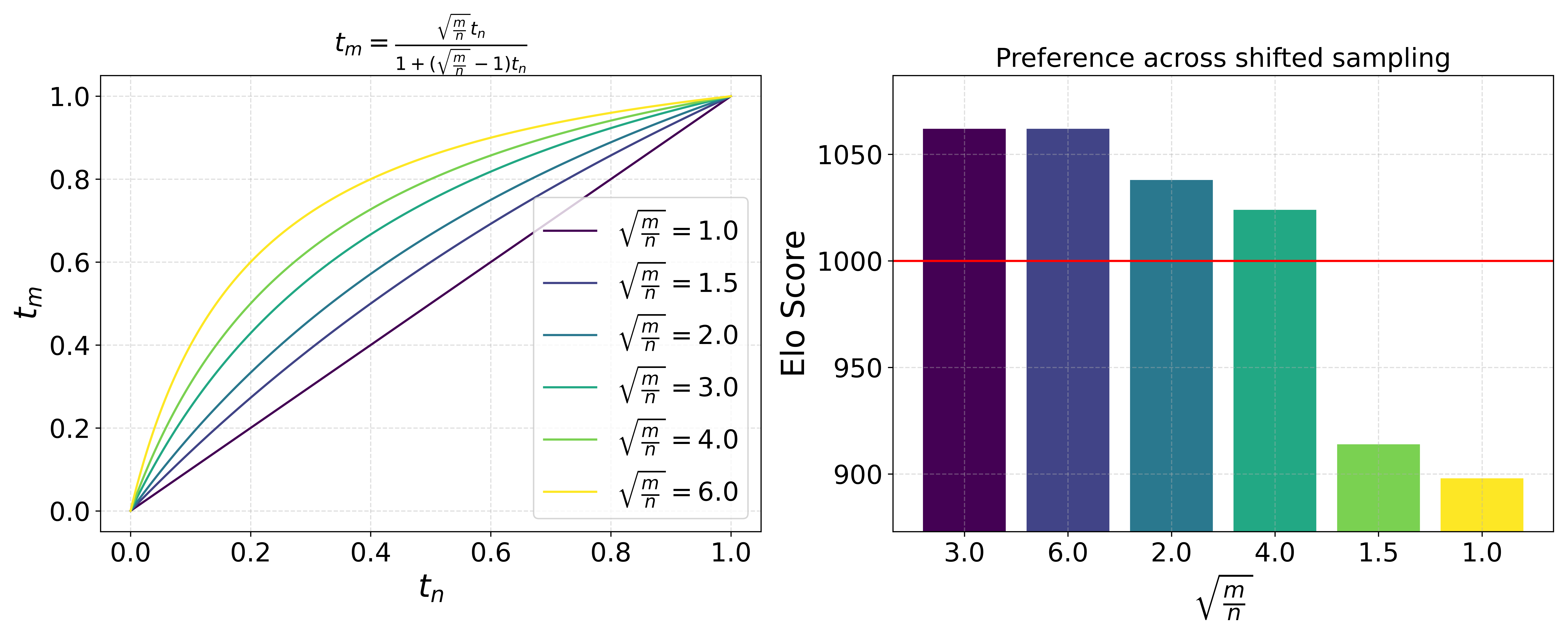} \\
\includegraphics[width=1.0\linewidth,valign=m]{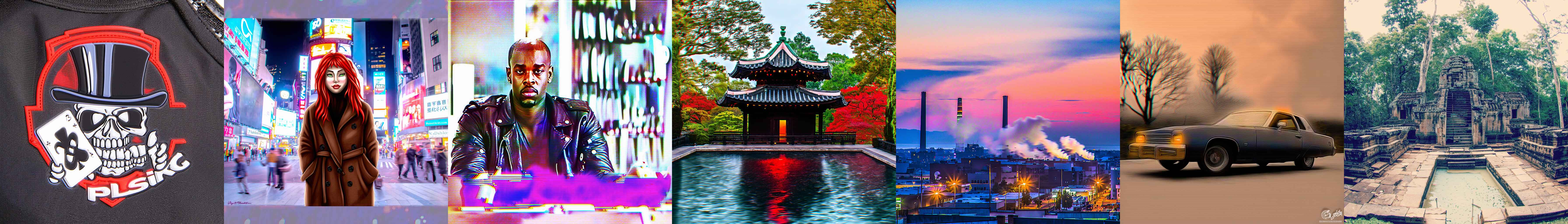} \\
\includegraphics[width=1.0\linewidth,valign=m]{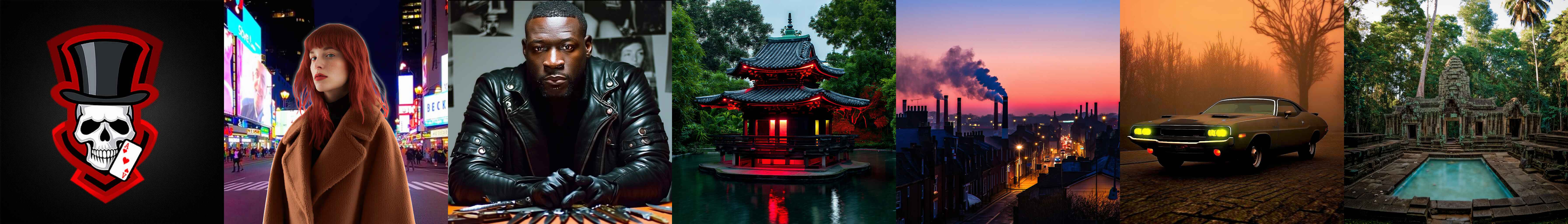}
\caption{\label{fig:timeshift}\textbf{Timestep shifting at higher resolutions.} 
\emph{Top right:} Human quality preference rating when applying the shifting based on \Cref{eq:timeshift}. 
\emph{Bottom row:} A $512^2$ model trained and sampled with $\sqrt{m/n} = 1.0$ (\emph{top}) and $\sqrt{m/n} = 3.0$ (\emph{bottom}).
See \Cref{subsec:shifting}.}
\end{figure}
}

\newcommand{\horizontalcherries}{
\begin{figure*}[htp]
    \centering
    \begin{minipage}[t]{0.1042\linewidth}
        \centering
        \includegraphics[width=1.0\linewidth]{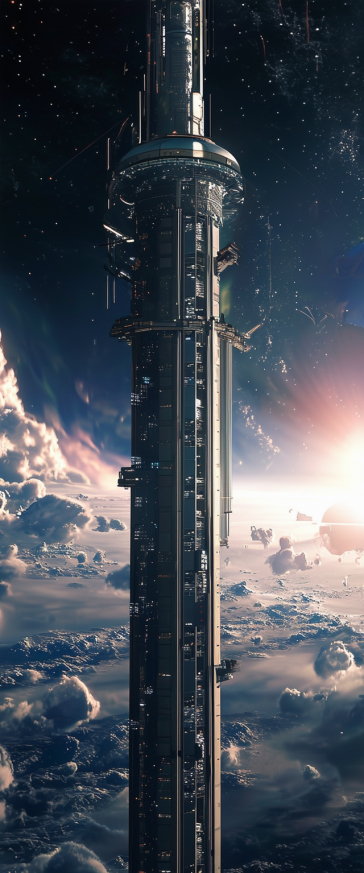} \\
        \tiny{a space elevator, cinematic scifi art}
    \end{minipage}
    \begin{minipage}[t]{0.1425\linewidth}
        \centering
        \includegraphics[width=1.0\linewidth]{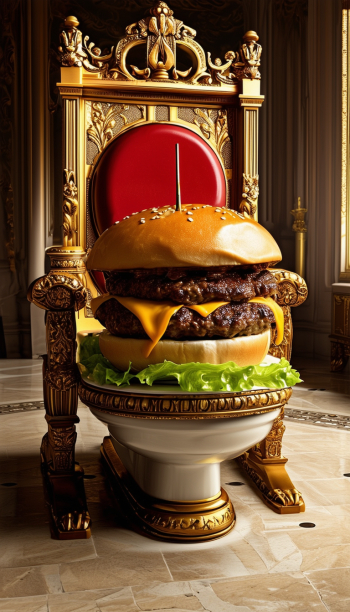} \\
        \tiny{A cheeseburger with juicy beef patties and melted cheese sits on top of a toilet that looks like a throne and stands in the middle of the royal chamber.}
    \end{minipage}
    \begin{minipage}[t]{0.1425\linewidth}
        \centering
        \includegraphics[width=1.0\linewidth]{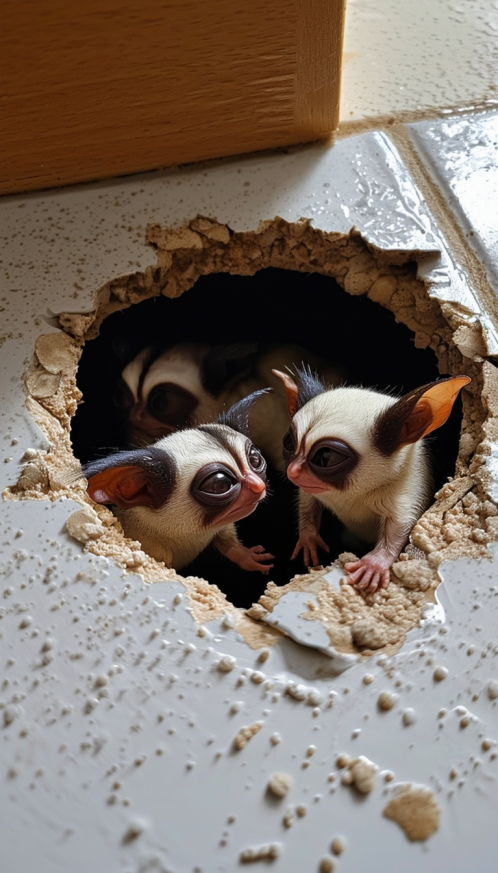} \\
        \tiny{a hole in the floor of my bathroom with small gremlins living in it}
    \end{minipage}
    \begin{minipage}[t]{0.1425\linewidth}
        \centering
        \includegraphics[width=1.0\linewidth]{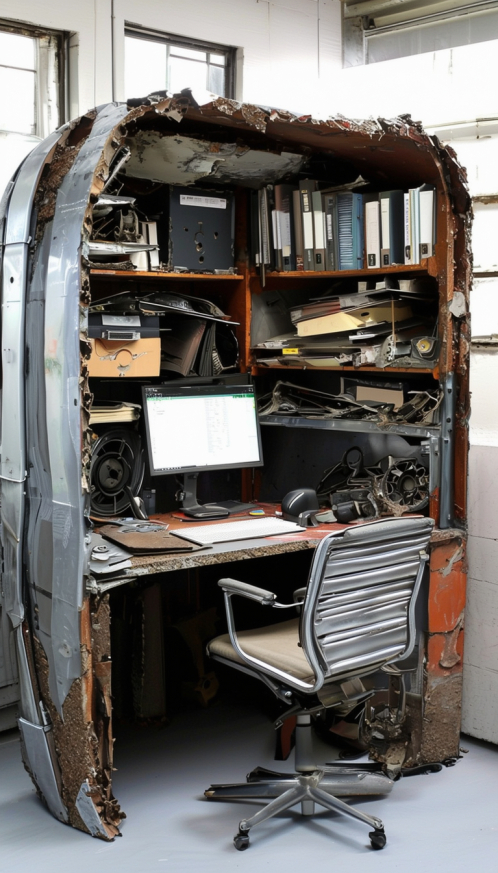} \\
        \tiny{a small office made out of car parts}
    \end{minipage}
    \begin{minipage}[t]{0.1425\linewidth}
        \centering
        \includegraphics[width=1.0\linewidth]{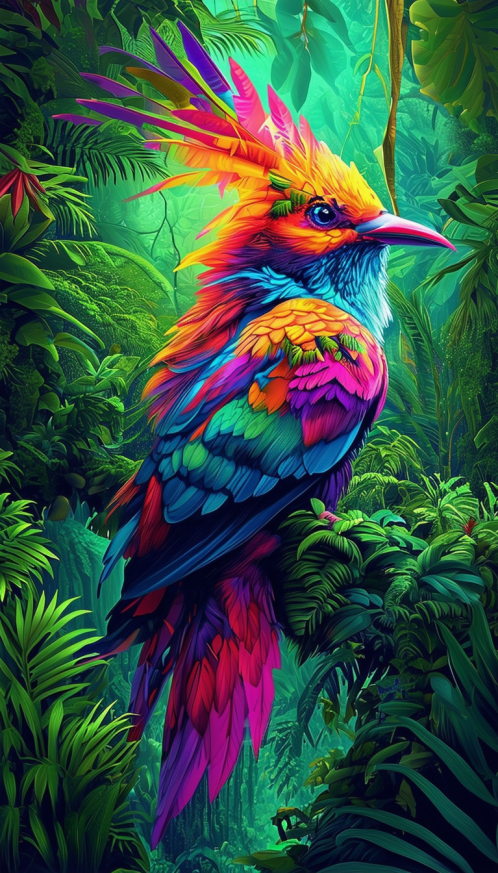} \\
        \tiny{This dreamlike digital art captures a vibrant, kaleidoscopic bird in a lush rainforest.}
    \end{minipage}
    \begin{minipage}[t]{0.1425\linewidth}
        \centering
        \includegraphics[width=1.0\linewidth]{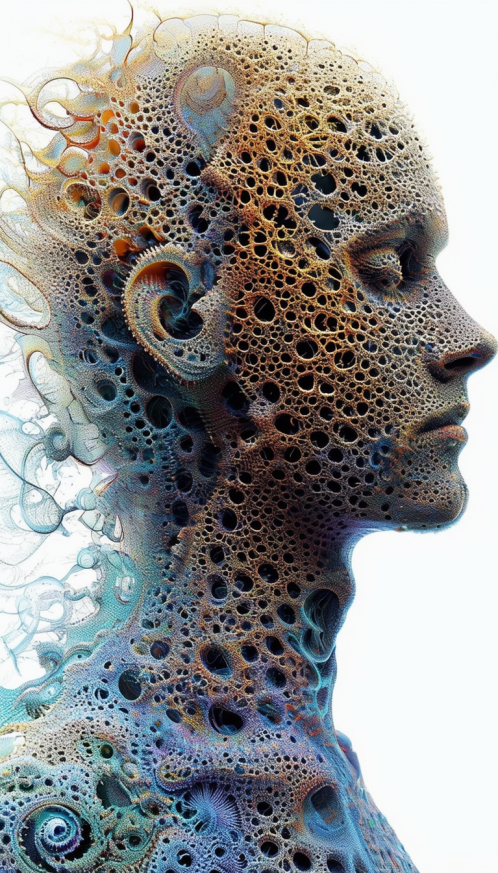} \\
        \tiny{human life depicted entirely out of fractals}
    \end{minipage}
    \begin{minipage}[t]{0.1042\linewidth}
        \centering
        \includegraphics[width=1.0\linewidth]{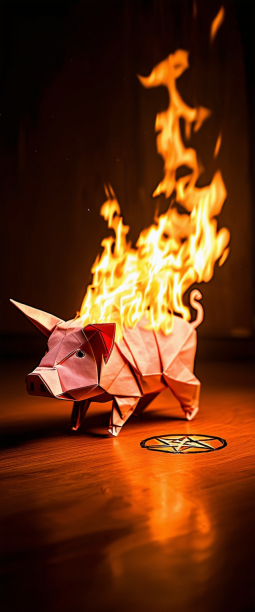} \\
        \tiny{an origami pig on fire in the middle of a dark room with a pentagram on the floor}
    \end{minipage}
    \hfill
    \begin{minipage}[t]{0.495\linewidth}
        \centering
        \includegraphics[width=1.0\linewidth]{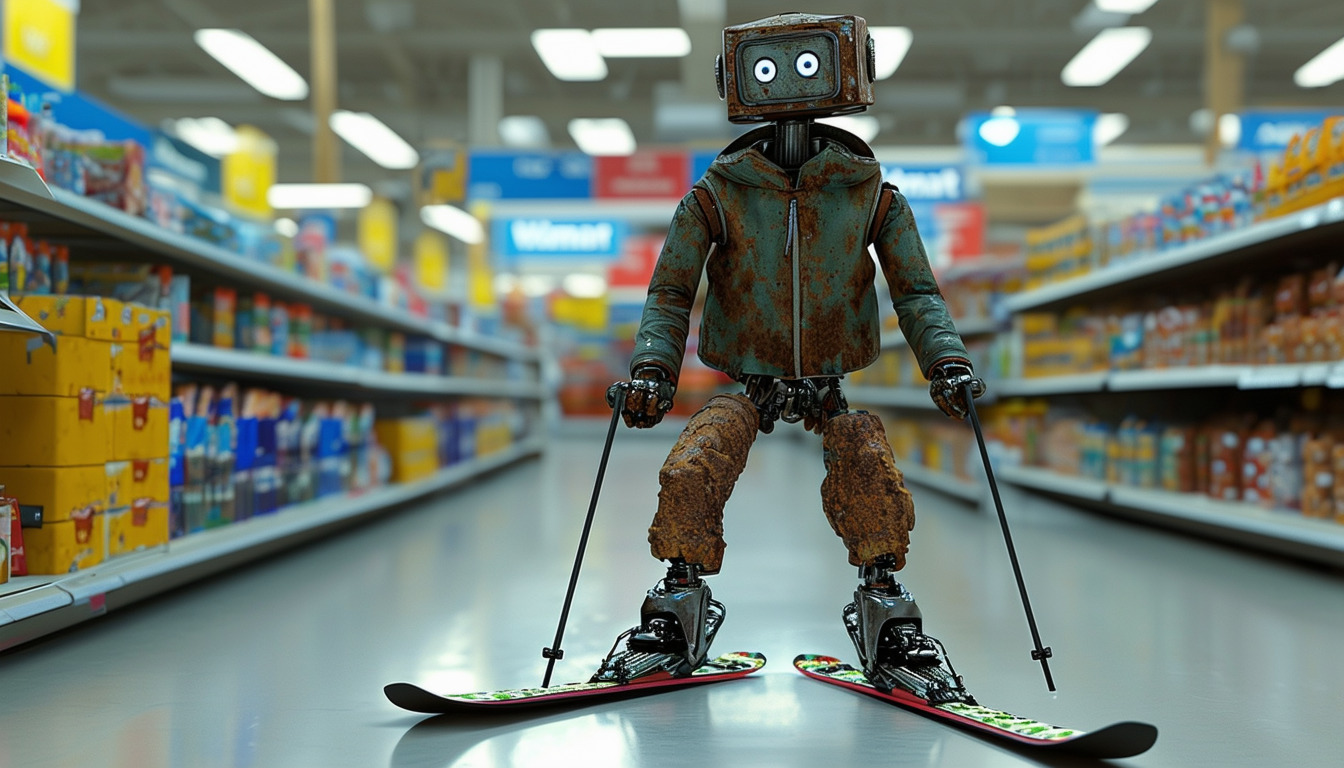} \\
        \tiny{an old rusted robot wearing pants and a jacket riding skis in a supermarket.}
    \end{minipage}
    \hfill
    \begin{minipage}[t]{0.495\linewidth}
        \centering
        \includegraphics[width=1.0\linewidth]{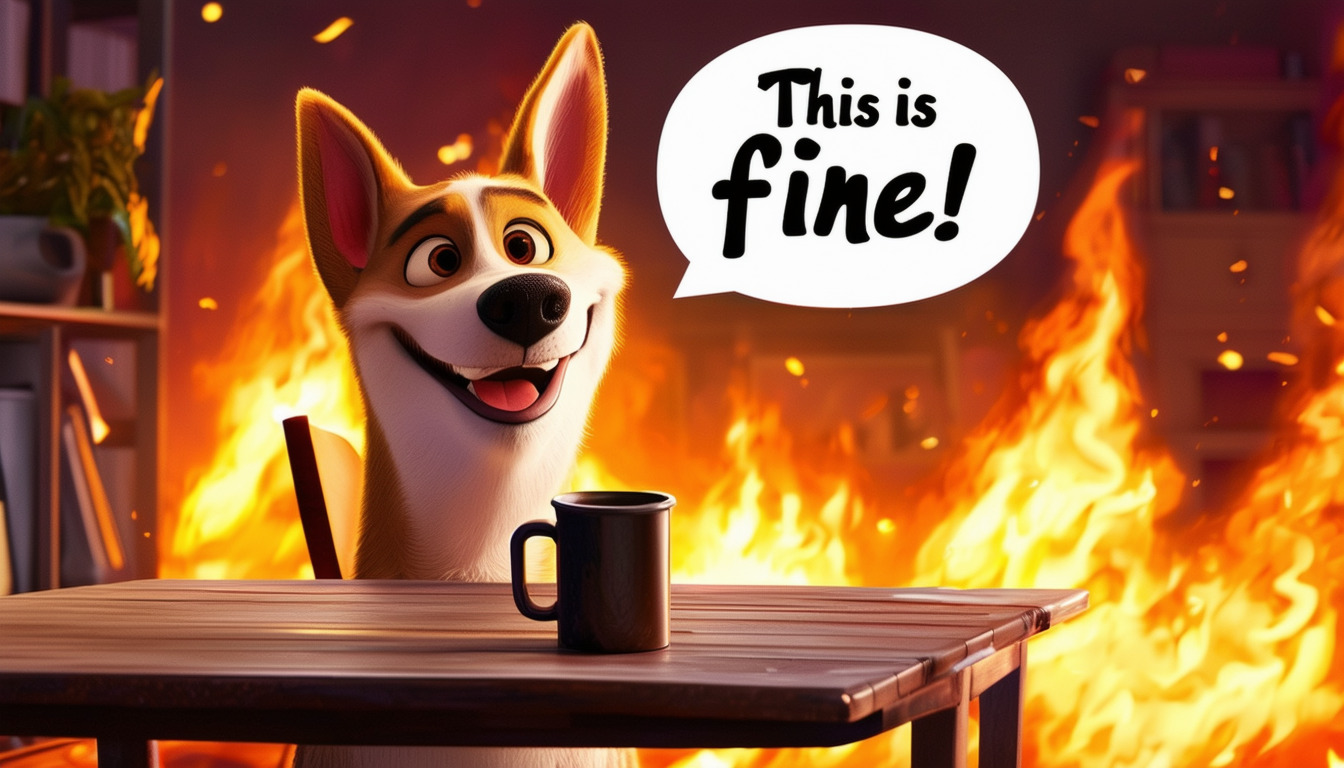} \\
        \tiny{smiling cartoon dog sits at a table, coffee mug on hand, as a room goes up in flames. ``This is fine,'' the dog assures himself.}
    \end{minipage}
    \hfill
    \vspace{1em}
    \begin{minipage}{1.0\linewidth}
        \centering
        \includegraphics[width=1.0\linewidth]{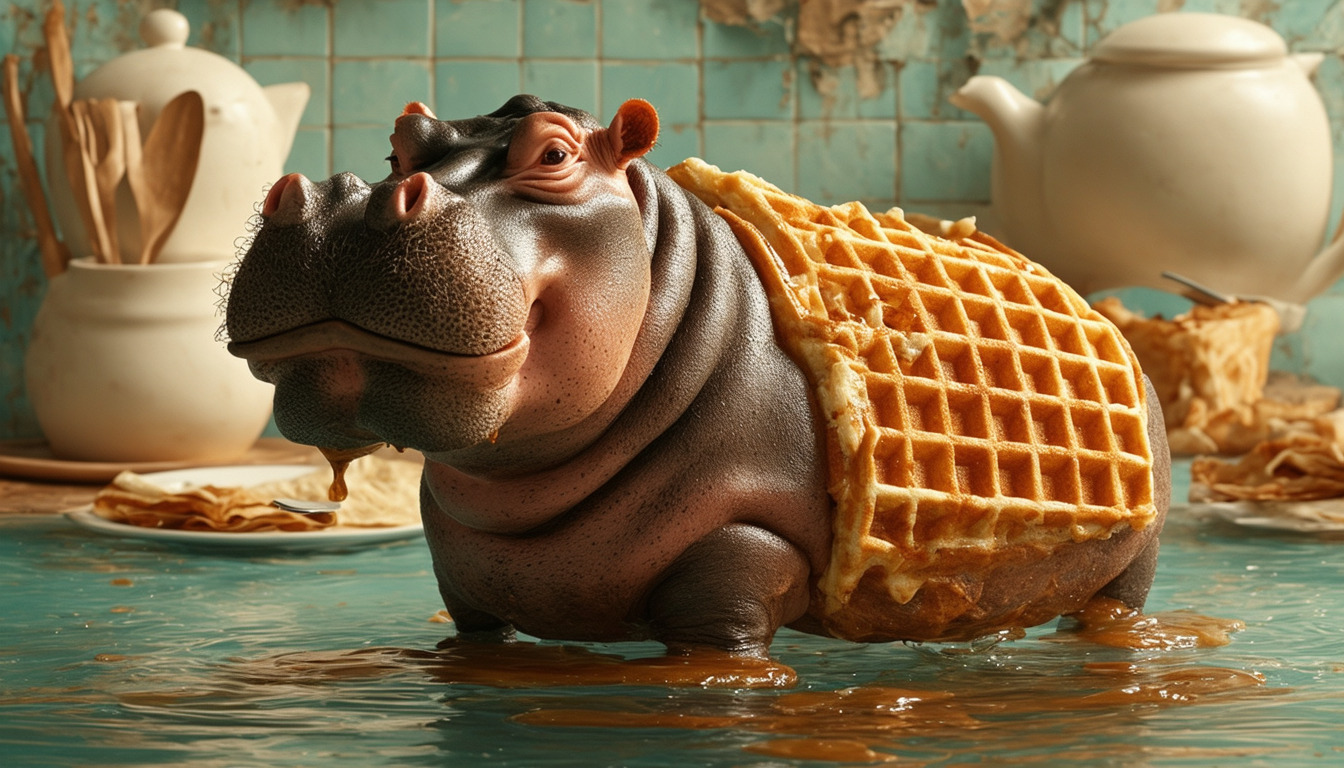} 
        \tiny{A whimsical and creative image depicting a hybrid creature that is a mix of a waffle and a hippopotamus. This imaginative creature features the distinctive, bulky body of a hippo, but with a texture and appearance resembling a golden-brown, crispy waffle. The creature might have elements like waffle squares across its skin and a syrup-like sheen. It's set in a surreal environment that playfully combines a natural water habitat of a hippo with elements of a breakfast table setting, possibly including oversized utensils or plates in the background. The image should evoke a sense of playful absurdity and culinary fantasy.}
    \end{minipage}
\end{figure*}
}

\newcommand{\morehorizontalcherries}{
\begin{figure*}[htp]
    \centering
    \begin{minipage}{0.495\linewidth}
        \centering
        \includegraphics[width=1.0\linewidth]{img/samples/001/sd3magazine.jpg} \\
        \captionof{figure}{\emph{the words ``STABLE DIFFUSION 3'' composed of an abstract collage of torn + glued magazine clippings from 1970’s era magazine, arranged in the shape of the words, white background.}}
        \label{fig:sample4}
    \end{minipage}
    \hfill
    \begin{minipage}{0.495\linewidth}
        \centering
        \includegraphics[width=1.0\linewidth]{img/samples/001/pigdragon.png} \\
        \captionof{figure}{\emph{epic fantasy art of a massive pig dragon destroying a village at night. \\ \phantom{abg}}}
        \label{fig:sample5}
    \end{minipage}
    \hfill
\vspace{1em}
    \begin{minipage}{1.0\linewidth}
        \centering
        \includegraphics[width=1.0\linewidth]{img/samples/001/highresdog.jpg} 
        \captionof{figure}{\emph{a cute dog wearing sunglasses on a boat in a lake in the mountains at dawn.}}
        \label{fig:sample6}
    \end{minipage}
\end{figure*}
}

\newcommand{\horizontalcherriesthree}{
\begin{figure*}[htp]
    \begin{minipage}[t]{0.495\linewidth}
        \centering
        \includegraphics[width=1.0\linewidth]{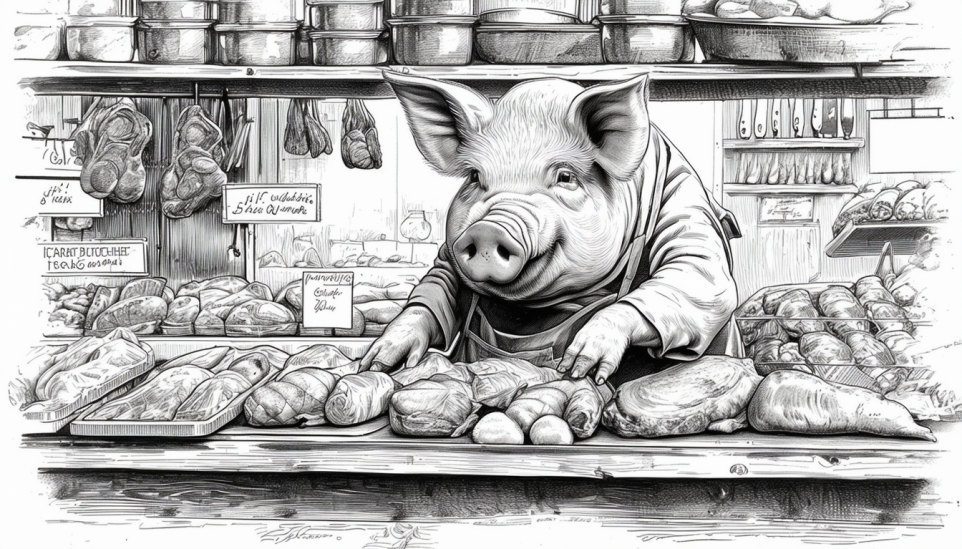} \\
        \tiny{Detailed pen and ink drawing of a happy pig butcher selling meat in its shop.}
    \end{minipage}
    \hfill
    \begin{minipage}[t]{0.495\linewidth}
        \centering
        \includegraphics[width=1.0\linewidth]{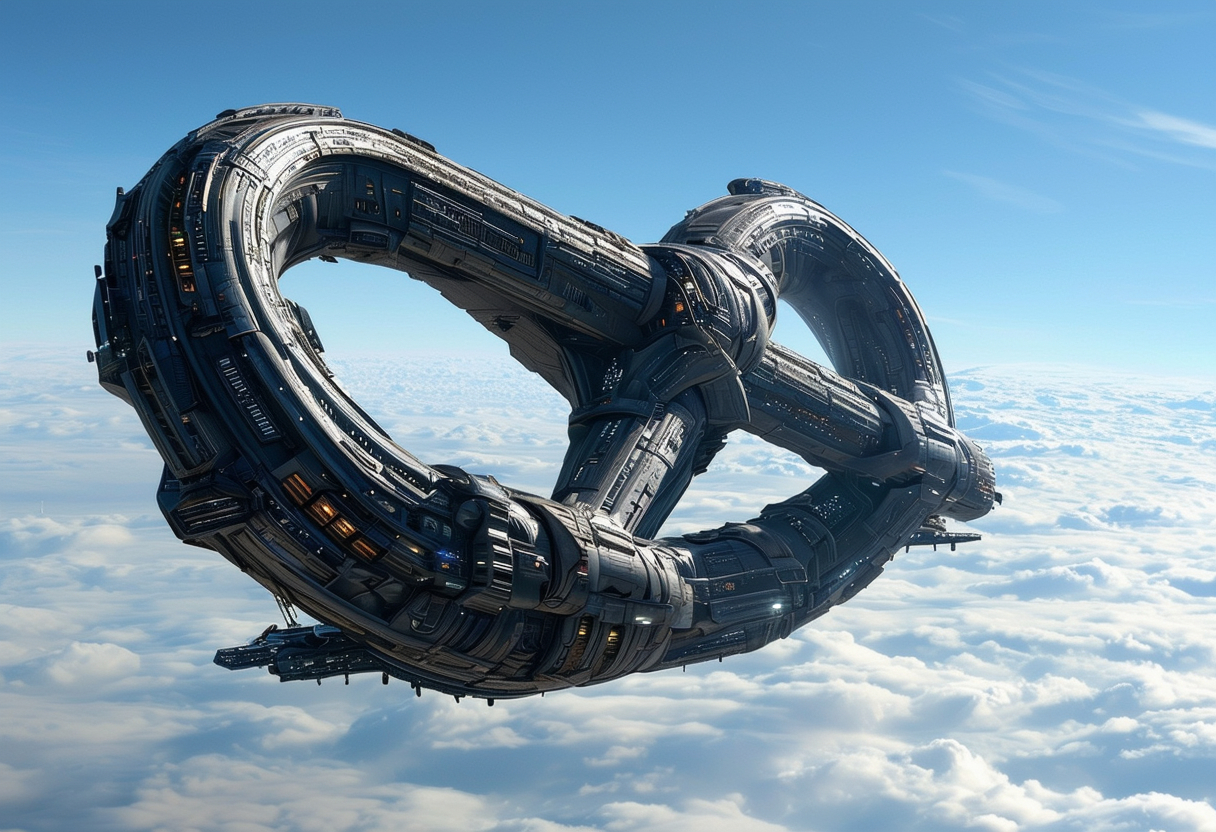} \\
        \tiny{a massive alien space ship that is shaped like a pretzel.}
    \end{minipage}
    \hfill
    \vspace{0.5em}
    \centering
    \hspace{0.5em}
    \begin{minipage}[t]{0.1425\linewidth}
        \centering
        \includegraphics[width=1.0\linewidth]{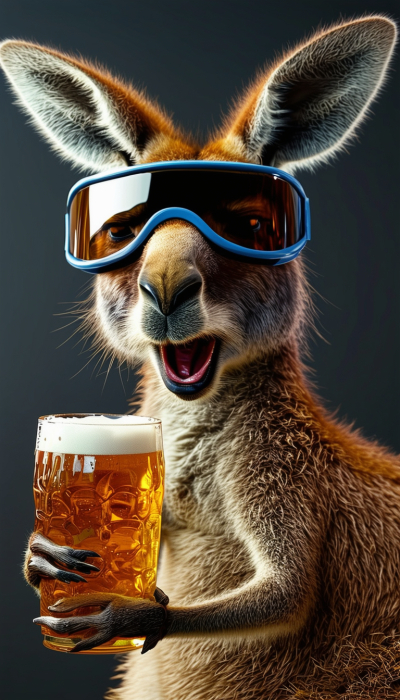} \\
        \tiny{A kangaroo holding a beer, wearing ski goggles and passionately singing silly songs.}
    \end{minipage}
    \hfill
    \centering
    \begin{minipage}[t]{0.1425\linewidth}
        \centering
        \includegraphics[width=1.0\linewidth]{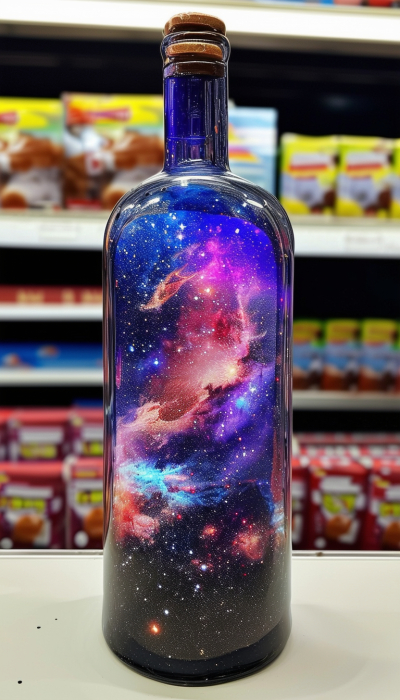} \\
        \tiny{An entire universe inside a bottle sitting on the shelf at walmart on sale.}
    \end{minipage}
    \hfill
    \centering
    \begin{minipage}[t]{0.1425\linewidth}
        \centering
        \includegraphics[width=1.0\linewidth]{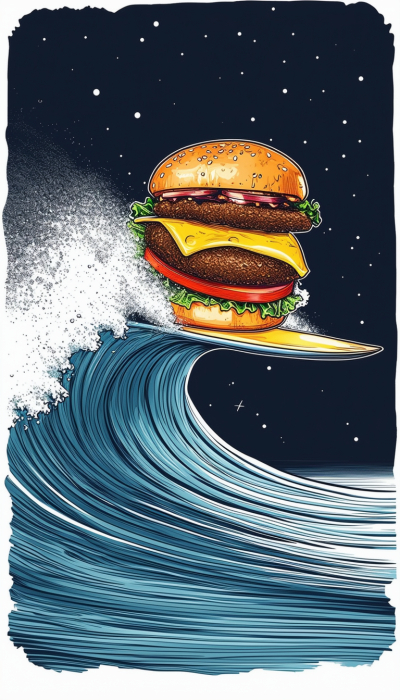} \\
        \tiny{A cheesburger surfing the vibe wave at night}
    \end{minipage}
    \hfill
    \centering
    \begin{minipage}[t]{0.1425\linewidth}
        \centering
        \includegraphics[width=1.0\linewidth]{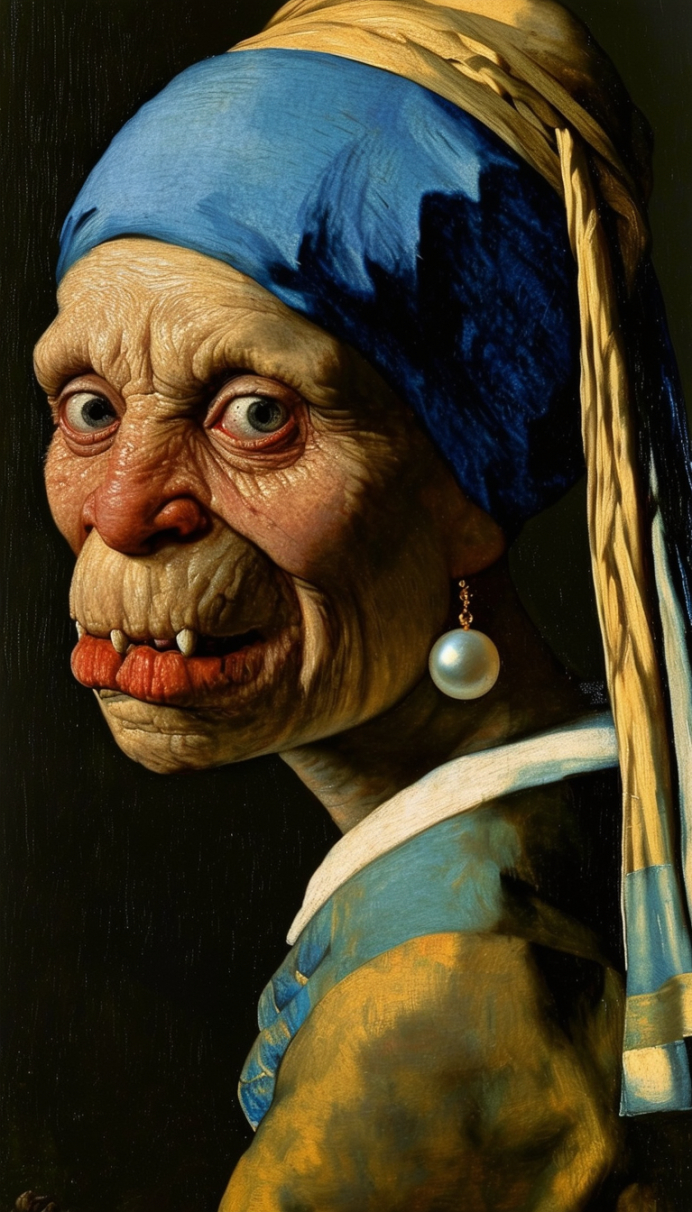} \\
        \tiny{A swamp ogre with a pearl earring by Johannes Vermeer}
    \end{minipage}
    \hfill
    \centering
    \begin{minipage}[t]{0.1425\linewidth}
        \centering
        \includegraphics[width=1.0\linewidth]{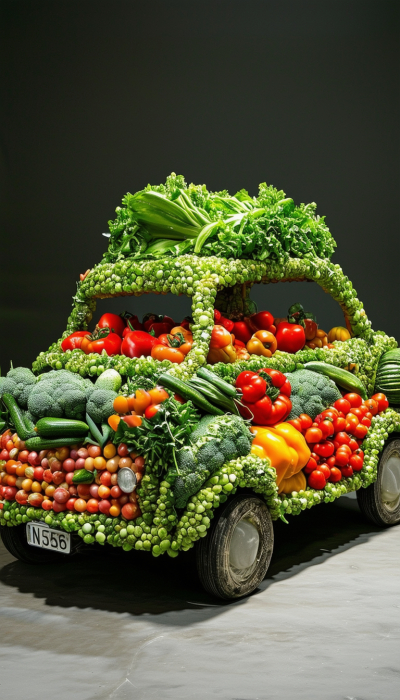} \\
        \tiny{A car made out of vegetables.}
    \end{minipage}
    \hfill
    \centering
    \begin{minipage}[t]{0.1425\linewidth}
        \centering
        \includegraphics[width=1.0\linewidth]{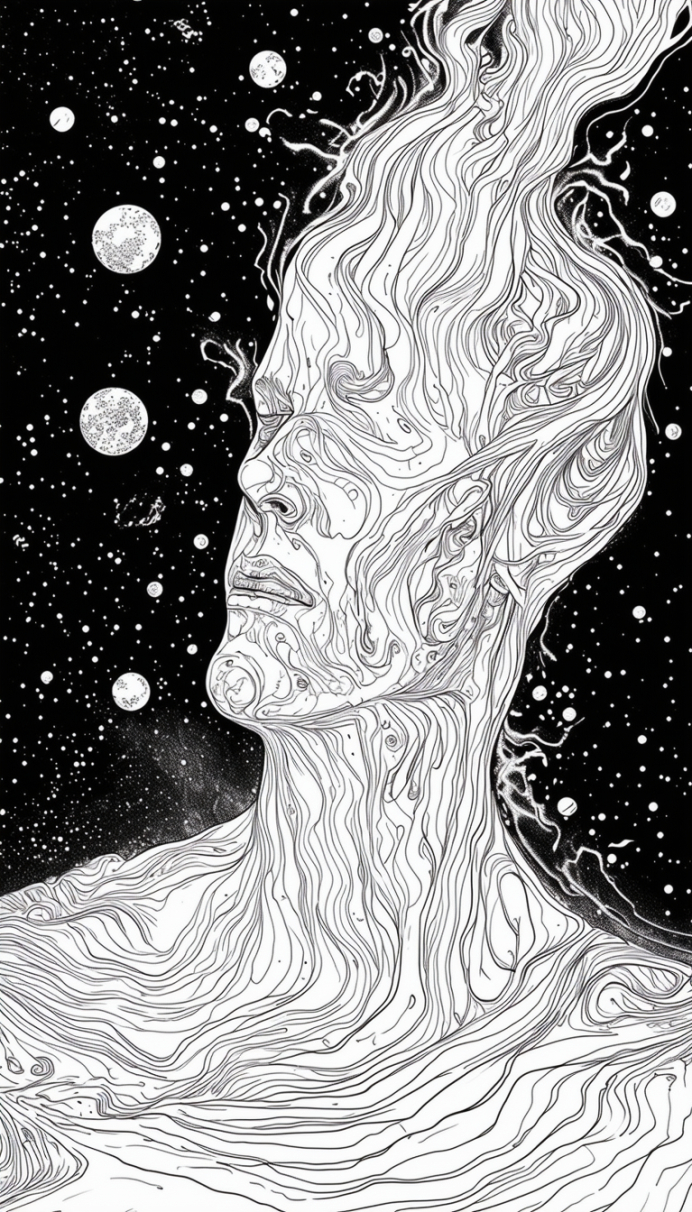} \\
        \tiny{heat death of the universe, line art}
    \end{minipage}
    \hfill
    \vspace{0.5em}
    \begin{minipage}[t]{0.495\linewidth}
        \centering
        \includegraphics[width=1.0\linewidth]{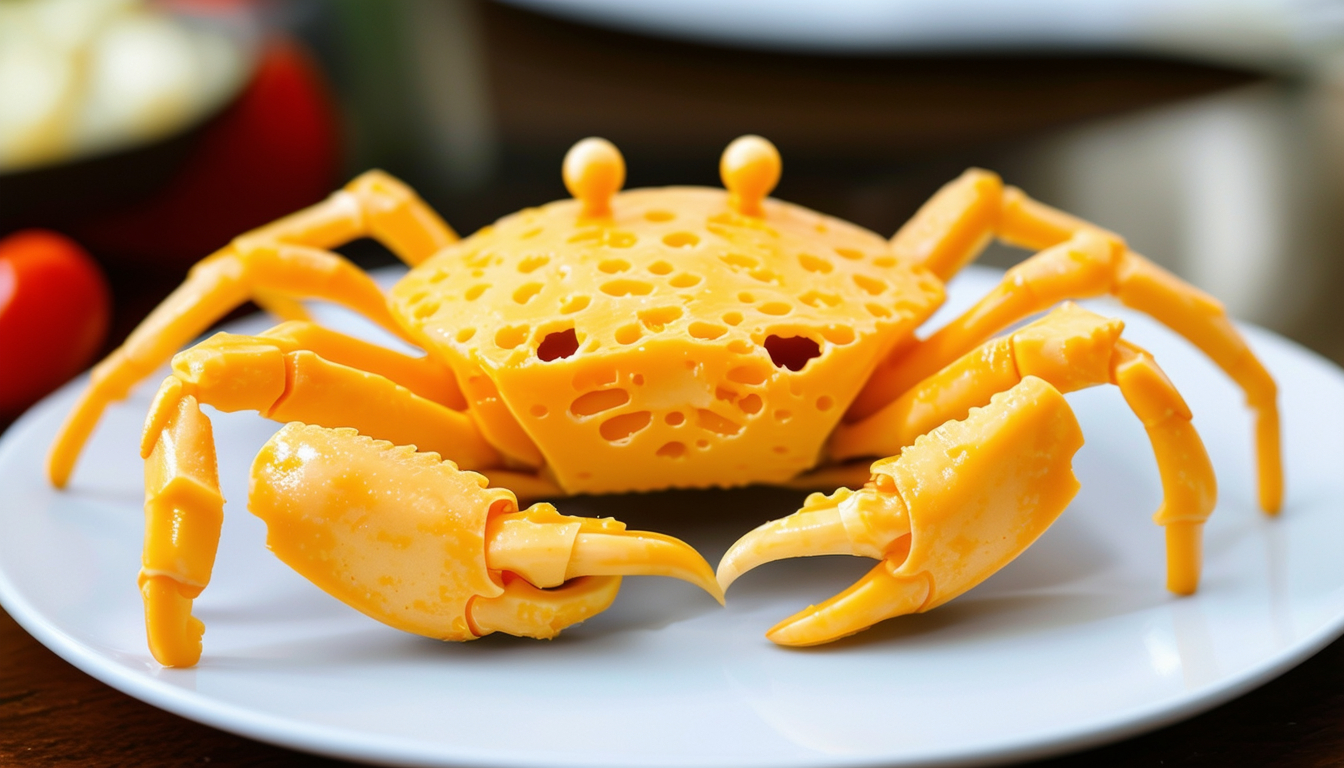} \\
        \tiny{A crab made of cheese on a plate}
    \end{minipage}
    \hfill
    \begin{minipage}[t]{0.495\linewidth}
        \centering
        \includegraphics[width=1.0\linewidth]{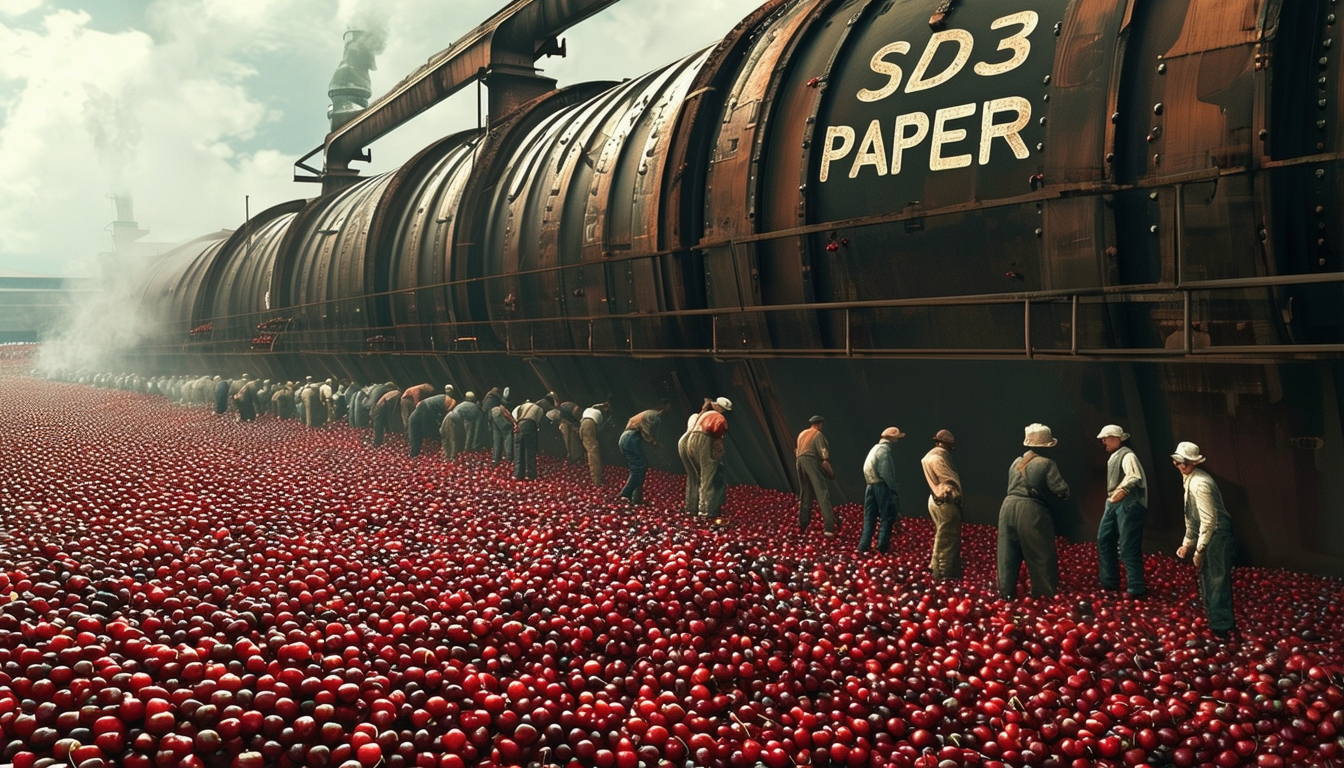} \\
        \tiny{Dystopia of thousand of workers picking cherries and feeding them into a machine that runs on steam and  is as large as a skyscraper. Written on the side of the machine: "SD3 Paper"}
    \end{minipage}
    \hfill
    \vspace{0.5em}
    \begin{minipage}[t]{0.495\linewidth}
        \centering
        \includegraphics[width=1.0\linewidth]{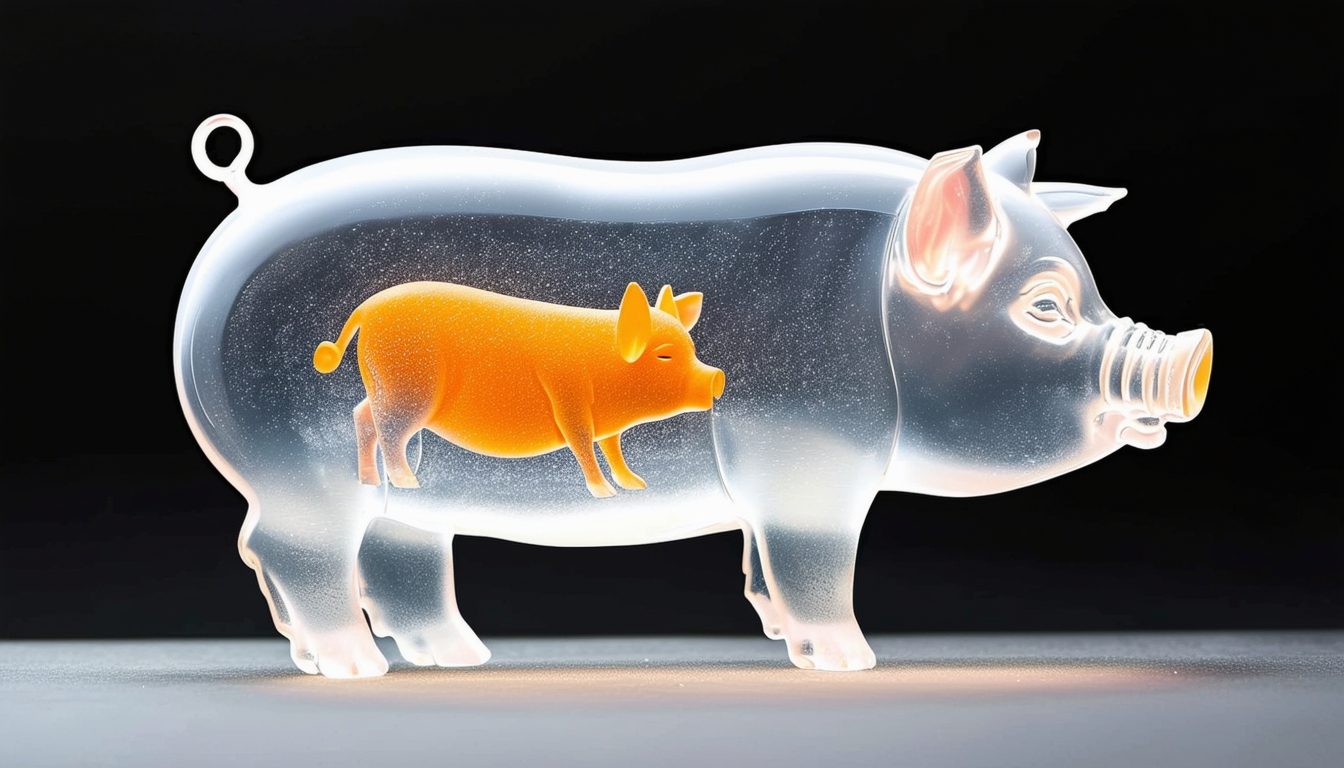} \\
        \tiny{translucent pig, inside is a smaller pig.}
    \end{minipage}
    \hfill
    \begin{minipage}[t]{0.495\linewidth}
        \centering
        \includegraphics[width=1.0\linewidth]{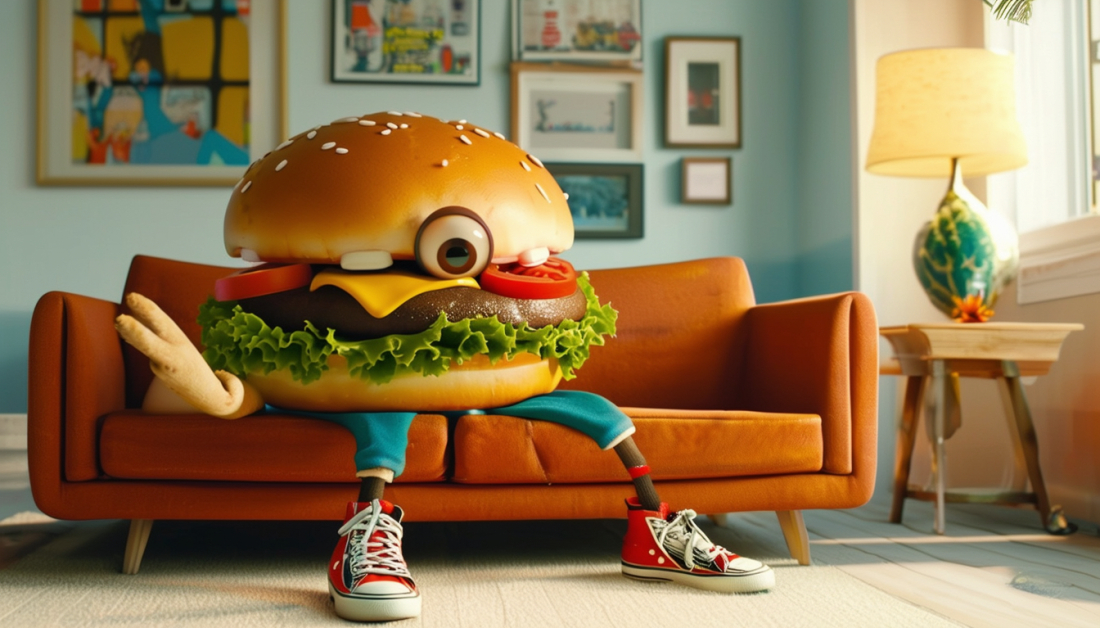} \\
        \tiny{Film still of a long-legged cute big-eye anthropomorphic cheeseburger wearing sneakers relaxing on the couch in a sparsely decorated living room.}
    \end{minipage}
\end{figure*}
}

\newcommand{\horizontalcherriesgrids}{
\begin{figure*}[htp]
    \begin{minipage}[t]{0.495\linewidth}
        \centering
        \includegraphics[width=1.0\linewidth]{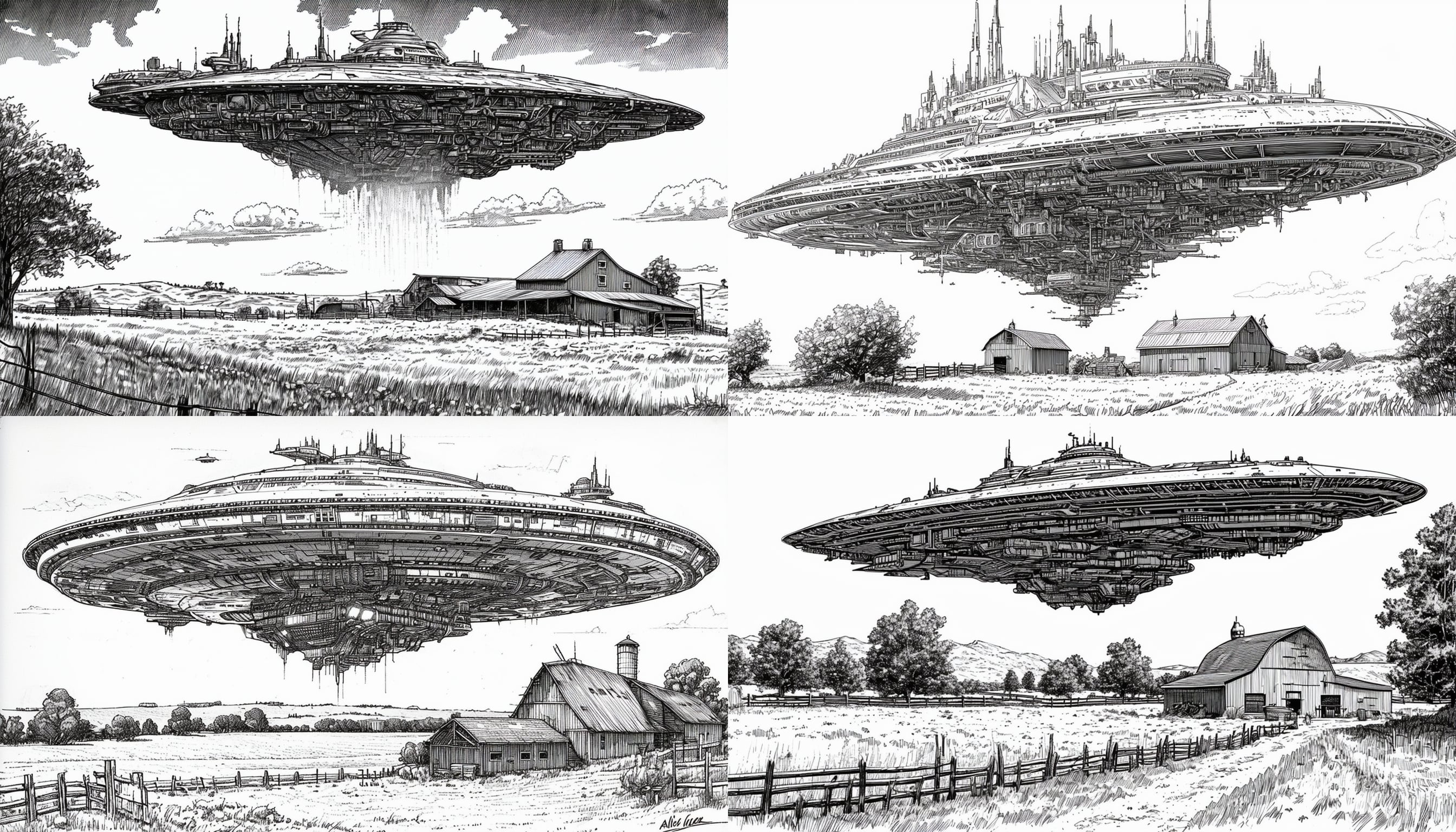} \\
        \tiny{detailed pen and ink drawing of a massive complex alien space ship above a farm in the middle of nowhere.}
    \end{minipage}
    \hfill
    \begin{minipage}[t]{0.495\linewidth}
        \centering
        \includegraphics[width=1.0\linewidth]{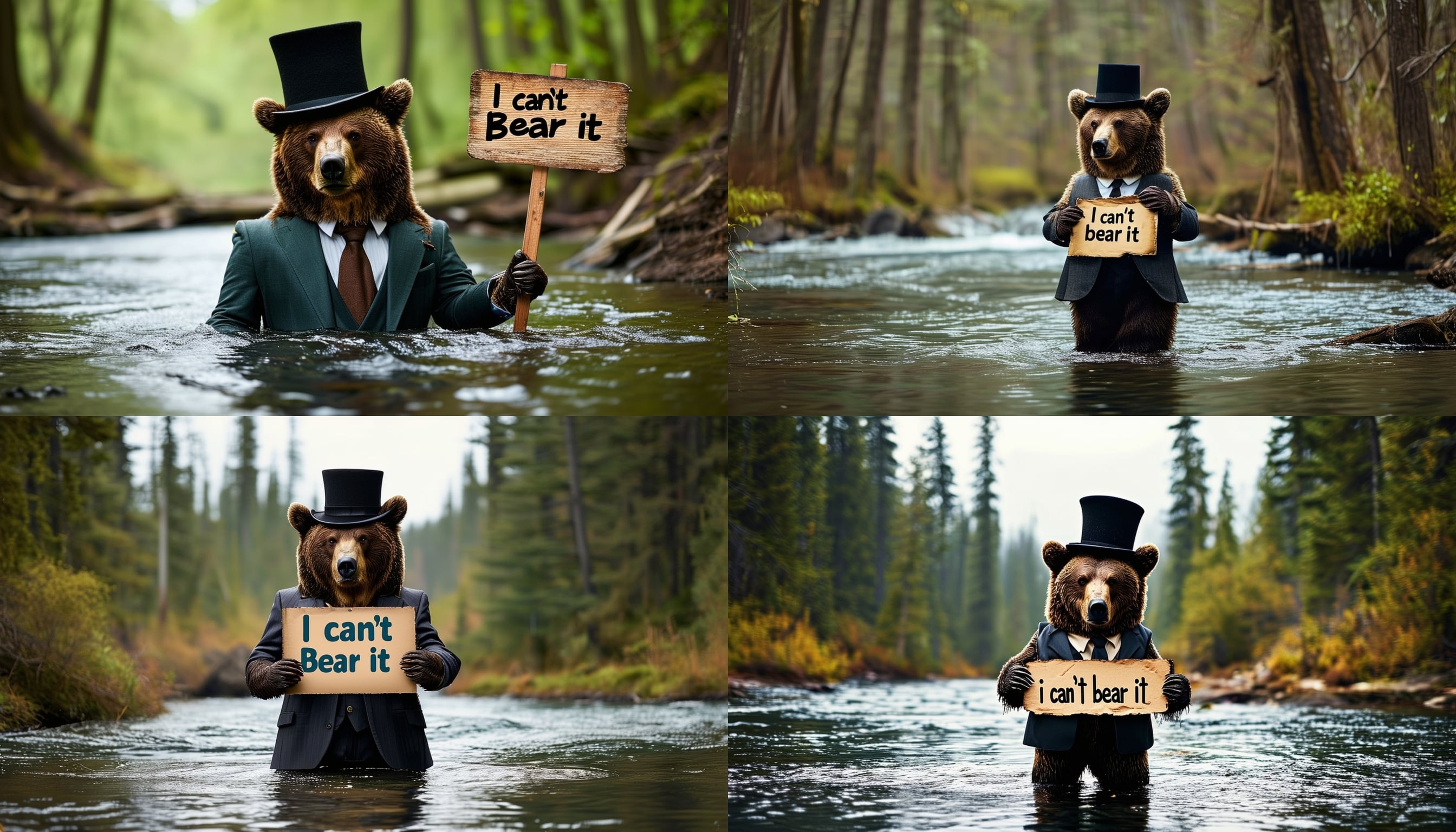} \\
        \tiny{photo of a bear wearing a suit and tophat in a river in the middle of a forest holding a sign that says "I cant bear it".}
    \end{minipage}

    \begin{minipage}[t]{0.495\linewidth}
        \centering
        \includegraphics[width=1.0\linewidth]{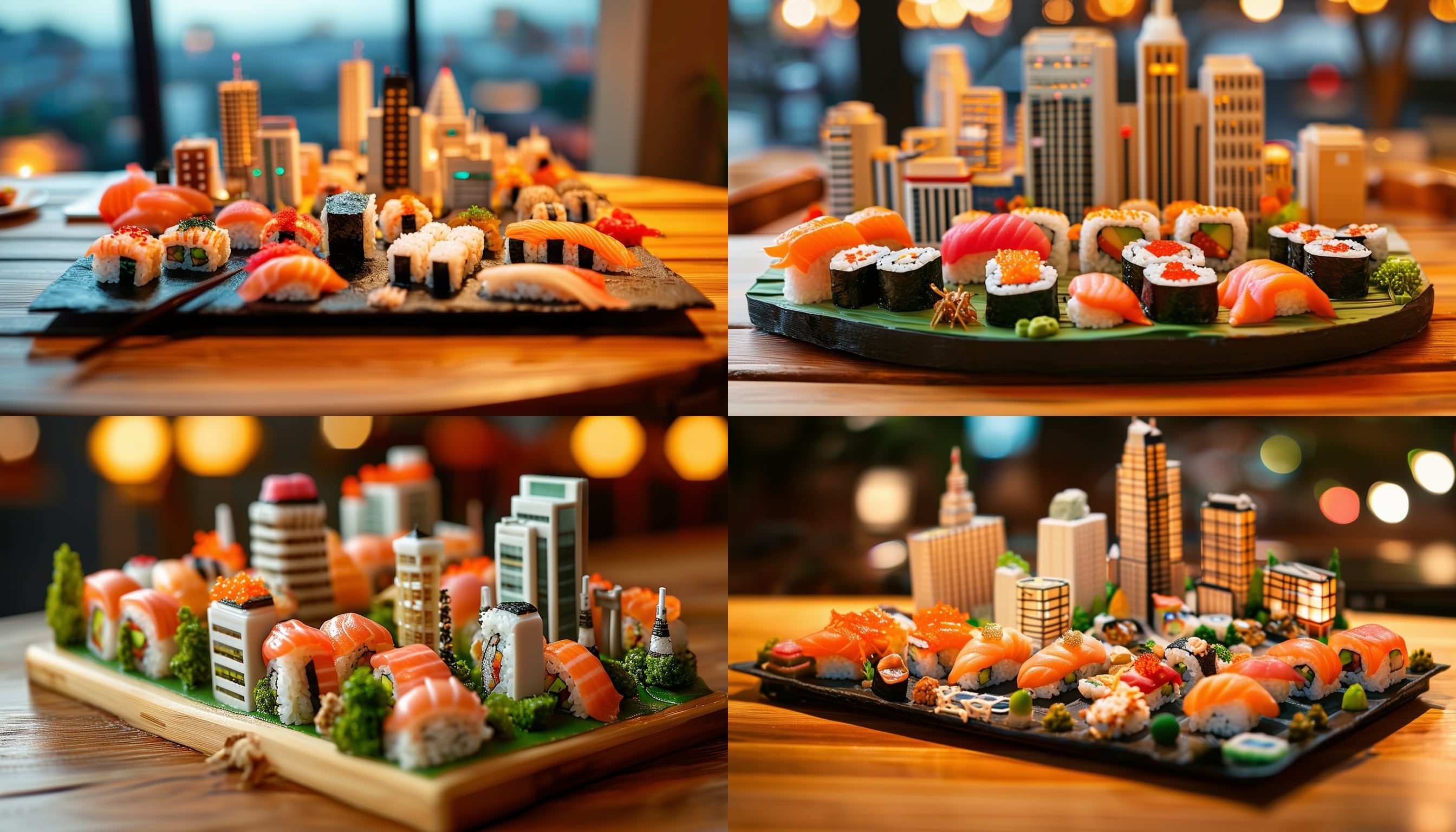} \\
        \tiny{tilt shift aerial photo of a cute city made of sushi on a wooden table in the evening.}
    \end{minipage}
    \hfill
    \begin{minipage}[t]{0.495\linewidth}
        \centering
        \includegraphics[width=1.0\linewidth]{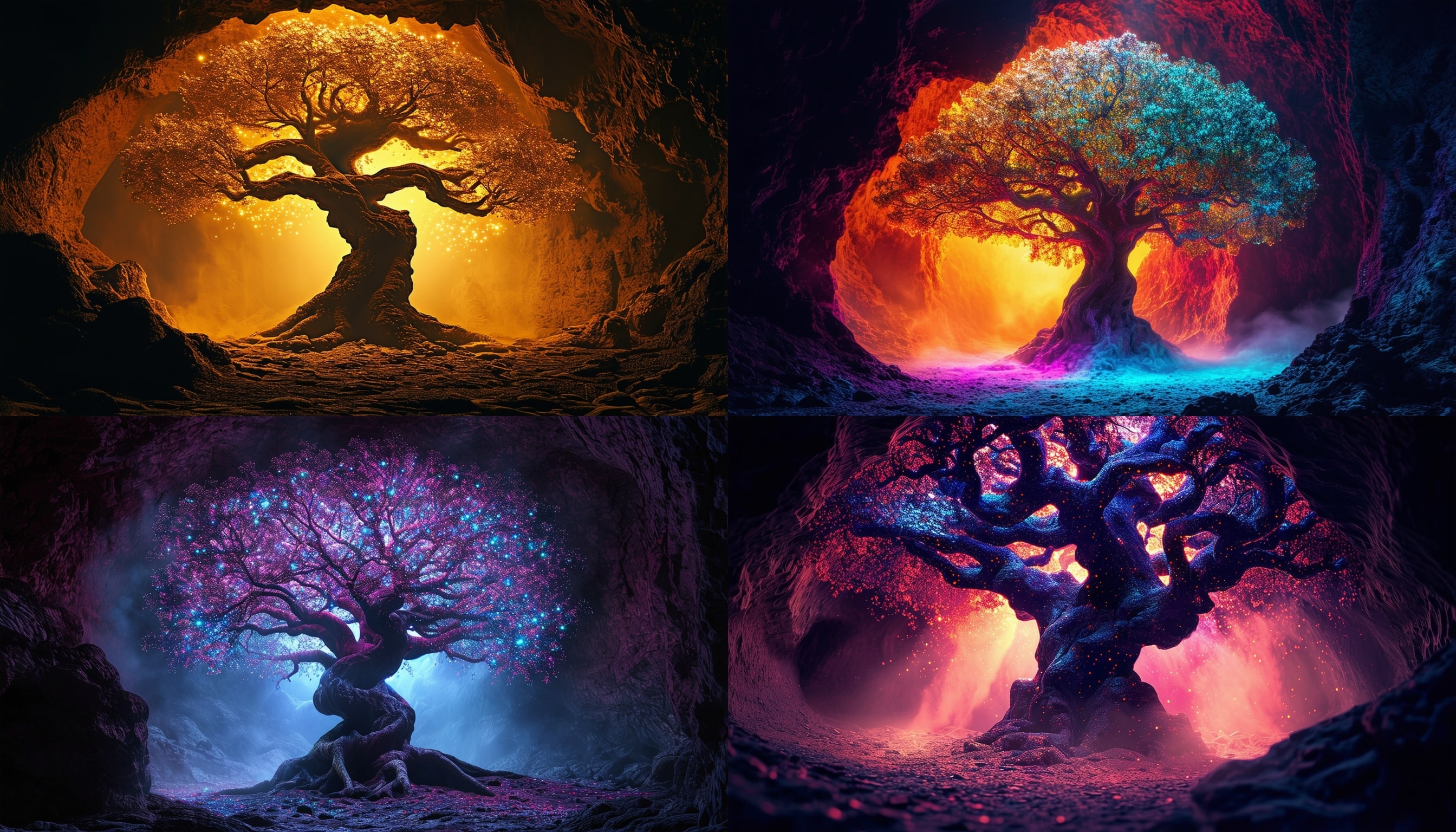} \\
        \tiny{dark high contrast render of a psychedelic tree of life illuminating dust in a mystical cave.}
    \end{minipage}  

    \begin{minipage}[t]{0.495\linewidth}
        \centering
        \includegraphics[width=1.0\linewidth]{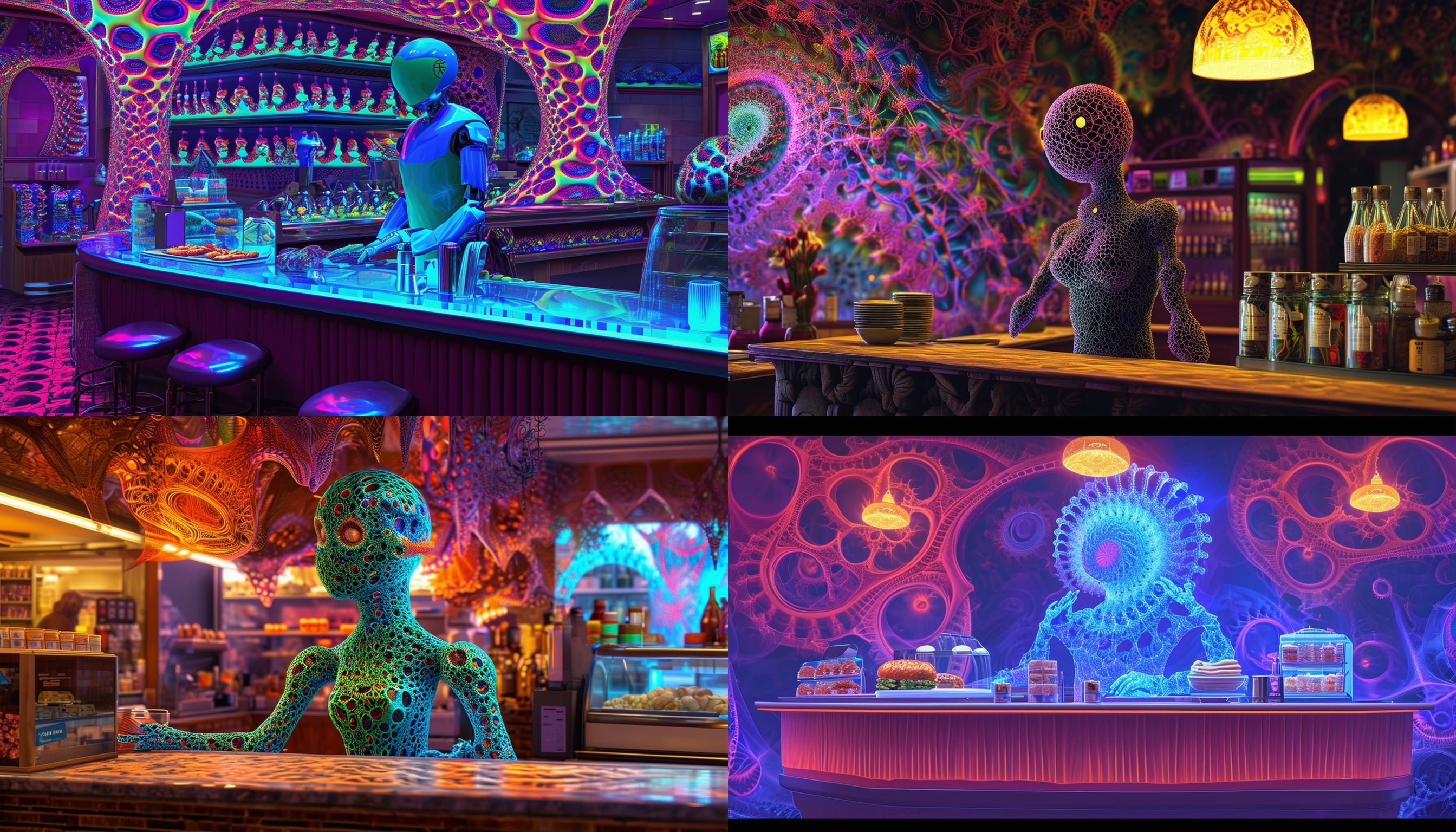} \\
        \tiny{an anthropomorphic fractal person behind the counter at a fractal themed restaurant.}
    \end{minipage}
    \hfill
    \begin{minipage}[t]{0.495\linewidth}
        \centering
        \includegraphics[width=1.0\linewidth]{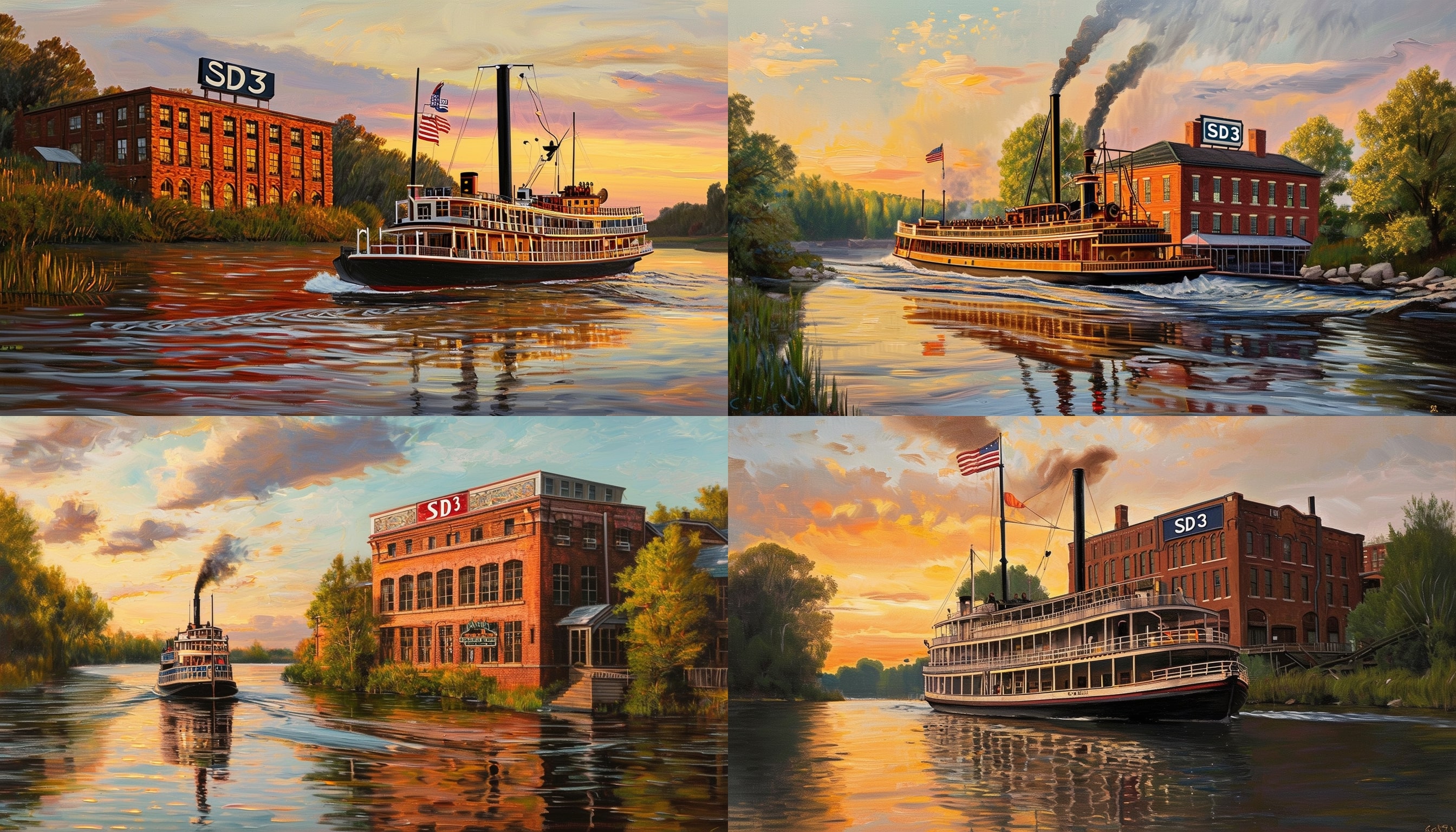} \\
        \tiny{beautiful oil painting of a steamboat in a river in the afternoon. On the side of the river is a large brick building with a sign on top that says \"SD3\".}
    \end{minipage}  

    \begin{minipage}[t]{0.495\linewidth}
        \centering
        \includegraphics[width=1.0\linewidth]{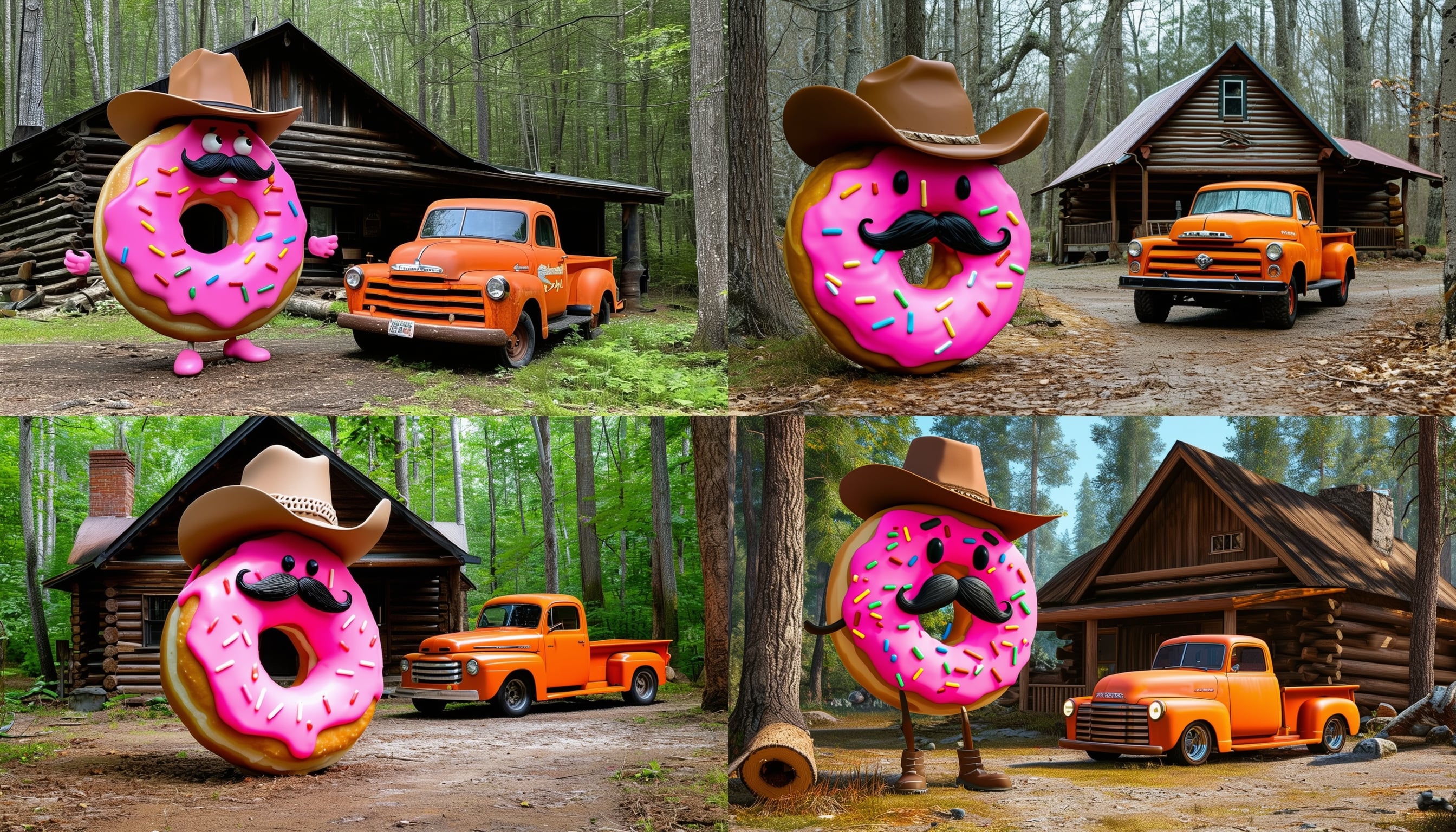} \\
        \tiny{an anthopomorphic pink donut with a mustache and cowboy hat standing by a log cabin in a forest with an old 1970s orange truck in the driveway}
    \end{minipage}
    \hfill
    \begin{minipage}[t]{0.495\linewidth}
        \centering
        \includegraphics[width=1.0\linewidth]{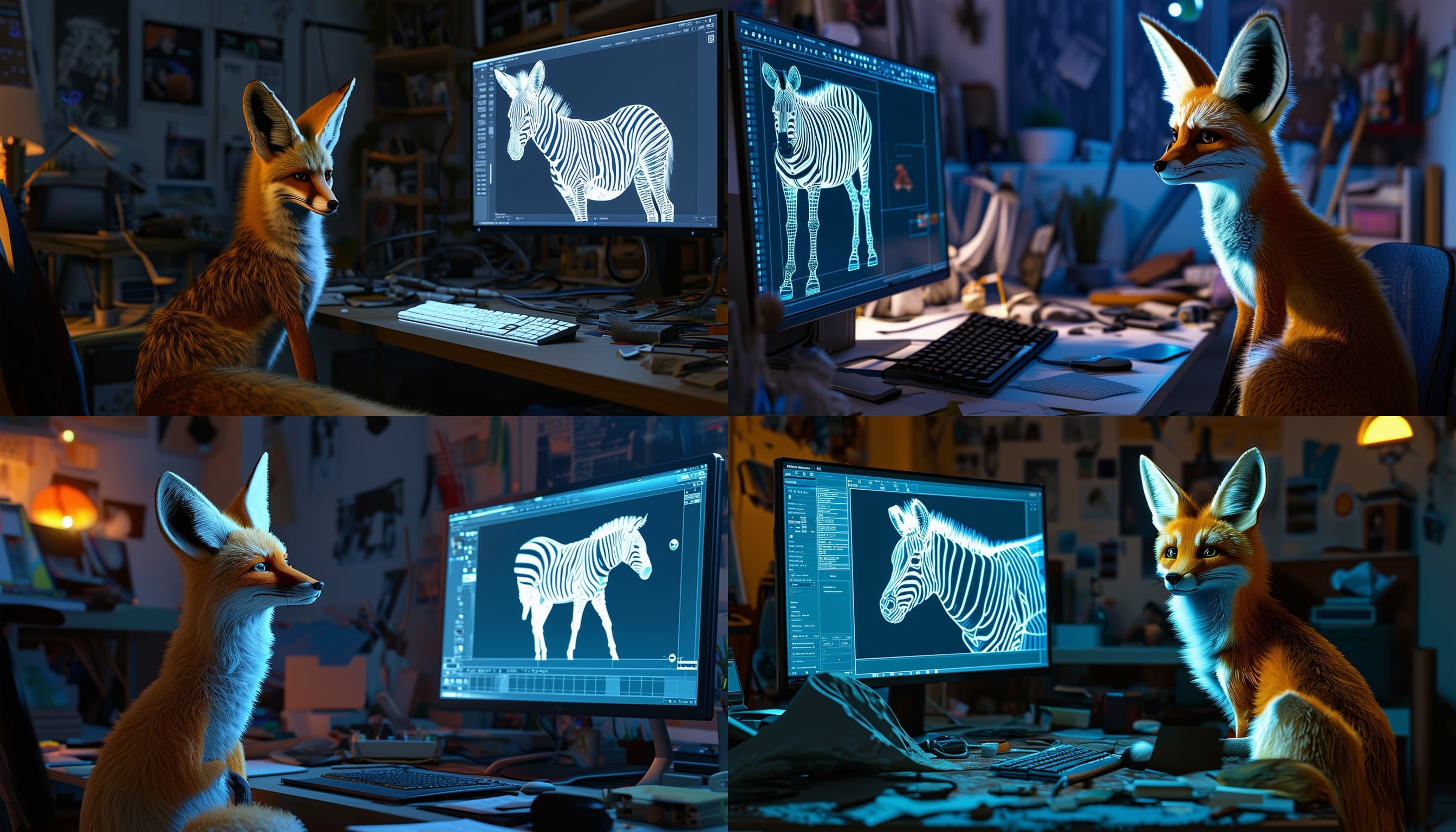} \\
        \tiny{fox sitting in front of a computer in a messy room at night. On the screen is a 3d modeling program with a line render of a zebra.}
    \end{minipage}  
\end{figure*}
}

\newcommand{\aereconstructiontable}{
\begin{table}[t]
  \centering
  \begin{tabular}{lccc}
    \toprule
    Metric & 4 chn & 8 chn & 16 chn \\
    \midrule
    FID ($\downarrow$) & \s2.41 & \s1.56 & \textbf{\s1.06} \\
    Perceptual Similarity ($\downarrow$) & \s0.85 & \s0.68 & \textbf{\s0.45} \\
    SSIM ($\uparrow$) & \s0.75 & \s0.79 & \textbf{\s0.86} \\
    PSNR ($\uparrow$) & 25.12 & 26.40 & \textbf{28.62} \\
    \bottomrule
  \end{tabular}
  \vspace{-0.5em}
  \caption{\textbf{Improved Autoencoders.} Reconstruction performance metrics for different channel configurations. The downsampling factor for all models is $f=8$. \vspace{-1em}}
  \label{tab:aereconstructiontable}
\end{table}
}

\newcommand{\aereconstructiontablefullprecision}{
\begin{table}[ht]
  \centering
  \begin{tabular}{lccc}
    \toprule
    Metric & 4 chn & 8 chn & 16 chn \\
    \midrule
    \textbf{Fidelity} & & & \\
    FID ($\downarrow$) & 2.4132 & 1.5607 & 1.0638 \\
    \midrule
    \textbf{Image Similarity} & & & \\
    PSNR ($\uparrow$) & 25.1210 & 26.3944 & 28.6241 \\
    Perceptual Similarity ($\downarrow$) & 0.8480 & 0.6761 & 0.4544 \\
    SSIM ($\uparrow$) & 0.7536 & 0.7920 & 0.8574 \\
    \bottomrule
  \end{tabular}
  \caption{\textbf{Improved} Reconstruction performance metrics for different channel configurations.}
  \label{tab:aereconstructiontable}
\end{table}
}

\newcommand{\biggerismorestepefficient}{
\begin{table}[t]
\centering
\begin{tabular}{@{}lccc@{}c}
\toprule
  & \multicolumn{4}{c}{relative CLIP score decrease [\%]} \\
\cmidrule(l){2-5}
 & \multicolumn{1}{p{2cm}}{\centering 5/50 steps} & \multicolumn{1}{p{2cm}}{\centering 10/50 steps} & \multicolumn{1}{p{2cm}}{\centering 20/50 steps} & path length \\
\midrule
  depth=15 & 4.30 & 0.86 & 0.21 & 191.13 \\
  depth=30 & 3.59 & 0.70 & 0.24 & 187.96 \\
  depth=38 & 2.71 & 0.14 & 0.08 & 185.96 \\
\bottomrule
\end{tabular}
\caption{
\textbf{Impact of model size on sampling efficiency.}
The table shows the relative performance decrease relative to CLIP scores evaluated using 50 sampling steps at a fixed seed. 
Larger models can be sampled using fewer steps, which we attribute to increased
robustness and better fitting the straight-path objective of rectified flow
models, resulting in shorter path lengths. Path length is calculated by summing up $ \Vert v_\theta \cdot dt \Vert $ over 50 steps. \vspace{-1em}
\label{tab:biggerismorestepefficient}
}
\end{table}
}

\newcommand{\genEvalSotaTable}{
\begin{table}[t]
\tiny
\centering
\setlength{\tabcolsep}{2pt}
\resizebox{.99\linewidth}{!}{
\begin{tabular}{@{}lccccccc@{}}
\toprule
\multirow{2}{*}[-9.7pt]{Model} & \multirow{2}{*}[-9.9pt]{Overall} & \multicolumn{2}{c}{Objects} & \multirow{2}{*}[-9.7pt]{Counting} & \multirow{2}{*}[-9.7pt]{Colors} & \multirow{2}{*}[-9.7pt]{Position} & \multirow{2}{*}[-0.5pt] {Color} \\
\cmidrule(lr){3-4}
 &  & Single & Two &  &  &  & Attribution \\
\midrule
minDALL-E & 0.23 & 0.73 & 0.11 & 0.12 & 0.37 & 0.02 & 0.01 \\
SD v1.5 & 0.43 & 0.97 & 0.38 & 0.35 & 0.76 & 0.04 & 0.06 \\
PixArt-alpha & 0.48 & 0.98 & 0.50 & 0.44 & 0.80 & 0.08 & 0.07 \\
SD v2.1 & 0.50 & 0.98 & 0.51 & 0.44 & \underline{0.85} & 0.07 & 0.17 \\
DALL-E 2 & 0.52 & 0.94 & 0.66 & 0.49 & 0.77 & 0.10 & 0.19 \\
SDXL & 0.55 & 0.98 & 0.74 & 0.39 & \underline{0.85} & 0.15 & 0.23 \\
SDXL Turbo          & 0.55 & \textbf{1.00} & 0.72 & 0.49 & 0.80 & 0.10 & 0.18 \\
IF-XL & 0.61 & 0.97 & 0.74 & \textit{0.66} & 0.81 & 0.13 & 0.35 \\
DALL-E 3 & 0.67 & 0.96 & \underline{0.87} & 0.47 & \textit{0.83} & \textbf{0.43} & \textit{0.45} \\
\midrule
Ours (depth=18), $512^2$ & 0.58 & 0.97 & 0.72 & 0.52 & 0.78 & 0.16 & 0.34 \\
Ours (depth=24), $512^2$ & 0.62 & 0.98 & 0.74 & 0.63 & 0.67 & \textit{0.34} & 0.36 \\
Ours (depth=30), $512^2$ & 0.64 & 0.96 & 0.80 & 0.65 & 0.73 & 0.33 & 0.37 \\
Ours (depth=38), $512^2$ & \textit{0.68} & 0.98 & 0.84 & \textit{0.66} & 0.74 & \underline{0.40} & 0.43 \\
Ours (depth=38), $512^2$ w/DPO & \underline{0.71} & 0.98 & \underline{0.89} & \textbf{0.73} & \textit{0.83} & \textit{0.34} & \underline{0.47} \\
Ours (depth=38), $1024^2$ w/DPO & \textbf{0.74} & \underline{0.99} & \textbf{0.94} & \underline{0.72} & \textbf{0.89} & 0.33 & \textbf{0.60} \\
\bottomrule
\end{tabular}
}
\caption{\textbf{GenEval comparisons}. Our largest model (depth=38) outperforms all current open models and DALLE-3~\citep{betker2023improving} on GenEval~\citep{ghosh2023geneval}.
We highlight the \textbf{best}, \underline{second best}, and \textit{third best} entries. For DPO, see \Cref{supsec:dpo}.
}\label{tab:genEvalSotaTable}
\end{table}
}

\newcommand{\spelling}{
\vspace{-1em}
\begin{table}
\centering
\setlength{\tabcolsep}{5pt}
\begin{tabular}{@{}l@{\hspace{3em}}cc@{}}
\toprule
ours (depth=38) w/DPO & vs. Ideogram v2 & vs. DALLE-3  \\
\midrule
Win Rate & 61\% & 54 \% \\
\bottomrule
\end{tabular}
\vspace{-5pt}
\caption{\label{tab:spelling} 
\textbf{Spelling.} 
Our model outperforms \emph{Ideogram v2}~\citep{ideogramv02} and \emph{DALLE-3}~\citep{betker2023improving} in typography. 
We report human preference with respect to correct spelling based on the ``Writing`` category of PartiPrompts~\citep{yu2022scaling}.}
\vspace{-1.5em}
\end{table}
}

\newcommand{\captionstudy}{
\begin{table}
\centering

\begin{tabular}{@{}lcc@{}}
\toprule
& \multicolumn{1}{c}{Original Captions} & \multicolumn{1}{c}{50/50 Mix} \\
\cmidrule(lr){2-3}
& success rate [\%] & success rate [\%]  \\
\midrule
Color Attribution & 11.75 & 24.75 \\
Colors & 71.54 & 68.09 \\
Position & \s6.50 & 18.00 \\
Counting & 33.44 & 41.56 \\
Single Object & 95.00 & 93.75 \\
Two Objects & 41.41 & 52.53 \\ \midrule
Overall score & 43.27 & 49.78 \\
\bottomrule
\end{tabular}
\vspace{-0.5em}
\caption{\textbf{Improved Captions}. Using a 50/50 mixing ratio of synthetic (via CogVLM~\citep{wang2023cogvlm}) and original captions improves text-to-image performance. Assessed via the GenEval~\citep{ghosh2023geneval} benchmark.}
\label{tab:captionstudy}
\end{table}
}

\newcommand{\preencodingspeedups}{
\begin{table}
    \setlength{\tabcolsep}{5pt}
    \centering
    \begin{tabular}{lcccc}
        \toprule
         Model & Mem [GB] & FP [ms] & Storage [kB] & Delta [\%] \\ \midrule
         VAE (Enc) & 0.14 & 2.45 & 65.5 & 13.8 \\
         CLIP-L & 0.49 & 0.45 & 121.3 & 2.6 \\
         CLIP-G & 2.78 & 2.77 & 202.2 & 15.6 \\
         T5 & 19.05 & 17.46 & 630.7 & 98.3 \\
         \bottomrule
    \end{tabular}
\caption{
\textbf{   Key figures for preencoding frozen input networks.}
Mem is the memory required to load the model on the GPU. FP [ms] is the time per sample for the forward pass with per-device batch size of 32. Storage is the size to save a single sample. Delta [\%] is how much longer a training step takes, when adding this into the loop for the 2B MMDiT-Model (568ms/it).}
\label{tab:preencodingspeedups}
\end{table}
}

\begin{abstract}
\vspace{-0.25em} 
Diffusion models create data from noise by inverting the forward paths of data towards noise and have emerged as a powerful generative modeling technique for high-dimensional, perceptual data such as images and videos. Rectified flow is a recent generative model formulation that connects data and noise in a straight line. Despite its better theoretical properties and conceptual simplicity, it is not yet decisively established as standard practice. In this work, we improve existing noise sampling techniques for training rectified flow models by biasing them towards perceptually relevant scales. Through a large-scale study, we demonstrate the superior performance of this approach compared to established diffusion formulations for high-resolution text-to-image synthesis.  Additionally, we present a novel transformer-based architecture for text-to-image generation that uses separate weights for the two modalities and enables a bidirectional flow of information between image and text tokens, improving text comprehension, typography, and human preference ratings. We demonstrate that this architecture follows predictable scaling trends and correlates lower validation loss to improved text-to-image synthesis as measured by various metrics and human evaluations. Our largest models outperform state-of-the-art models, and we will make our experimental data, code, and model weights publicly available.
\vspace{-1.0em}
\end{abstract}
    
\section{Introduction}%
\label{sec:intro}
Diffusion models create data from noise~\citep{Song2020ScoreBasedGM}. 
They are trained to 
invert forward paths of data towards random noise and, thus, %
in 
conjunction with approximation and generalization properties of neural networks, %
can be used to generate new data points that are not present in the training data but follow the distribution of the training data~\citep{SohlDickstein2015DeepUL, song2020generative}. This generative modeling technique has proven to be very effective for modeling high-dimensional, perceptual data such as images~\citep{ho2020denoising}. In recent years, diffusion models have become the de-facto approach for generating high-resolution images and videos from natural language inputs with impressive generalization capabilities~\citep{saharia2022photorealistic, ramesh2022hierarchical, Rombach_2022, podell2023sdxl, dai2023emu, esser2023structure, blattmann2023align, betker2023improving, blattmann2023stable, singer2022makeavideo}. Due to their iterative nature and the associated computational costs, as well as the long sampling times during inference, research on formulations for more efficient training and/or faster sampling of these models has increased~\citep{karras2023analyzing, liu2022flow}. 

While specifying a forward path from data to noise leads to efficient training, it also raises the question of which path to choose. 
This choice can have important implications for sampling. For example, a forward process that fails to remove all noise from the data can lead to a discrepancy in training and test distribution and result in artifacts such as gray image samples~\citep{lin2024common}. 
Importantly, the choice of the forward process also influences the learned backward process and, thus, the sampling efficiency. While curved paths require many integration steps to simulate the process, a straight path could be simulated with a single step and is less prone to error accumulation. Since each step corresponds to an evaluation of the neural network, this has a direct impact on the sampling speed.

A particular choice for the forward path is a so-called \emph{Rectified Flow}~\citep{liu2022flow,albergo2022building,lipman2023flow}, which connects data and noise on a straight line. Although this model class has better theoretical properties, it has not yet become decisively established in practice. So far, some advantages have been empirically demonstrated in small and medium-sized experiments~\cite{ma2024sit}, but these are mostly limited to class-conditional models. 
In this work, we change this by introducing a re-weighting of the noise scales in rectified flow models, similar to noise-predictive diffusion models~\citep{ho2020denoising}. Through a large-scale study, we compare our new formulation to existing diffusion formulations and demonstrate its benefits.  

We show that the widely used approach for text-to-image synthesis, where a fixed text representation is fed directly into the model (e.g., via cross-attention~\citep{vaswani2017attention, Rombach_2022}), is not ideal, and present a new architecture that incorporates learnable streams for both image and text tokens, which enables a two-way flow of information between them. 
We combine this with our improved rectified flow formulation and investigate its scalability. 
We demonstrate a predictable scaling trend in the validation loss and show that a lower validation loss correlates strongly with improved automatic and human evaluations.

Our largest models outperform state-of-the art open models such as \emph{SDXL}~\citep{podell2023sdxl},  \emph{SDXL-Turbo}~\citep{sauer2023adversarial}, \emph{Pixart-$\alpha$}~\citep{chen2023pixart}, and closed-source models such as DALL-E 3~\citep{betker2023improving} both in quantitative evaluation~\citep{ghosh2023geneval} of prompt understanding and human preference ratings.

The core contributions of our work are: 
(i) We conduct a large-scale, systematic study on different diffusion model and rectified flow formulations to identify the best setting. For this purpose, we introduce new noise samplers for rectified flow models that improve performance over previously known samplers. %
(ii) We devise a novel, scalable architecture for text-to-image synthesis that allows bi-directional mixing between text and image token streams within the network. We show its benefits compared to established backbones such as UViT~\citep{hoogeboom2023simple} and DiT~\citep{Peebles_2023}. 
Finally, we (iii) perform a scaling study of our model and demonstrate that it follows predictable scaling trends. We show that a lower validation loss correlates strongly with improved text-to-image performance assessed via metrics such as T2I-CompBench~\citep{huang2023t2i}, GenEval~\citep{ghosh2023geneval} and human ratings.
We make results, code, and model weights publicly available.
\vspace{-1em}
\section{Simulation-Free Training of Flows}
\label{sec:method}

We consider generative models that define a mapping between samples $x_1$ from a
noise distribution $p_1$ to samples $x_0$ from a data distribution $p_0$ in
terms of an ordinary differential equation (ODE),
\begin{equation}
  \label{eq:ode}
  dy_t = v_\Theta(y_t, t)\,dt\;,
\end{equation}
where the velocity $v$ is parameterized by the weights $\Theta$ of a neural network.
Prior work by~\citet{Chen2018NeuralOD} suggested to directly solve \Cref{eq:ode} via differentiable ODE solvers. However, this process is computationally expensive, especially for large network architectures that parameterize $v_\Theta(y_t, t)$. 
A more efficient alternative is to directly regress a vector field $u_t$ that generates a probability path between $p_0$ and $p_1$.
To construct such a $u_t$, we define a forward process, corresponding to a probability path $p_t$ between $p_0$ and $p_1=\mathcal{N}(0, 1)$, as
\begin{equation}
  \label{eq:forwardprocess}
  z_t = a_t x_0 + b_t \epsilon\quad\text{where}\;\epsilon \sim \mathcal{N}(0,I)\;. 
\end{equation}
For $a_0 = 1, b_0 = 0, a_1 = 0$ and $b_1 = 1$, the marginals,
\begin{align}
  \label{eq:marginals}
  p_t(z_t) &=
  \mathbb{E}_{\epsilon \sim \mathcal{N}(0,I)}
  p_t(z_t \vert \epsilon)\;,
\end{align}
are consistent with the data and noise distribution.

To express the relationship between $z_t, x_0$ and $\epsilon$, we introduce
$\psi_t$ and $u_t$ as 
\begin{align}
      \psi_t(\cdot | \epsilon) &: x_0 \mapsto a_t x_0 + b_t \epsilon \\
        u_t(z| \epsilon) &\coloneqq \psi'_t(\psi_t^{-1}(z| \epsilon)  \vert \epsilon) \label{eq:utmapsdiffinverse}
\end{align}

Since $z_t$ can be written as solution to the ODE  $z_t' = u_t(z_t |\epsilon)$, with initial value $z_0=x_0$, $u_t(\cdot | \epsilon)$ generates $p_t(\cdot | \epsilon)$.
Remarkably, one can construct a marginal vector field $u_t$ which generates the marginal probability paths $p_t$  \cite{lipman2023flow} (see \ref{subsec:flowproofs}), using the conditional vector fields $u_t(\cdot | \epsilon)$:
\begin{align}
    u_t(z) = \mathbb{E}_{\epsilon \sim
  \mathcal{N}(0,I)} u_t(z \vert \epsilon) \frac{p_t(z \vert
  \epsilon)}{p_t(z)}
  \label{eq:marginal_u}
\end{align}
While regressing $u_t$ with the \emph{Flow Matching} objective 
\begin{align}
   \mathcal{L}_{FM} =  \mathbb{E}_{t, p_t(z)} || v_{\Theta}(z, t) - u_t(z) ||_2^2.
\end{align}
directly is intractable due to the marginalization in \eqref{eq:marginal_u}, 
\emph{Conditional Flow Matching} (see \ref{subsec:flowproofs}),
\begin{align}
   \mathcal{L}_{CFM} =  \mathbb{E}_{t, p_t(z | \epsilon), p(\epsilon) }|| v_{\Theta}(z, t) - u_t(z | \epsilon)  ||_2^2\;, \label{eq:condflowmatch}
\end{align}
with the conditional vector fields $u_t(z \vert \epsilon)$ provides an equivalent yet tractable objective.

To convert the loss into an explicit form we insert
 $\psi_t'(x_0 \vert \epsilon) = a_t'x_0 + b_t' \epsilon$ and $\psi_t^{-1}(z\vert \epsilon) = \frac{z -b_t \epsilon}{a_t}$ into (\ref{eq:utmapsdiffinverse})
\begin{equation}
  z_t' = u_t(z_t \vert \epsilon) = \frac{a_t'}{a_t} z_t - \epsilon b_t (\frac{a_t'}{a_t} - \frac{b_t'}{b_t})\;.
  \label{eq:zode}
\end{equation}
Now, consider the \emph{signal-to-noise ratio} $\lambda_t := \log \frac{a_t^2}{b_t^2}$. 
With $\lambda_t' = 2 (\frac{a_t'}{a_t} - \frac{b_t'}{b_t})$, we can rewrite \Cref{eq:zode} as 
\begin{equation}
  u_t(z_t \vert \epsilon) = \frac{a_t'}{a_t} z_t - \frac{b_t}{2} \lambda_t' \epsilon 
  \label{eq:zodewithlambda}
\end{equation}

Next, we use \Cref{eq:zodewithlambda} to reparameterize \Cref{eq:condflowmatch} as a noise-prediction objective:
\begin{align}
    \mathcal{L}_{CFM} &= \mathbb{E}_{t, p_t(z | \epsilon), p(\epsilon) } || v_{\Theta}(z, t) - \frac{a_t'}{a_t} z  + \frac{b_t}{2} \lambda_t' \epsilon ||_2^2 \\
    &= \mathbb{E}_{t, p_t(z | \epsilon), p(\epsilon) } \left(-\frac{b_t}{2}\lambda_t' \right)^2  || \epsilon_\Theta(z, t) - \epsilon ||_2^2
\end{align}
where we defined $\epsilon_\Theta \coloneqq \frac{-2}{\lambda_t' b_t} (v_\Theta - \frac{a_t'}{a_t} z)$. 

Note that the optimum of the above objective does not change when
introducing a time-dependent weighting. Thus, one can derive various weighted
loss functions that provide a signal towards the desired solution but might
affect the optimization trajectory. For a unified analysis of different
approaches, including classic diffusion formulations, we can write the objective
in the following form (following \citet{Kingma2023UnderstandingDO}):
\begin{equation*}
  \mathcal{L}_w(x_0) = -\frac{1}{2} \mathbb{E}_{t\sim\mathcal{U}(t), \epsilon\sim \mathcal{N}(0, I)}
  \left[ w_t \lambda_t' \Vert \epsilon_\Theta(z_t, t) - \epsilon
  \Vert^2 \right]\;,
\end{equation*}
where $w_t = -\frac{1}{2} \lambda_t' b_t^2$ corresponds to $\mathcal{L}_{CFM}$.

\section{Flow Trajectories}
\label{subsec:variants}
In this work, we consider different variants of the above formalism that we
briefly describe in the following.

\paragraph{Rectified Flow}
Rectified Flows (RFs) \cite{liu2022flow,albergo2022building,lipman2023flow} define the forward process as straight paths
between the data distribution and a standard normal distribution, i.e.
\begin{equation}
z_t = (1-t) x_0 + t \epsilon\;,
\end{equation}
and uses $\mathcal{L}_{CFM}$ which then corresponds to $w_t^\text{RF} = \frac{t}{1-t}$. The network output directly parameterizes the velocity $v_\Theta$.

\paragraph{EDM}
EDM \cite{Karras2022ElucidatingTD} uses a forward process of the form
\begin{equation}
  z_t = x_0 + b_t \epsilon
\end{equation}
where \cite{Kingma2023UnderstandingDO}
$ b_t = \exp{F_{\mathcal{N}}^{-1}(t \vert P_m, P_s^2)} $
with $F_{\mathcal{N}}^{-1}$ being the quantile function of the normal
distribution with mean $P_m$ and variance $P_s^2$. Note that this choice results in
\begin{equation}
  \lambda_t \sim \mathcal{N}(-2P_m, (2P_s)^2)\quad\text{for}\;t\sim
  \mathcal{U}(0,1)
\end{equation}
The %
network is parameterized through an \textbf{F}-prediction
\cite{Kingma2023UnderstandingDO,Karras2022ElucidatingTD} and the loss can be
written as $\mathcal{L}_{w_t^\text{EDM}}$ with
\begin{equation}
  w_t^\text{EDM} = \mathcal{N}(\lambda_t \vert -2P_m, (2P_s)^2)(e^{-\lambda_t}+0.5^2)
\end{equation}

\paragraph{Cosine}
\cite{nichol2021improved} proposed a forward process of the form
\begin{equation}
  z_t = \cos\bigl(\frac{\pi}{2} t\bigr) x_0 + \sin\bigl(\frac{\pi}{2} t\bigr) \epsilon\;.
\end{equation}
In combination with an $\epsilon$-parameterization and loss, this corresponds
to a weighting $w_t = \operatorname{sech}(\lambda_t/2)$.
When combined with a \textbf{v}-prediction loss
\cite{Kingma2023UnderstandingDO}, the weighting is given by $w_t = e^{-\lambda_t/2}$.

\paragraph{(LDM-)Linear}
LDM \cite{Rombach_2022} uses a modification of the DDPM schedule \cite{ho2020denoising}. Both are
variance preserving schedules, i.e. $b_t=\sqrt{1-a_t^2}$, and define $a_t$ for
discrete timesteps $t = 0, \dots, T-1$ in terms of diffusion coefficients
$\beta_t$ as
$ a_t  = (\prod_{s=0}^t (1 - \beta_s))^{\frac{1}{2}}$.
For given boundary values $\beta_0$ and $\beta_{T-1}$, DDPM uses
$ \beta_t = \beta_0 + \frac{t}{T-1} (\beta_{T-1} - \beta_0) $
and LDM uses
$\beta_t = \left( \sqrt{\beta_0\vphantom{\beta_{T-1}}} + \frac{t}{T-1} (\sqrt{\beta_{T-1}} - \sqrt{\beta_0\vphantom{\beta_{T-1}}}) \right)^2$. 

\subsection{Tailored SNR Samplers for RF models}
\label{subsubsec:tailoredsnrforrf}
The RF loss trains the velocity $v_\Theta$ uniformly on all timesteps in
$[0, 1]$.
Intuitively, however, the resulting velocity prediction target $\epsilon - x_0$ is more
difficult for $t$ in the middle of $[0, 1]$, since for $t=0$, the optimal
prediction is the mean of $p_1$, and for $t=1$ the optimal prediction is the
mean of $p_0$. 
In general, changing the distribution over $t$ from the commonly used
uniform distribution $\mathcal{U}(t)$ to a distribution with density $\pi(t)$
is equivalent to a weighted loss $\mathcal{L}_{w_t^\pi}$ with
\begin{equation}
  w_t^\pi = \frac{t}{1-t}\pi(t)
\end{equation}
Thus, we aim to give more weight to intermediate timesteps by sampling them more frequently. Next, we describe the timestep densities $\pi(t)$ that we use to train our models.

\paragraph{Logit-Normal Sampling}
One option for a distribution that puts more weight on intermediate steps
is the logit-normal distribution~\cite{atchison1980logistic}. Its density,
\begin{equation}
\pi_{\text{ln}}(t; m, s) = \frac{1}{s\sqrt{2\pi}} \frac{1}{t(1-t)}\exp\Bigl(-\frac{(\text{logit}(t)-m)^2}{2s^2}\Bigr),
\label{eq:logitnormal}
\end{equation}
where $\text{logit}(t) = \log \frac{t}{1-t}$, has a location parameter, $m$,
and a scale parameter, $s$. The location parameter enables us to bias
the training timesteps towards either data $p_0$ (negative $m$) or noise $p_1$
(positive $m$). As shown in \Cref{fig:tsamplers}, the scale parameters controls how
wide the distribution is.

In practice, we sample the random variable $u$ from a normal distribution $u \sim \mathcal{N}(u; m, s)$ and map it through the standard logistic function.

\paragraph{Mode Sampling with Heavy Tails}
The logit-normal density always vanishes at the endpoints $0$ and $1$. To study
whether this has adverse effects on the performance, we also use a timestep
sampling distribution with strictly positive density on $[0, 1]$. For a scale
parameter $s$, we define
\begin{equation}
f_{\text{mode}}(u; s) = 1 - u - s\cdot\Bigl(\cos^2\bigl(\frac{\pi}{2} u\bigr) - 1 + u\Bigr).
\label{eq:modesampler}
\end{equation}
For $ -1 \leq s \leq \frac{2}{\pi -2}$, this function is
monotonic, and we can use it to sample from the implied density
$\pi_{\text{mode}}(t; s) = \left| \frac{d}{dt} f_{\text{mode}}^{-1}(t) \right|$.
As seen in \Cref{fig:tsamplers}, the scale parameter controls the degree to which either
the midpoint (positive $s$) or the endpoints (negative $s$) are favored during sampling.
This formulation also includes a \textbf{uniform weighting}
$\pi_{\text{mode}}(t; s=0) = \mathcal{U}(t)$ for $s=0$, which has been used
widely in previous works on Rectified Flows \cite{liu2022flow,ma2024sit}.

\paragraph{CosMap}
Finally, we also consider the \emph{cosine} schedule \cite{nichol2021improved}
from \Cref{subsec:variants} in the RF setting. 
In particular, we are looking for a mapping $f: u \mapsto f(u) = t, \; u \in [0, 1]$, %
such that the log-snr matches that of the cosine schedule: $2 \log \frac{\cos(\frac{\pi}{2}u)}{\sin(\frac{\pi}{2}u)} = 2 \log \frac{1-f(u)}{f(u)}$. 
Solving for $f$, we obtain for $u \sim \mathcal{U}(u)$
\begin{equation}
t = f(u) = 1 - \frac{1}{\tan(\frac{\pi}{2} u) + 1}, 
\label{eq:cosmap}
\end{equation}
from which we obtain the density 
\begin{equation}
\pi_{\text{CosMap}}(t) = \left| \frac{d}{dt} f^{-1}(t) \right| = \frac{2}{\pi - 2\pi t + 2\pi t^2}.
\label{eq:cosmapdensity}
\end{equation}

\section{Text-to-Image Architecture}
\label{subsec:methodarch}
\begin{figure*}[t]
\centering
\begin{subfigure}[t]{0.49\textwidth}\centering
        \resizebox{!}{0.46\textheight}{\begin{tikzpicture}[
        block/.style={draw, fill=white, rectangle, minimum width=3cm,rounded corners=0.2cm},
        trainableblock/.style={block,fill=blue!10},
        frozenblock/.style={block,fill=blue!10},
        fnblock/.style={block,fill=green!10},
        operation/.style={draw, fill=red!10, circle, minimum width=2em},
        tensor/.style={draw, fill=VioletRed!50, circle},
        input/.style={block,fill=WildStrawberry!40},
    ]

    \pgfdeclarelayer{arrowlayer}
    \pgfdeclarelayer{residuallayer}
    \pgfdeclarelayer{decorations}
    \pgfdeclarelayer{bg}
    \pgfsetlayers{bg,arrowlayer,residuallayer,main,decorations}

    \newcommand{\clipGcolor}{LimeGreen!40}
    \newcommand{\clipLcolor}{ForestGreen!40}
    \newcommand{\tfiveColor}{YellowGreen!40}

    \node [input] (text) {Caption};

    \node [frozenblock, fill=\clipGcolor, below=1cm of text] (clipG) {CLIP-L/14};
    \node [frozenblock, fill=\clipLcolor, left=1.5cm of clipG.center] (clipL) {CLIP-G/14};
    \node [frozenblock, fill=\tfiveColor, right=1.5cm of clipG.center] (t5) {T5 XXL\phantom{/}};

    \coordinate [below=2.5cm of clipG] (conditioning);
    \coordinate [left=3.5cm of conditioning] (pool_conditioning);
    \draw [fill=\tfiveColor] (conditioning) |-++ (1.5, 1) coordinate (conditioning_br) --++ (0, -1.5) coordinate (conditioning_tr) -| cycle;
    \draw [fill=\clipLcolor] ($(conditioning)+(0, 0.25)$) -|++ (-1.5, 0.5) coordinate (conditioning_bl) -| cycle;
    \draw [fill=\clipGcolor] ($(conditioning)+(0, 0.75)$) -|++ (-1.5, 0.25) -| cycle;
    \draw [fill=black!10] ($(conditioning)+(0, 0.25)$) -|++ (-1.5, -0.75) coordinate (conditioning_tl) -| cycle;

    \draw [->] (t5) to [out=270, in=0] ($(conditioning_br)!0.5!(conditioning_tr)+(-0.5, 0)$);
    \draw [->] (clipG) to [out=270, in=90] ($(conditioning_bl)+(0.75, 0.125)$);
    \draw [->] (clipL) to [out=270, in=180] ($(conditioning_bl)+(0.2, -0.25)$);

    \draw [fill=\clipGcolor] ($(pool_conditioning)+(0, 0.5)$) -|++ (0.2, 0.5) -|++ (-0.4, -0.5) -- cycle;
    \draw [fill=\clipLcolor] ($(pool_conditioning)+(0, 0.5)$) -|++ (0.2, -1) -|++ (-0.4, 1) -- cycle;
    \path ($(pool_conditioning)+(-0.2, -0.5)$) --++ (0, 1.5) node [midway, rotate=90, yshift=0.2cm] {Pooled};

    \draw [->] (clipG) to [out=270, in=90] ($(pool_conditioning)+(0, 0.75)$);
    \draw [->] (clipL) to [out=270, in=0] ($(pool_conditioning)+(0, 0.)$);

    \begin{pgfonlayer}{decorations}
        \path [decoration={brace, amplitude=8pt, raise=4pt}, decorate, draw] ($(conditioning_bl)+(0, 0.25)$) -- (conditioning_br) node [midway, yshift=0.7cm, fill=white] {$77 + 77$ tokens};
        \path [decoration={brace, amplitude=8pt, raise=4pt}, decorate, draw] (conditioning_br) -- (conditioning_tr) node [midway, xshift=1.0cm, align=center] {$4096$\\ channel};
    \end{pgfonlayer}

    \node [trainableblock, below=1cm of conditioning] (c_lin) {Linear};
    \node [tensor, below=0.5cm of c_lin] (c_in) {$c$};

    \node [trainableblock, below=1cm of pool_conditioning] (y_mlp) {MLP};

    \begin{pgfonlayer}{arrowlayer}
        \draw [->] (text.south) to [out=270, in=90] (clipG);
        \draw [->] (text.south) to [out=270, in=90] (clipL);
        \draw [->] (text.south) to [out=270, in=90] (t5);

        \draw [->] (conditioning) -- (c_in);
    \end{pgfonlayer}


    \node [trainableblock, below=2cm of y_mlp] (t_emb) {MLP};
    \node [fnblock, below=0.2cm of t_emb] (t_enc) {Sinusoidal Encoding};
    \node [input, below=0.5cm of t_enc] (timestep) {Timestep};

    \node [operation, below=0.5cm of y_mlp] (y_plus) {$+$};
    \node [tensor, right=1cm of y_plus] (y_final) {y};

    \begin{pgfonlayer}{arrowlayer}
        \draw [->] (timestep) -- (t_emb) -- (y_plus);
        \draw [->] (pool_conditioning) -- (y_plus);
        \draw [->] (y_plus) -- (y_final);
    \end{pgfonlayer}

    \node [input, below right=2.15cm and 2.5cm of text] (image) {Noised Latent};

    \node [fnblock, below=0.5cm of image] (patching) {Patching};
    \node [trainableblock, below=0.2cm of patching] (patch_proj) {Linear};
    \node [operation, below=0.5cm of patch_proj] (image_plus_pos) {$+$};
    \node [trainableblock, left=0.2cm of image_plus_pos,align=center] (pos_emb) {Positional\\ Embedding};

    \node [tensor, below=0.5cm of image_plus_pos] (x_in) {$x$};
    \begin{pgfonlayer}{arrowlayer}
        \draw [->] (image) -- (image_plus_pos);
        \draw [->] (pos_emb) -- (image_plus_pos);
        \draw [->] (image_plus_pos) -- (x_in);
    \end{pgfonlayer}

    \node [trainableblock, below=1.5cm of $(c_in)!0.5!(x_in)$, minimum width=5cm] (attn1) {\emph{MM-DiT}-Block 1};
    \node [trainableblock, below=0.3cm of attn1, minimum width=5cm] (attn2) {\emph{MM-DiT}-Block 2};
    \node [below=0.4cm of attn2, minimum width=5cm] {$\ldots$};
    \node [trainableblock, below=1.3cm of attn2, minimum width=5cm] (attnN) {\emph{MM-DiT}-Block $d$};

    \coordinate (attn1_mid_left) at ($(attn1.north west)!0.5!(attn1.north)$);
    \coordinate (attn1_mid_right) at ($(attn1.north east)!0.5!(attn1.north)$);

    \coordinate (attnN_mid_left) at ($(attnN.north west)!0.5!(attnN.north)$);
    \coordinate (attnN_mid_right) at ($(attnN.north east)!0.5!(attnN.north)$);

    \node [trainableblock, below=1.5cm of attnN_mid_right] (out_mod) {Modulation};
    \node [trainableblock, below=0.2cm of out_mod] (out_mlp) {Linear};
    \node [fnblock, below=0.2cm of out_mlp] (depatch) {Unpatching};
    \node [input, below=0.5cm of depatch] (output) {Output};

    \coordinate (attn1_left) at ($(attn1.west)+(-2, 0)$);

    \begin{pgfonlayer}{arrowlayer}
        \draw [->] (x_in) -| (attn1_mid_right) -- (output);
        \draw [->] (c_in) -| (attn1_mid_left) -- (attnN_mid_left);

        \draw [->] (y_final) -| (attn1_left) -- (attn1);
        \draw [->] (attn1_left) |- (attn2);
        \draw [->] (attn1_left) |- (attnN);
        \draw [->] (attn1_left) |- (out_mod);
    \end{pgfonlayer}

    \begin{pgfonlayer}{bg}
        \draw [fill=blue!5] ($(attn1.north east)+(1, 0.5)$) -| ($(attnN.south west)+(-1, -0.5)$) -| cycle;
    \end{pgfonlayer}

\end{tikzpicture}}
        \caption{Overview of all components.}
        \label{fig:modelfig:overview_mmdit_only}
    \end{subfigure}\hfill%
    \begin{subfigure}[t]{0.49\textwidth}\centering
        \resizebox{!}{0.46\textheight}{\input{tikz/mm_block2}}
        \caption{One \modelname block}
        \label{fig:modelfig:mm}
    \end{subfigure}
    \caption{\textbf{Our model architecture.} Concatenation is indicated by $\odot$ and element-wise multiplication by $*$. The RMS-Norm for $Q$ and $K$ can be added to stabilize training runs. Best viewed zoomed in. \vspace{-1em}}
    \label{fig:modelfig:total}
\end{figure*}

\vspace{-0.5em}
For text-conditional sampling of images, our model has to take both modalities,
text and images, into account. We use pretrained models to derive suitable
representations and then describe the architecture of our diffusion backbone.
An overview of this is presented in \cref{fig:modelfig:total}.

Our general setup follows LDM \citep{Rombach_2022} for training text-to-image models in the latent space of a pretrained autoencoder.
Similar to the encoding of images to latent representations, we also follow
previous approaches \cite{saharia2022photorealistic, balaji2022ediffi} and encode the text conditioning $c$ using
pretrained, frozen text models. Details can be found in \Cref{suppsec:detailsonreps}.

\textbf{Multimodal Diffusion Backbone}
Our architecture builds upon the DiT \cite{Peebles_2023}
architecture. DiT only considers class conditional image generation and uses a
modulation mechanism to condition the network on both the timestep of the
diffusion process and the class label. Similarly, we use
embeddings of the timestep $t$ and $c_\text{vec}$ as inputs to the
modulation mechanism. However, as the pooled text representation retains only
coarse-grained information about the text input \cite{podell2023sdxl}, the network also
requires information from the sequence representation $c_\text{ctxt}$.

We construct a sequence consisting of embeddings of the text and image inputs.
Specifically, we add positional encodings and flatten $2\times 2$ patches of
the latent pixel representation $x \in \RR^{h \times w \times c}$ to a patch
encoding sequence of length $\frac{1}{2} \cdot h \cdot \frac{1}{2} \cdot w$. 
After embedding this
patch encoding and the text encoding $c_\text{ctxt}$ to a common
dimensionality, we concatenate the two sequences. We then follow DiT and apply
a sequence of modulated attention and MLPs. %

Since text and image embeddings are conceptually quite different, we use two
separate sets of weights for the two modalities. As shown in \cref{fig:modelfig:mm}, this
is equivalent to having two independent transformers for each modality, but
joining the sequences of the two modalities for the attention operation, such
that both representations can work in their own space yet take the other one
into account.

For our scaling experiments, we parameterize the size of the model in terms of
the model's depth $d$, \ie the number of attention blocks, by setting the
hidden size to $64\cdot d$ (expanded to $4\cdot64\cdot d$ channels in the MLP
blocks), and the number of attention heads equal to $d$.

\section{Experiments}
\label{sec:experiments}
\tabglobalrank
\tabsota

\subsection{Improving Rectified Flows}
\label{subsec:improvingrfs}
We aim to understand which of the approaches for simulation-free training of
normalizing flows as in \eqref{eq:ode} is the most efficient.  To enable
comparisons across different approaches, we control for the
optimization algorithm, the model architecture, the dataset and samplers. In
addition, the losses of different approaches are incomparable and also do not
necessarily correlate with the quality of output samples; hence we need
evaluation metrics that allow for a comparison between approaches.
We train models on ImageNet~\cite{Russakovsky2014ImageNetLS} and CC12M~\cite{Changpinyo2021Conceptual1P}, 
and evaluate both the training and the EMA weights of
the models during training using validation losses, CLIP scores
\cite{radford2021learning,Hessel_2021}, and
FID \cite{heusel2017gans} under different sampler settings (different guidance scales and sampling steps). 
We calculate the FID on CLIP features as proposed by~\cite{sauer2021projected}. 
All metrics are evaluated on the COCO-2014 validation split~\cite{Lin_2014}.
Full details on the training and sampling hyperparameters are provided in \Cref{subsubsec:exp_prelim}.

\subsubsection{Results}
We train each of 61 different formulations on the two datasets. We include the
following variants from \Cref{subsec:variants}:
\begin{itemize}[noitemsep,topsep=0pt,parsep=0pt,partopsep=0pt]
  \item Both $\epsilon$- and \textbf{v}-prediction loss with linear
    (\texttt{eps/linear}, \texttt{v/linear}) and cosine
    (\texttt{eps/cos}, \texttt{v/cos}) schedule.
  \item RF loss with $\pi_{\text{mode}}(t; s)$ (\texttt{rf/mode(s)}) with 7 values for $s$ chosen
    uniformly between $-1$ and $1.75$, and additionally for $s=1.0$ and $s=0$ which
    corresponds to uniform timestep sampling (\texttt{rf/mode}).
  \item RF loss with $\pi_{\text{ln}}(t; m, s)$ (\texttt{rf/lognorm(m, s)})
    with 30 values for $(m, s)$ in the grid with $m$ uniform between $-1$ and
    $1$, and $s$ uniform between $0.2$ and $2.2$.
  \item RF loss with $\pi_{\text{CosMap}}(t)$ (\texttt{rf/cosmap}).
  \item EDM (\texttt{edm($P_m, P_s$)}) with 15 values for $P_m$ chosen
    uniformly between $-1.2$ and $1.2$ and $P_s$ uniform between $0.6$ and
    $1.8$. Note that $P_m, P_s=(-1.2, 1.2)$ corresponds to the parameters
    in \cite{Karras2022ElucidatingTD}.
  \item EDM with a schedule such that it matches the log-SNR weighting of
    \texttt{rf} (\texttt{edm/rf}) and one that matches the log-SNR weighting of
    \texttt{v/cos} (\texttt{edm/cos}).
\end{itemize}

For each run, we select the step with minimal validation loss when evaluated with EMA weights and then collect CLIP scores and FID obtained with 6 different
sampler settings both with and without EMA weights.

For all 24 combinations of sampler settings, EMA weights, and dataset choice, we
rank the different formulations using a non-dominated sorting algorithm. For
this, we repeatedly compute the variants that are Pareto optimal according to
CLIP and FID scores, assign those variants the current iteration index, remove
those variants, and continue with the remaining ones until all variants get
ranked. Finally, we average those ranks over the 24 different control settings.

We present the results in Tab.~\ref{tab:globalranking}, where we only show the
two best-performing variants for those variants that were evaluated with different
hyperparameters. We also show ranks where we restrict the averaging over
sampler settings with 5 steps and with 50 steps.

We observe that \texttt{rf/lognorm(0.00, 1.00)} consistently achieves a good
rank. It outperforms a rectified flow formulation with uniform timestep
sampling (\texttt{rf}) and thus confirms our hypothesis that
intermediate timesteps are more important. Among all the variants, \emph{only}
rectified flow formulations with modified timestep sampling perform better than
the LDM-Linear~\citep{Rombach_2022} formulation (\texttt{eps/linear}) used previously.

We also observe that some variants perform well in some settings but worse in
others, \eg \texttt{rf/lognorm(0.50, 0.60)} is the best-performing variant with
50 sampling steps but much worse (average rank 8.5) with 5 sampling steps. We
observe a similar behavior with respect to the two metrics in
Tab.~\ref{tab:tabsota}. The first group shows representative variants and their
metrics on both datasets with 25 sampling steps. The next group shows the
variants that achieve the best CLIP and
FID scores. With the exception of \texttt{rf/mode(1.75)}, these
variants typically perform very well in one metric but relatively badly
in the other. In contrast, we once again observe that \texttt{rf/lognorm(0.00,
1.00)} achieves good performance across metrics and datasets, where it obtains
the third-best scores two out of four times and once the second-best
performance.

Finally, we illustrate the qualitative behavior of different formulations in
\Cref{fig:fidvssteps}, where we use different colors for different groups
of formulations (\textcolor[HTML]{d95f02}{\texttt{edm}},
\textcolor[HTML]{1b9e77}{\texttt{rf}}, \textcolor[HTML]{7570b3}{\texttt{eps}}
and \textcolor[HTML]{e7298a}{\texttt{v}}).
Rectified flow formulations generally perform well and, compared to other
formulations, their performance degrades less when reducing the number of
sampling steps.

\figfidvssteps

\subsection{Improving Modality Specific Representations}
\label{subsec:modalityspecifics}
Having found a formulation in the previous section that allows rectified flow models to not only 
compete with established diffusion formulations such as LDM-Linear~\citep{Rombach_2022} 
or EDM~\citep{Karras2022ElucidatingTD}, but even outperforms them, 
we now turn to the application of our formulation to high-resolution text-to-image synthesis. 
Accordingly, the final performance of our algorithm depends not only 
on the training formulation, but also on the parameterization via a neural network
and the quality of the image and text representations we use. 
In the following sections, we describe how we improve all these components before scaling our final method in \Cref{subsec:scaling}. 

\subsubsection{Improved Autoencoders}
\label{sec:improvedae}
Latent diffusion models achieve high efficiency by operating in the latent space of a pretrained autoencoder~\citep{Rombach_2022}, 
which maps an input RGB  $X\in \RR^{H\times W\times 3}$ into a lower-dimensional space $x=E(X) \in \RR^{h\times w \times d}$.
The reconstruction quality of this autoencoder provides an upper bound on the achievable image quality after latent diffusion training. 
Similar to \citet{dai2023emu}, we find that increasing the number of latent channels $d$ significantly boosts reconstruction performance, see \Cref{tab:aereconstructiontable}. Intuitively, predicting latents with higher $d$ is a more difficult task, and thus models with increased capacity should be able to perform better for larger $d$, ultimately achieving higher image quality. We confirm this hypothesis in \Cref{fig:aefidstudy}, where we see that the $d=16$ autoencoder exhibits better scaling performance in terms of sample FID. For the remainder of this paper, we thus choose $d=16$. 

\aereconstructiontable
\horizontalcherries

\subsubsection{Improved Captions}
\label{subsubsec:improvedcaptions}
\citet{betker2023improving} demonstrated that synthetically generated captions can greatly improve text-to-image models trained at scale. This is due to the oftentimes simplistic 
nature of the human-generated captions that come with large-scale image datasets, which overly focus on the image subject and usually omit details describing the background or composition of the scene, or, if applicable, displayed text~\citep{betker2023improving}. 
We follow their approach and use an off-the-shelf, state-of-the-art vision-language model, \emph{CogVLM}~\citep{wang2023cogvlm}, to create synthetic annotations for our large-scale image dataset. As synthetic captions may cause a text-to-image model to forget about certain concepts not present in the VLM's knowledge corpus, we use a  ratio of 50 \% original and 50 \% synthetic captions. 

To assess the effect of training on this caption mix, we train two $d=15$ \modelname models for 250k steps, one on only original captions and the other on the 50/50 mix. 
We evaluate the trained models using the GenEval benchmark~\citep{ghosh2023geneval} in \Cref{tab:captionstudy}. The results demonstrate that the model trained with the addition of synthetic captions clearly outperforms the model that only utilizes original captions. We thus use the 50/50 synthetic/original caption mix for the remainder of this work. 
\captionstudy

\subsubsection{Improved Text-to-Image Backbones}
\label{sec:netbattle}

In this section, we compare the performance of existing transformer-based diffusion backbones with our novel multimodal transformer-based diffusion backbone, \modelname, as introduced in \Cref{subsec:methodarch}. \modelname is specifically designed to handle different domains, here text and image tokens, using (two) different sets of trainable model weights. More specifically, we follow the experimental setup from \Cref{subsec:improvingrfs} and compare text-to-image performance on CC12M of DiT, CrossDiT (DiT but with cross-attending to the text tokens instead of sequence-wise concatenation~\citep{chen2023pixart}) and our \modelname. For \modelname, we compare models with two sets of weights and three sets of weights, where the latter handles the CLIP~\citep{radford2021learning} and T5~\citep{raffel2019exploring} tokens (\cf \Cref{subsec:methodarch}) separately. 
Note that DiT (w/ concatenation of text and image tokens as in \Cref{subsec:methodarch}) can be interpreted as a special case of \modelname with one shared set of weights for all modalities. Finally, we consider the UViT~\citep{hoogeboom2023simple} architecture as a hybrid between the widely used UNets and transformer variants. 
We analyze the convergence behavior of these architectures in \Cref{fig:archstraining}:
Vanilla DiT underperforms UViT. The cross-attention DiT variant CrossDiT achieves better performance than UViT, although UViT seems to learn much faster initially. 
Our \modelname variant significantly outperforms the cross-attention and vanilla variants. 
We observe only a small gain when using three parameter sets instead of two (at the cost of increased parameter count and VRAM usage), and thus opt for the former option for the remainder of this work. 

\archstrainingsqueezed

\subsection{Training at Scale}
\label{subsec:scaling}
Before scaling up, we filter and preencode our data to ensure safe and efficient pretraining.
Then, all previous considerations of diffusion formulations, architectures, and data culminate in the last section, where we scale our models up to 8B parameters.

\subsubsection{Data Preprocessing}
\paragraph{Pre-Training Mitigations}
Training data significantly impacts a generative model's abilities. Consequently, data filtering is effective at constraining undesirable capabilities~\citep{dalle2mitigations}. 
Before training at sale, we filter our data for the following categories:
(i) Sexual content: We use NSFW-detection models to filter for explicit content. 
(ii) Aesthetics: We remove images for which our rating systems predict a low score.
(iii) Regurgitation: We use a cluster-based deduplication method to remove perceptual and semantic duplicates from the training data; see \Cref{sec:dedup}.

\paragraph{Precomputing Image and Text Embeddings}
Our model uses the output of multiple pretrained, frozen networks as inputs (autoencoder latents and text encoder representations). %
Since these outputs are constant during training, we precompute them once for the entire dataset. We provide a detailed discussion of our approach in \Cref{suppsec:precomputingembeddings}.

\subsubsection{Finetuning on High Resolutions}
\label{subsec:shifting}
\maxattnlogitqk
\paragraph{QK-Normalization} 
In general, we pretrain all of our models on low-resolution images of size $256^2$ pixels. 
Next, we finetune our models on higher resolutions with mixed aspect ratios (see next paragraph for details). 
We find that, when moving to high resolutions, mixed precision training can become unstable and the loss diverges.
This can be remedied by switching to full precision training --- but comes with a $\sim 2\times$ performance drop compared to mixed-precision training. 
A more efficient alternative is reported in the (discriminative) ViT literature: \citet{dehghani2023scaling} observe that the training of large vision transformer models diverges because the attention entropy grows uncontrollably. To avoid this, \citet{dehghani2023scaling} propose to normalize Q and K before the attention operation. We follow this approach and use RMSNorm~\citep{zhang2019root} with learnable scale in both streams of our MMDiT architecture for our models, see \Cref{fig:modelfig:total}.
As demonstrated in \Cref{fig:qk_vis}, the additional normalization prevents the attention logit growth instability, confirming findings by \citet{dehghani2023scaling} and \citet{wortsman2023smallscale} and enables efficient training at bf16-mixed~\citep{bfloat16} precision when combined with $\epsilon=10^{-15}$ in the AdamW~\citep{Loshchilov2017FixingWD} optimizer.
This technique can also be applied on pretrained models that have not used qk-normalization during pretraining: The model quickly adapts to the additional normalization layers and trains more stably. 
Finally, we would like to point out that although this method can generally help to stabilize the training of large models, it is not a universal recipe and may need to be adapted depending on the exact training setup.

\paragraph{Positional Encodings for Varying Aspect Ratios}
After training on a fixed $256\times256$ resolution we aim to (i) increase the
resolution and resolution and (ii) enable inference with flexible aspect
ratios. Since we use 2d positional frequency embeddings we have to adapt them
based on the resolution. In the multi-aspect ratio setting, a direct
interpolation of the embeddings as in \cite{dosovitskiy2020image} would not reflect the side
lengths correctly. Instead we use a combination of extended and interpolated
position grids which are subsequently frequency embedded.

For a target resolution of $S^2$ pixels, we use bucketed sampling
\cite{novelai,podell2023sdxl} such that
that each batch consists of images of a homogeneous size $H \times W$, where $H \cdot W
\approx S^2$.
For the maximum and minimum training aspect ratios,
this results in the maximum values for width, $W_\text{max}$, and height,
$H_\text{max}$, that will be encountered. Let $h_\text{max}=H_\text{max}/16, w_\text{max}=W_\text{max}/16$ and $s=S/16$ be the corresponding sizes in latent space (a factor 8) after patching (a factor 2).
Based on these values, we construct a vertical position
grid with the values
$((p-\frac{h_\text{max}-s}{2})\cdot\frac{256}{S})_{p=0}^{{h_\text{max}-1}}$
and correspondingly for the horizontal positions. We then
center-crop from the resulting positional 2d grid before embedding it.

\figtimeshift

\paragraph{Resolution-dependent shifting of timestep schedules}
Intuitively, since higher resolutions have more pixels, we need more noise to
destroy their signal. %
Assume we are working in a resolution with $n=H\cdot W$ pixels. Now, consider a
"constant" image, i.e. one where every pixel has the value $c$. The forward
process produces $z_t = (1-t)c \mathbbm{1} + t \epsilon$, where both
$\mathbbm{1}$ and $\epsilon \in \mathbb{R}^n$. Thus, $z_t$ provides $n$
observations of the random variable $Y=(1-t)c + t \eta$ with $c$ and $\eta$ in
$\mathbb{R}$, and $\eta$ follows a standard normal distribution. Thus,
$\mathbb{E}(Y)=(1-t)c$ and $\sigma(Y)=t$. We can therefore recover $c$ via
$c=\frac{1}{1-t}\mathbb{E}(Y)$, and the error between $c$ and its sample
estimate $\hat{c} = \frac{1}{1-t}\sum_{i=1}^n z_{t,i}$ has a standard deviation
of $\sigma(t, n)=\frac{t}{1-t}\sqrt{\frac{1}{n}}$ (because the standard error
of the mean for $Y$ has deviation $\frac{t}{\sqrt{n}}$).  So if one already
knows that the image $z_0$ was constant across its pixels, $\sigma(t, n)$
represents the degree of uncertainty about $z_0$. For example, we immediately
see that doubling the width and height leads to half the uncertainty at any
given time $0 < t < 1$.  But, we can now map a timestep $t_n$ at resolution $n$
to a timestep $t_m$ at resolution $m$ that results in the same degree of
uncertainty via the ansatz $\sigma(t_n, n)=\sigma(t_m, m)$. Solving for $t_m$
gives
\begin{equation}
  \label{eq:timeshift}
t_m = \frac{\sqrt{\frac{m}{n}}t_n}{1+(\sqrt{\frac{m}{n}}-1)t_n}
\end{equation}
\sotaeval
We visualize this shifting function in \Cref{fig:timeshift}. Note that the
assumption of constant images is not realistic. To find good values for the
shift value $\alpha\coloneq\sqrt{\frac{m}{n}}$ during inference, we apply them to the
sampling steps of a model trained at resolution $1024\times 1024$ and run a
human preference study. The results in \Cref{fig:timeshift} show a strong
preference for samples with shifts greater than $1.5$ but less drastic
differences among the higher shift values. In our subsequent experiments, we
thus use a shift value of $\alpha=3.0$ both during training and sampling at
resolution $1024\times 1024$. A qualitative comparison between samples after 8k
training steps with and without such a shift can be found in
\Cref{fig:timeshift}. Finally, note that \eqref{eq:timeshift} implies a 
log-SNR shift of $\log \frac{n}{m}$ similar to \cite{hoogeboom2023simple}: 
 \begin{align}
     \lambda_{t_m} &= 2 \log \frac{1-t_n}{\sqrt{\frac{m}{n}} t_n} \\
     &= \lambda_{t_n} - 2 \log \alpha = \lambda_{t_n} - \log \frac{m}{n} \; .
 \end{align}

After the shifted training at resolution $1024 \times 1024$, we align the model
using Direct Preference Optimization (DPO) as described in
\Cref{supsec:dpo}.

\subsubsection{Results}\label{subsec:results}
\figvalscaling
In \Cref{fig:valscaling}, we examine the effect of training our \modelname at scale. For images, we conduct a large scaling study and train models with different numbers of parameters for 500k steps on $256^2$ pixels resolution using preencoded data, \cf \Cref{suppsec:precomputingembeddings}, with a batch size of 4096. We train on $2\times2$ patches~\citep{Peebles_2023}, and report validation losses on the CoCo dataset~\citep{Lin_2014} every 50k steps. In particular, to reduce noise in the validation loss signal, we sample loss levels equidistant in $t \in (0,1)$ and compute validation loss for each level separately. We then average the loss across all but the last ($t=1$) levels.

Similarly, we conduct a preliminary scaling study of our \modelname on videos.
To this end we start from the pretrained image weights and additionally use a
2x temporal patching. We follow \citet{blattmann2023align} and feed data to the
pretrained model by collapsing the temporal into the batch axis. In each
attention layer we rearrange the representation in the visual stream and add a
full attention over all spatio-temporal tokens after the spatial attention
operation before the final feedforward layer. Our video models are trained for
140k steps with a batch size of 512 on videos comprising 16 frames with $256^2$
pixels. We report validation losses on the Kinetics
dataset~\cite{carreira2018quo} every 5k steps. Note that our reported FLOPs for
video training in \Cref{fig:valscaling} are only FLOPs from video training and
do not include the FLOPs from image pretraining.

For both the image and video domains, we observe a smooth decrease in the
validation loss when increasing model size and training steps. We find the validation loss to be highly correlated to comprehensive evaluation metrics
(CompBench~\cite{huang2023t2i}, GenEval~\cite{ghosh2023geneval}) and to human preference.
These results support the validation loss as a simple and general measure of
model performance. Our results do not show saturation neither for image not for video models.

\Cref{fig:qualitativescaleparti} illustrates how training a larger model for longer impacts sample quality.
Tab.~\ref{tab:genEvalSotaTable} shows the results of GenEval in full. When applying the methods presented in \Cref{subsec:shifting} and increasing training image resolution, our biggest model excels in most categories and
outperforms DALLE~3~\cite{betker2023improving}, the current state of the art in prompt comprehension, in overall score. 
\dropthememb

Our $d=38$ model outperforms current proprietary~\cite{betker2023improving,ideogramv1} and open~\cite{sauer2023adversarial,playgroundv25,chen2023pixart,pernias2023wuerstchen} SOTA generative image models in human preference evaluation on the Parti-prompts benchmark~\cite{yu2022scaling} in the categories \emph{visual aesthetics}, \emph{prompt following} and \emph{typography generation}, \cf \Cref{fig:sota_human_eval}. For evaluating human preference in these categories, raters were shown pairwise outputs from two models, and asked to answer the following questions: \\
\textbf{Prompt following:} \textit{Which image looks more \emph{representative} to the \emph{tex}t shown above and \emph{faithfully} follows it?} \\
\textbf{Visual aesthetics:} \textit{Given the prompt, which image is of \emph{higher-quality} and \emph{aesthetically more pleasing}?} \\
\textbf{Typography:} \textit{Which image more accurately shows/displays the text specified in the above description? More accurate spelling is preferred! Ignore other aspects.}

Lastly,~\Cref{tab:biggerismorestepefficient} highlights an intriguing result: not only do bigger models perform better, they also require fewer steps to reach their peak performance.

\genEvalSotaTable
\biggerismorestepefficient

\paragraph{Flexible Text Encoders}
While the main motivation for using multiple text-encoders is boosting the overall model performance~\cite{balaji2022ediffi}, we now show that this choice additionally increases the flexibility of our \modelname-based rectified flow during inference.
As described in \Cref{subsubsec:exp_prelim} we train our model with three text encoders, with an individual drop-out rate of 46.3\%. 
Hence, at inference time, we can use an arbitrary subset of all three text
encoders. This offers means for trading off model performance for improved
memory efficiency, which is particularly relevant for the 4.7B parameters of
T5-XXL~\citep{raffel2019exploring} that require significant amounts of VRAM.
Interestingly, we observe limited performance drops when using only the two
CLIP-based text-encoders for the text prompts and replacing the T5 embeddings
by zeros. We provide a qualitative visualization in
\Cref{fig:embedder_dropping}. Only for complex prompts involving either highly
detailed descriptions of a scene or larger amounts of written text do we find
significant performance gains when using all three text-encoders. These
observations are also verified in the human preference evaluation results in
\Cref{fig:sota_human_eval} (\emph{Ours w/o T5}).
Removing T5 has no effect on aesthetic quality ratings ($50\%$ win rate), and only a small impact on prompt adherence ($46\%$ win rate), whereas its contribution to the capabilities of generating written text are more significant ($38\%$ win rate).

\section{Conclusion}
\label{sec:conclusion}
In this work, we presented a scaling analysis of rectified flow models for
text-to-image synthesis. We proposed a novel timestep sampling for rectified
flow training that improves over previous diffusion training formulations for
latent diffusion models and retains the favourable properties of rectified
flows in the few-step sampling regime. We also demonstrated the advantages of
our transformer-based \modelname architecture that takes the multi-modal nature of the
text-to-image task into account. Finally, we performed a scaling study of this
combination up to a model size of 8B parameters and $5\times 10^{22}$ training
FLOPs. We showed that validation loss improvements correlate with both existing
text-to-image benchmarks as well as human preference evaluations. This, in
combination with our improvements in generative modeling and scalable, multimodal
architectures achieves performance that is competitive with state-of-the-art
proprietary models. 
The scaling trend shows no signs of saturation, which makes us optimistic
that we can continue to improve the performance of our models in the future.
\FloatBarrier
\section*{Broader Impact}
This paper presents work whose goal is to advance the field of machine learning in general and image synthesis in particular. 
There are many potential societal consequences of our work, none of which we feel must be specifically highlighted here.
For an extensive discussion of the general ramifications of diffusion models, we point interested readers towards~\cite{po2023state}.

\bibliography{references}
\bibliographystyle{icml2024}

\newpage
\appendix
\onecolumn

\pagebreak

\begin{center}
\textbf{\large Supplementary}
\end{center}

\horizontalcherriesthree
\horizontalcherriesgrids

\section{Background}
\label{sec:background}

\paragraph{Diffusion Models}\citep{SohlDickstein2015DeepUL,Song2020ScoreBasedGM,ho2020denoising} generate data by approximating the reverse ODE to a stochastic forward process which transforms data to noise. They have become the standard approach for generative modeling of  images~\cite{dhariwal2021diffusion,ramesh2022hierarchical,saharia2022photorealistic,Rombach_2022,balaji2022ediffi} and videos~\citep{singer2022makeavideo,ho2022imagen,esser2023structure, blattmann2023align,gupta2023photorealistic}. Since these models can be derived both via a variational lower bound on the negative likelihood~\citep{SohlDickstein2015DeepUL} and score matching~\citep{Hyvrinen2005EstimationON,Vincent2011ACB,song2020generative}, 
various formulations of forward- and reverse processes~\citep{Song2020ScoreBasedGM,dockhorn2021score}, model parameterizations~\citep{ho2020denoising,ho2022classifierfree,Karras2022ElucidatingTD}, loss weightings~\citep{ho2020denoising, Karras2022ElucidatingTD} and ODE 
solvers~\citep{song2022denoising,lu2023dpmsolver, dockhorn2022genie} have led to a large number of different training objectives and sampling procedures. More recently, the seminal works of \citet{Kingma2023UnderstandingDO} and \citet{Karras2022ElucidatingTD} have proposed unified formulations and introduced new theoretical and practical insights for training~\cite{Karras2022ElucidatingTD,Kingma2023UnderstandingDO} and inference~\cite{Karras2022ElucidatingTD}. However, despite these improvements, the trajectories of common ODEs involve partly significant amounts of curvature~\citep{Karras2022ElucidatingTD,liu2022flow}, which requires increased amounts of solver steps and, thus, renders fast inference difficult. To overcome this, we adopt rectified flow models whose formulation allows for learning straight ODE trajectories.  

\paragraph{Rectified Flow Models}
\cite{liu2022flow,albergo2022building,lipman2023flow} approach generative modeling by
constructing a transport map between two distributions through an ordinary
differential equation (ODE). This approach has close connections to continuous
normalizing flows (CNF) \cite{Chen2018NeuralOD} as well as diffusion models.
Compared to CNFs, Rectified Flows and Stochastic Interpolants have the
advantage that they do not require simulation of the ODE during training.
Compared to diffusion models, they can result in ODEs that are faster to
simulate than the probability flow ODE \cite{Song2020ScoreBasedGM} associated
with diffusion models. Nevertheless, they do not result in optimal transport
solutions, and multiple works aim to minimize the trajectory curvature further
\cite{lee2023minimizing,alex2023improving,aramalex2023multisample}.
\cite{dao2023flow,ma2024sit} demonstrate the feasibility of rectified flow
formulations for class-conditional image synthesis, \cite{fischer2023boosting} for
latent-space upsampling, and \cite{liu2023instaflow}
apply the reflow procedure of \cite{liu2022flow} to distill a pretrained
text-to-image model \cite{Rombach_2022}. Here, we are interested in rectified
flows as the foundation for text-to-image synthesis with fewer sampling steps.
We perform an extensive comparison between different formulations and loss
weightings and propose a new timestep schedule for training of
rectified flows with improved performance.

\paragraph{Scaling Diffusion Models}
The transformer architecture~ \cite{vaswani2017attention} is well known for its scaling properties in NLP~\cite{kaplan2020scaling} and computer vision tasks~\cite{dosovitskiy2020image,zhai2022scaling}. For diffusion models, U-Net architectures~\cite{Ronneberger_2015} have been the  dominant choice~\cite{ho2020denoising,Rombach_2022,balaji2022ediffi}.
While some recent works explore diffusion transformer backbones~\cite{Peebles_2023,chen2023pixart,ma2024sit}, scaling laws for text-to-image diffusion models remain unexplored.

\newcommand*\diff{\mathop{}\!\mathrm{d}}
\newcommand*\Diff[1]{\mathop{}\!\mathrm{d^#1}}
\DeclarePairedDelimiter\bra{\langle}{\rvert}
\DeclarePairedDelimiter\ket{\lvert}{\rangle}
\DeclarePairedDelimiterX\braket[2]{\langle}{\rangle}{#1\,\delimsize\vert\,\mathopen{}#2}

\newpage
 \section{On Flow Matching}
 \subsection{Details on Simulation-Free Training of Flows}
 \label{subsec:flowproofs}

Following~\citep{lipman2023flow}, to see that $u_t(z)$ generates $p_t$, we note that the continuity equation provides a necessary and sufficient condition \cite{Villani2008OptimalTO}:
\begin{align}
     & \frac{d}{dt} p_t(x) + \nabla \cdot [p_t(x) v_t(x)] = 0 \leftrightarrow  \text{$v_t$ generates probability density path $p_t$}. \label{eq:cont_eq}
\end{align}

Therefore it suffices to show that
\begin{align}
    - \nabla \cdot [u_t(z) p_t(z)]
      &=
    - \nabla \cdot [\mathbb{E}_{\epsilon \sim
   \mathcal{N}(0,I)} u_t(z \vert \epsilon) \frac{p_t(z \vert
   \epsilon)}{p_t(z)}  p_t(z) ] \label{eq:marg_cont_line1}\\
     &=  \mathbb{E}_{\epsilon \sim \mathcal{N}(0,I)} - \nabla \cdot  [u_t(z \vert \epsilon) p_t(z \vert \epsilon) ] \label{eq:marg_cont_line2} \\
     &= \mathbb{E}_{\epsilon \sim \mathcal{N}(0,I)}  \frac{d}{dt} p_t(z \vert \epsilon) \label{eq:marg_cont_line3}  = \frac{d}{dt} p_t(z),
 \end{align}
 where we used the continuity equation \Cref{eq:cont_eq} for $u_t(z \vert \epsilon)$ in line \Cref{eq:marg_cont_line2} to \Cref{eq:marg_cont_line3} since $u_t(z \vert \epsilon)$ generates $p_t(z \vert \epsilon)$ and the definition of \Cref{eq:marginal_u} in line \Cref{eq:marg_cont_line1}

 The equivalence of objectives $\mathcal{L}_{FM} \leftrightharpoons \mathcal{L}_{CFM}$~\citep{lipman2023flow} follows from
 \begin{align}
 \mathcal{L}_{FM}(\Theta) &= \mathbb{E}_{t, p_t(z)}|| v_{\Theta}(z, t) - u_t(z)  ||_2^2 \\
     &=  \mathbb{E}_{t, p_t(z)}|| v_{\Theta}(z, t) ||_2^2  - 2 \mathbb{E}_{t, p_t(z)} \braket{v_{\Theta}(z, t) }{  u_t(z) } + c
     \label{eq:equiv_line2} \\
       &=\mathbb{E}_{t, p_t(z)}|| v_{\Theta}(z, t) ||_2^2  - 2 \mathbb{E}_{t, p_t(z | \epsilon),  p(\epsilon) } \braket{v_{\Theta}(z, t) }{u_t(z | \epsilon )} + c \label{eq:equiv_line3} \\
      &=   \mathbb{E}_{t, p_t(z | \epsilon), p(\epsilon) }|| v_{\Theta}(z, t) - u_t(z | \epsilon)  ||_2^2 + c'  =   \mathcal{L}_{CFM}(\Theta) + c'
 \end{align}
 where $c, c'$ do not depend on $\Theta$ and line \Cref{eq:equiv_line2} to line \Cref{eq:equiv_line3} follows from:
 \begin{align}
 \mathbb{E}_{p_t(z | \epsilon),  p(\epsilon) } \braket{v_{\Theta}(z, t) }{u_t(z | \epsilon )}
 &=  \int \diff z  \int \diff \epsilon p_t(z | \epsilon) p(\epsilon)\braket{v_{\Theta}(z, t) }{u_t(z | \epsilon )} \\
 &=  \int \diff z p_t(z) \braket{v_{\Theta}(z, t) }{ \int \diff \epsilon \frac{ p_t(z| \epsilon)}{p_t(z)} p(\epsilon) u_t(z | \epsilon )} \label{eq:equiv_intergral_line2} \\
 &=  \int \diff z p_t(z)  \braket{v_{\Theta}(z, t) }{  u_t(z) } =  \mathbb{E}_{p_t(z)} \braket{v_{\Theta}(z, t) }{  u_t(z) } \label{eq:equiv_intergral_line3} 
 \end{align}
 where we extended with $\frac{p_t(z)}{p_t(z)}$ in line \Cref{eq:equiv_intergral_line2} and used the definition of \Cref{eq:marginal_u} in line \Cref{eq:equiv_intergral_line2} to \Cref{eq:equiv_intergral_line3}.

\subsection{Details on Image and Text Representations}
\label{suppsec:detailsonreps}
\textbf{Latent Image Representation}
We follow LDM \cite{Rombach_2022} and use a pretrained autoencoder to represent RGB images $X
\in \RR^{H\times W\times 3}$ in a smaller latent space $x=E(X) \in \RR^{h\times
w \times d}$. We use a spatial downsampling factor of $8$, such that
$h=\frac{H}{8}$ and $w=\frac{W}{8}$, and experiment with different values for
$d$ in \Cref{sec:improvedae}. We always apply the forward process
from \eqref{eq:forwardprocess} in the latent space, and when sampling a %
representation $x$ via \eqref{eq:ode}, we decode it back into pixel space $X =
D(x)$ via the decoder $D$.
We follow \citet{Rombach_2022} and normalize the latents by their mean and 
standard deviation, which are globally computed over a subset
of the training data. 
\Cref{fig:aefidstudy} shows how generative model training for different $d$ evolves as a function of model capacity, 
as discussed in \Cref{sec:improvedae}.
\aefidstudy

\textbf{Text Representation}
Similar to the encoding of images to latent representations, we also follow
previous approaches \cite{saharia2022photorealistic, balaji2022ediffi} and encode the text conditioning $c$ using
pretrained, frozen text models. In particular, for all experiments, we use a
combination of CLIP \cite{radford2021learning} models and a encoder-decoder text model.
Specifically, we encode $c$ with the text encoders of both a CLIP L/14 model of
\citet{radford2021learning} as well as an OpenCLIP bigG/14 model of \citet{Cherti_2023}. We
concatenate the pooled outputs, of sizes $768$ and $1280$ respectively, to
obtain a vector conditioning $c_\text{vec} \in \RR^{2048}$. We also concatenate
the penultimate hidden representations channel-wise to a CLIP context conditioning
$c_\text{ctxt}^{\text{CLIP}} \in \RR^{77\times2048}$. Next, we encode $c$ also
to the final hidden representation,
$c_\text{ctxt}^{\text{T5}}\in\RR^{77\times4096}$, of the encoder of a T5-v1.1-XXL model
\cite{raffel2019exploring}. Finally, we zero-pad $c^\text{CLIP}_\text{ctxt}$ along the channel
axis to $4096$ dimensions to match the T5 representation and concatenate it
along the sequence axis with $c_\text{ctxt}^\text{T5}$ to obtain the final
context representation $c_\text{ctxt}\in\RR^{154\times4096}$. These two caption
representations, $c_\text{vec}$ and $c_\text{ctxt}$, are used in two different
ways as described in \Cref{subsec:methodarch}.

\subsection{Preliminaries for the Experiments in \Cref{subsec:improvingrfs}.}
\label{subsubsec:exp_prelim}

\textbf{Datasets}
We use two datasets to account for the missing of a standard text-to-image
benchmark. As a widely used dataset, we convert the ImageNet dataset \cite{Russakovsky2014ImageNetLS}
into a dataset suitable for text-to-image models by adding captions of the
form ``a photo of a \texttt{\textlangle{}class name\textrangle{}}'' to images, where \texttt{\textlangle{}class name\textrangle{}} is randomly
chosen from one of the provided names for the image's class label.
As a more realistic text-to-image dataset, we use the CC12M dataset \cite{Changpinyo2021Conceptual1P} for
training.

\textbf{Optimization}
In this experiment, we train all models using a global batch size of 1024 using the AdamW optimizer
\cite{Loshchilov2017FixingWD} with a learning rate of $10^{-4}$ and 1000 linear warmup
steps. We use mixed-precision training and keep a copy of the model weights
which gets updated every 100 training batches with an exponential moving
average (EMA) using a decay factor of $0.99$. For unconditional diffusion
guidance \cite{ho2022classifierfree}, we set the outputs of each of the three text encoders
independently to zero with a probability of $46.4\%$, such that we roughly
train an unconditional model in $10\%$ of all steps.

\textbf{Evaluation}
As described in \Cref{subsec:improvingrfs}, we use CLIP scores, 
FID and validation losses to evaluate our models 
regularly during training on the COCO-2014 validation split~\cite{Lin_2014}.

As the loss values differ widely in magnitude and variance for different
timesteps, we evaluate them in a stratified way on eight equally spaced values
in the time interval $[0, 1]$.

To analyze how different approaches behave under different sampler settings, we
produce 1000 samples for each of the samplers which differ in guidance scales
as well as number of sampling steps. We evaluate these samples with CLIP scores
using CLIP L/14 \cite{radford2021learning} and also compute FID between CLIP L/14 image features of these
samples and the images of the validation set. For sampling, we always use a
Euler discretization \cite{euler1768institutionum} of \eqref{eq:ode} and six
different settings: 50 steps with classifier-free-guidance scales 1.0, 2.5,
5.0, and 5, 10, 25 steps with classifier-free-guidance scale 5.0.

\subsection{Improving SNR Samplers for Rectified Flow Models}
As described in \Cref{sec:method}, we introduce novel densities $\pi(t)$ for the timesteps
that we use to train our rectified flow models. \Cref{fig:tsamplers} visualizes the distributions of the \textbf{logit-normal sampler} and the \textbf{mode sampler} introduced in \Cref{subsubsec:tailoredsnrforrf}. Notably, as we demonstrate in \Cref{subsec:improvingrfs}, the \textbf{logit-normal sampler} outperforms the classic uniform rectified flow formulation~\citep{liu2022flow} and established diffusion baselines such as EDM~\citep{Karras2022ElucidatingTD} and LDM-Linear~\citep{Rombach_2022}.
\tsamplers

\figqualitativescaleparti

\section{Direct Preference Optimization}
\label{supsec:dpo}
\figdpo
\figdpohumaneval
Direct Preference Optimization~(DPO)~\citep{rafailov2023direct} is a technique to finetune LLMs with preference data. Recently, this method has been adapted to preference finetuning of text-to-image diffusion models~\citep{wallace2023diffusion}. In this section, we verify that our model is also amenable to preference optimization. In particular, we apply the method introduced in~\citet{wallace2023diffusion} to our 2B and 8B parameter base model. Rather than finetuning the entire model, we introduce learnable Low-Rank Adaptation (LoRA) matrices (of rank 128) for all linear layers as is common practice. We finetune these new parameters for 4k and 2k iteration for the 2B and 8B base model, respectively. We then evaluate the resulting model in a human preference study using a subset of 128 captions from the Partiprompts set~\citep{yu2022scaling} (roughly three voter per prompt and comparison). \Cref{fig:dpo_human_eval} shows that our base models can be effectively tuned for human preference. \Cref{fig:dpo} shows samples of the respective base models and DPO-finetuned models.

\section{Finetuning for instruction-based image editing}

A common approach for training instruction based image editing and general image-to-image diffusion models is to concatenate the latents of the input image to the noised latents of the diffusion target along the channel dimension before feeding the input into a U-Net ~\citep{brooks2023instructpix2pix, sheynin2023emu, saharia2022palette, saharia2022image}. We follow the same approach, concatenating input and target along the channels before patching, and demonstrate  that the same method is applicable to our proposed architecture. We finetune the 2B parameter base model on a dataset consisting of image-to-image editing tasks similar to the distribution of the InstructPix2Pix dataset \citep{brooks2023instructpix2pix} as well as inpainting, segmentation, colorization, deblurring and controlnet tasks similar to Emu Edit and Palette ~\citep{sheynin2023emu, saharia2022palette}. As shown in Fig \ref{fig:textmanipulation} we observe that the resulting 2B Edit model has the capability to manipulate text in a given image, even though no text manipulation tasks were included in the training data. We were not able to reproduce similar results when training a SDXL-based ~\citep{podell2023sdxl} editing model on the same data. 

\textmanipulation

\section{Data Preprocessing for Large-Scale Text-to-Image Training}
\label{suppsec:datepreproc}

\subsection{Precomputing Image and Text Embeddings}
\label{suppsec:precomputingembeddings}
Our model uses the output of multiple pretrained, frozen networks as inputs (autoencoder latents and text encoder representations). %
Since these outputs are constant during training, we precompute them once for the entire dataset.
This comes with two main advantages: (i) The encoders do not need to be available on the GPU during training, lowering the required memory. (ii) The forward encoding pass is skipped during training, saving time and total needed compute after the first epoch, see \cref{tab:preencodingspeedups}.

\preencodingspeedups

This approach has two disadvantages: First, random augmentation for each sample every epoch is not possible and we use square-center cropping during precomputation of image latents. For finetuning our model at higher resolutions, we specify a number of aspect ratio buckets, and resize and crop to the closest bucket first and then precompute in that aspect ratio. 
Second, the dense output of the text encoders is particularly large, creating additional storage cost and longer loading times during training (\cf \cref{tab:preencodingspeedups}).
We save the embeddings of the language models in half precision, as we do not observe a deterioration in performance in practice.

\subsection{Preventing Image Memorization}\label{sec:dedup}
In the context of generative image models memorization of training samples can lead to a number of issues~\citep{somepalli2023diffusion, carlini2023extracting, somepalli2023understanding}. 
To avoid verbatim copies of images by our trained models, we carefully scan our training dataset for duplicated examples and remove them. 

\paragraph{Details on Deduplication}
In accordance with the methods outlined by ~\citet{carlini2023extracting} and ~\citet{somepalli2023diffusion}, we opt for SSCD \citep{pizzi2022self} as the backbone for the deduplication process. The SSCD algorithm is a state-of-the-art technique for detecting near-duplicate images at scale, and it generates high-quality image embeddings that can be used for clustering and other downstream tasks. We also decided to follow~\citet{dalle2mitigations} to decide on a number of clusters $N$. For our experiments, we use $N = 16,000$.

We utilize~\citet{autofaiss} for clustering. ~\citet{autofaiss} is a library that simplifies the process of using Faiss (Facebook AI Similarity Search) for large-scale clustering tasks. Specifically, leverage FAISS index factory\footnote{\url{https://github.com/facebookresearch/faiss/wiki/The-index-factory}} functionality to train a custom index with predefined number of centroids. This approach allows for efficient and accurate clustering of high-dimensional data, such as image embeddings.

\Algref{alg:cluster_deduplication} details our deduplication approach. We ran an experiment to see how much data is removed by different SSCD threshold as shown in~\Figref{fig:sscd_ablation}. Based on these results we selected four thresholds for the final run~\Figref{fig:sscd_final_result}.

\subsection{Assessing the Efficacy of our Deduplication Efforts}
~\citet{carlini2023extracting} devise a two-stage data extraction attack that generates images using standard approaches, and flags those that exceed certain membership inference scoring criteria. \citet{carlini2023extracting} bias their search towards duplicated training examples because these are orders of magnitude more likely to be memorized than non-duplicated examples ~\citep{somepalli2023diffusion, somepalli2023diffusion,lee2021deduplicating}. 

To assess how well our SSCD-based deduplication works, we follow \citet{carlini2023extracting} to extract memorized samples from small, specifically for this purpose trained models and compare them before and after deduplication. Two main step of the mentioned procedure include: 1) Generate many examples using the diffusion model in the standard sampling manner and with the known prompts. 2) Perform membership inference to separate the model’s novel generations from those generations which are memorized training examples. \Algref{alg:memorization_detection} shows the steps to find the memorized samples based on~\citet{carlini2023extracting}. Note that we run this techniques two times; one for SD-2.1 model with only exact dedup removal as baseline, and for a model with the SD2.1 architecture but trained on removed exact duplication and near-duplication using SSCD~\citep{pizzi2022self}. 

We select the 350,000 most-duplicated examples from the training dataset based on SSCD~\citep{pizzi2022self} with threshold of 0.5, and generate 500 candidate images for each text prompt to increase the likelihood of finding memorization. The intuition is that for diffusion models, with high probability $Gen(p;r_1) \approx _d Gen(p;r_2)$ for two different random initial seeds $r_1$,$r_2$. On the other hand, if $Gen(p;r_1)\approx _d Gen(p;r_2)$ under some distance measure d, it is likely that these generated samples are memorized examples. To compute the distance measure $d$ between two images, we use a modified Euclidean $l_2$ distance. In particular, we found that many generations were often spuriously similar according to  $l_2$ distance (e.g., they all had gray backgrounds). We therefore instead divide each image into 16 non-overlapping 128 × 128 tiles and measure the maximum of the $l_2$ distance between any pair of image tiles between the two images. 
\Cref{fig:memorization_sscd} shows the comparison between number of memorized samples, before and after using SSCD with the threshold of 0.5 to remove near-duplicated samples. \citet{carlini2023extracting} mark images within clique size of 10 as memorized samples. Here we also explore different sizes for cliques. For all clique thresholds, SSCD is able to significantly reduce the number of memorized samples. Specifically, when the clique size is 10, trained SD models on the deduplicated training samples cut
off at SSCD$=0.5$ show a $5\times$ reduction in potentially memorized examples.

\algclusterdeduplication
\figdupcombined
\algmemorizationdetection
\figmembaselinessd

\end{document}